\documentclass[twoside]{article}

\usepackage[accepted]{aistats2023}
\usepackage{epsfig}
\usepackage{graphicx}
\usepackage{amsmath}
\usepackage{amssymb}
\usepackage{hyperref}
\usepackage[round]{natbib}
\usepackage{url}
\usepackage{booktabs}
\usepackage{amsfonts}
\usepackage{nicefrac}
\usepackage{microtype}
\usepackage{comment}
\usepackage{subfigure} 
\usepackage{pdflscape}
\usepackage{algorithm}
\usepackage{algorithmic}
\usepackage{bm}
\usepackage{soul}
\usepackage{listings}
\usepackage{bbm}
\usepackage{multirow}
\usepackage{multicol}

\newcommand\Tstrut{\rule{0pt}{2.2ex}}         % = `top' strut
\begin{document}



\twocolumn[

\aistatstitle{SwAMP: Swapped Assignment of Multi-Modal Pairs for Cross-Modal Retrieval}

\aistatsauthor{Minyoung Kim
}

\aistatsaddress{Samsung AI Center Cambridge, UK%\\
%{\tt\small mikim21@gmail.com}
} 
]

\begin{abstract}
We tackle the cross-modal retrieval problem, where learning is only supervised by relevant multi-modal pairs in the data. Although the contrastive learning is the most popular approach for this task, it makes potentially wrong assumption that the instances in different pairs are automatically irrelevant. To address the issue, we propose a novel loss function that is based on self-labeling of the unknown semantic classes. Specifically, we aim to predict class labels of the data instances in each modality, and assign those labels to the corresponding instances in the other modality (i.e., swapping the pseudo labels). With these swapped labels, we learn the data embedding for each modality using the supervised cross-entropy loss. This way, cross-modal instances from different pairs that are semantically related can be aligned to each other by the class predictor. We tested our approach on several real-world cross-modal retrieval problems, including text-based video retrieval, sketch-based image retrieval, and image-text retrieval.  For all these tasks our method achieves significant performance improvement over the contrastive learning. 
\end{abstract}

%%%%%%%%%%%%%%%%%%%%%%%%%%%%%%%%%%%%%%%%%%%%%%%%%%%%%%%%%%%%%%%%%%%%%%%%%%%%%%%
%%%%%%%%%%%%%%%%%%%%%%%%%%%%%%%%%%%%%%%%%%%%%%%%%%%%%%%%%%%%%%%%%%%%%%%%%%%%%%%
\section{Introduction}\label{sec:intro}

Cross-modal retrieval, the task of retrieving the most relevant items in the database of one modality (e.g., images) for a given query from another modality (e.g., texts), 
has received unprecedented attention in  computer vision and related areas~\citep{coco,vsepp,scan,vsr,caan,pcme,Miech_2021_CVPR,howto100m,howto100m_mil,t2vlad,doodle2search,Sain_2021_CVPR}. The crux of the problem is to learn the underlying relevance or similarity metric between data instances that live in heterogeneous modalities with highly different distributions. Although there are several different learning problem formulations in the literature, in this paper we mainly focus on the {\em paired} training data setup, in which training is only supervised by relevant pairs in the training data, and there are no semantic class labels annotated. That is, the training data consist of only pairs of relevant multi-modal data instances, %(Fig.~\ref{fig:concept}(a)), 
e.g., $(image, text)$, which may require minimal human annotation effort (e.g., web scraping of images and nearby texts). 

% Two approaches: modality-wise representation learning and cross-modal representation. The latter [inria group cvpr'21, SCAN] is more discriminative, but computationally more demanding... We focus on the former for computational benefit. 

%1) \textbf{Motivation} 
The contrastive (or triplet loss) learning~\citep{contrastive1,contrastive2} is recognised as the most popular and successful approach, which aims to learn the cross-modal similarity measure %%between instances from different modalities 
by the intuitive criteria that pull together relevant pairs and push away irrelevant ones. 
However, it makes potentially wrong assumption that instances in different pairs are automatically irrelevant. The pairs in the training data are usually collected by considering relevant pairs only (e.g., nearby images and texts %image/text snippets 
in a web page), and the relevance of instances in different pairs is usually not checked %(In Fig.~\ref{fig:concept}(b), the examples in the second and fourth rows are in fact relevant, but pushed away in the contrastive learning).
However, this is implicitly assumed in the contrastive loss. The issue was also raised in recent work~\citep{dml_bb19,ladder_loss,patrick20,video_retr21,mm_cluster}. In this paper we propose a novel learning algorithm that addresses the issue via self-labeled clustering approach. 
%However, it has two major drawbacks: i) its pair-based loss requires quadratic sampling complexity in the number of training data points (all possible pairs in Fig.~\ref{fig:concept}(b)), and ii) it makes potentially wrong assumption that the instances in different pairs are automatically irrelevant. The pairs in the training data are usually collected by considering relevant pairs only (e.g., nearby image/text snippets in a web page), and the relevance of instances in different pairs is usually not checked (In Fig.~\ref{fig:concept}(b), the examples in the second and fourth rows are in fact relevant, but pushed away in the contrastive learning). However, this is implicitly assumed in the contrastive loss. In this paper we propose a novel learning algorithm that addresses these drawbacks. 

Motivated from the recent %ideas of
clustering-based representation learning in the self-supervised learning literature~\citep{sela,swav}, we propose a novel loss function for cross-modal retrieval that is based on self-labeling of the unknown classes. Specifically, we introduce (latent) semantic class labels to be assigned to data instances, where class labels supposedly decide the relevance of cross-modal data instances (i.e., the same class label means relevant items, and vice versa). We predict class labels of the data instances in each modality, and assign the predicted labels to the corresponding instances in the other modality (i.e., swapping the pseudo labels). With these swapped pseudo labels, we learn the data  embedding for each modality using the supervised cross-entropy loss. %, hence leading to linear sampling complexity.
This way, cross-modal instances from different pairs that are semantically related can be aligned to each other by the class predictor. 
The whole process of label prediction and supervised learning with swapped classes is alternated to learn the optimal feature extraction networks. We call this approach {\em Swapped Assignment of Multi-modal Pairs} (SwAMP). %, and the overall idea is illustrated in Fig.~\ref{fig:concept}(c). %illustrates the idea of our class swapped assignment approach (SwAMP). \hl{Add a concept diagram with captions describing low sampling complexity and no-automatic negative relevance attained by the proposed swamp loss.}

The main benefits of the SwAMP are in two folds: 
i) Unlike the  contrastive loss, SwAMP does not make potentially wrong assumption that instances from different pairs are automatically irrelevant. The optimized class assignment finds similar instances from other pairs, and the feature extractor is trained in such a way that the same-class instances, even in different pairs, are well aligned. This feature of aligning instances in different pairs is hardly exploited in the contrastive loss. 
ii) Since the learning does not fully resort to pair-based losses as in contrastive learning, the sampling complexity can be reduced. This comes from the class-based loss adopted in the SwAMP, where similar ideas were exploited previously %in the proxy- or prototype-based learning (e.g., deep metric learning~\citep{proxy_nca,pa_cvpr20} and 
in self-supervised representation learning~\citep{selfsup_kmeans,sela,swav}.
%
%This can reduce the training time significantly... And avoids making the negative relevance assumption, and the relevance are discovered by the optimized cluster assignment..
%
%\textbf{Pair-based vs. proxy-based loss} Analogy to proxy-based (we learn class prototypes, and the features are learned to conform to relevance induced by the class assignments). When we emphasize the computational benefit of the SwAMP, perhaps it is important to mention other related proxy-based approaches in DML (the same motivation as swamp, but they are developed for supervised learning DML setup). 
%
Our approach is generically applicable to different types of cross-modal retrieval problems. We empirically demonstrate that the SwAMP loss improves retrieval performance significantly over the contrastive learning, on various real-world cross-modal retrieval problems, including text-video, sketch-image, and image-text retrieval. 

%\textbf{Contributions.} i) ii) iii) 

%%%%%%%%%%%%%%%%%%%%%%%%%%%%%%%%%%%%%%%%%%%%%%%%%%%%%%%%%%%%%%%%%%%%%%%%%%%%%%%
%%%%%%%%%%%%%%%%%%%%%%%%%%%%%%%%%%%%%%%%%%%%%%%%%%%%%%%%%%%%%%%%%%%%%%%%%%%%%%%
\section{Problem Setup $\&$ Background}\label{sec:setup}

%%%%%%%%%%%%%%%%%%%%%%%%%%%%%%%%%%%%%%%%%%%%%%%%%%%%%%%%%%%%%%%%%%%%%%%%%%%%%%%
%\subsection{Problem Setup and Notations}\label{sec:notations}

Let $x^A$ and $x^B$ denote data instances from modality A and modality B, respectively. For instance, $x^A$ is an image from the image modality, while $x^B$ is a text/caption from the text modality. 
Throughout the paper we deal with {\em modality-wise feature representation}, meaning that we have modality-wise feature extractors (neural networks) $\phi^A(\cdot)$ and $\phi^B(\cdot)$ applied to $x^A$ and $x^B$, respectively. Also known as {\em dual encoders}, it produces a succinct vector representation for each modality, $\phi^A(x^A)\in \mathbb{R}^d$ and $\phi^B(x^B)\in \mathbb{R}^d$. %Note that we assume the same feature dimensionality (denoted by $d$) for both modalities,  
The shared feature space ($\subset \mathbb{R}^d$) allows us to define the similarity score $s(x^A,x^B)$
as a cosine angle between $\phi^A(x^A)$ and $\phi^B(x^B)$. % as their cosine angle: 
% %%%%
% \vspace{-0.5em}
% \begin{equation}
% \vspace{-0.5em}
% s(x^A,x^B) = \frac{\phi^A(x^A)^\top \phi^B(x^B)} {||\phi^A(x^A)|| \cdot ||\phi^B(x^B)||}.
% \label{eq:cossim}
% \end{equation}
% %%%%
The goal %of the cross-modal retrieval 
is to learn the feature extractors so that the relevant pairs $x^A$ and $x^B$ have a high similarity score $s(x^A,x^B)$, while irrelevant pairs have a low similarity score. 
The main benefit of the modality-wise feature representation is the computational efficiency, scalable to billions of instances at training/test time, thanks to the efficient dot-product. There is an alternative approach that directly computes the similarity score without having modality-wise representation. A typical example is the {\em cross-modal attention} models~\citep{scan,ca1,ca2,ca3} (details in Sec.~\ref{sec:related}). %, where the details are discussed in Sec.~\ref{sec:related}. 
%that directly model $s(x^A,x^B)$ as a transformer-like neural network.  %which aim to capture interactions between local features in the instances from different modalities. 
%Although they can capture cross-modal interactions between local features of data instances from different modalities, they are computationally demanding and very slow due to the quadratic complexity in the number of local features. %Refer to Sec.~\ref{sec:related} for more details and related works.
%MOVE TO RELATED WORK???
%Oftentimes, each data instance $x$ is formed as a set of local elements (e.g., an image as a set of local patches/features, and a text as a collection of words). In the cross-modal attention, they model interaction between local elements of an instance in one modality with those in the other modality in the attention/transformer fashion to define the final similarity score. 
Although they can capture interactions between cross-modal local features, they are  computationally demanding, not scalable to large-scale data.  %as the complexity is quadratic in the number of local features. %Recently in~\citep{Miech_2021_CVPR}, an hybrid of the two approaches is introduced, which retains the two models, but performs re-ranking/distillation at test time for speed-up. 

The training data are composed of relevant pairs $\mathcal{D} = \{(x_i^A,x_i^B)\}_{i=1}^N$, where $x_i^A$ and $x_i^B$ are the instances in the $i$-th relevant pair. %in the $i$-th pair are regarded as relevant pairs. 
At test time, a query is given from the query modality, say $x^A$, and the goal is to find the most relevant instance, say $x^B$, from the other modality, where the search is performed on the given test set $\{x_i^B\}_{i=N+1}^{N+M}$. %We describe the contrastive learning in the following, which is %recognized as %one of 
%the most popular and successful method for cross-modal retrieval. 

%%%%%%%%%%%%%%%%%%%%%%%%%%%%%%%%%%%%%%%%%%%%%%%%%%%%%%%%%%%%%%%%%%%%%%%%%%%%%%%
\subsection{Contrastive Learning}\label{sec:contrastive}

In contrastive learning~\citep{contrastive1,contrastive2}, it is implicitly assumed that data instances from different pairs are irrelevant, although it may not be true. 
%(see the paragraph below). 
The loss function is defined to capture the intuition: penalize low (high) similarity scores for relevant (irrelevant, resp.) pairs. By introducing the margin $\alpha$ (e.g., 0.2) and considering the most violating irrelevant pairs (i.e., hard negatives), the loss can be written as (subscript {\em c} stands for contrastive): % hard negative mining):
%%%%
%\vspace{-0.5em}
\begin{align}
%\vspace{-0.5em}
% \mathcal{L}_{c}(\phi^A,\phi^B) = %; \mathcal{B}) =
% \sum_{i\in\mathcal{D}} 
% \big( s(x^A_i,x^B_i) - %\max_{ \substack{j\in \mathcal{B}\\ j\neq i}}
% \max_{j\in \mathcal{D}\setminus i} s(x^A_i,x^B_j) \big)_{\geq\alpha} + \big( s(x^A_i,x^B_i) - %\max_{\substack{j\in \mathcal{B}\\ j\neq i}} 
% \max_{j\in \mathcal{D}\setminus i} s(x^A_j,x^B_i) \big)_{\geq\alpha}
%
\mathcal{L}_{c}(\phi^A,\phi^B) &= %; \mathcal{B}) =
\sum_{i\in\mathcal{D}} 
\big( s(x^A_i,x^B_i) - %\max_{ \substack{j\in \mathcal{B}\\ j\neq i}}
\max_{j\in \mathcal{D}\setminus i} s(x^A_i,x^B_j) \big)_{\geq\alpha} \nonumber \\
& \ \ \ \ + \ \big( s(x^A_i,x^B_i) - %\max_{\substack{j\in \mathcal{B}\\ j\neq i}} 
\max_{j\in \mathcal{D}\setminus i} s(x^A_j,x^B_i) \big)_{\geq\alpha}
\label{eq:contrastive_loss}
\end{align}
%%%%
where $(z)_{\geq\alpha} = \max(0,\alpha-z)$ only incurs positive loss when $z<\alpha$. %, and $\alpha$ is the margin (e.g., 0.2).
A main issue of the contrastive learning is that we cannot guarantee that data instances from different pairs in the training data are irrelevant, because the data are usually collected by considering relevant pairs only (e.g., web scraping of images and nearby texts), and the relevance of instances in different pairs is usually not checked. However, this is assumed in the contrastive loss. 

%There are two main issues of the contrastive learning: i) We cannot guarantee that data instances from different pairs in the training data are irrelevant, because the data are usually collected by considering relevant pairs only (e.g., web scraping of images and nearby texts), and the relevance of instances in different pairs is usually not checked. However, this is assumed in the contrastive loss. ii) As shown in (\ref{eq:contrastive_loss}), the loss function contains all possible (relevant and irrelevant) cross-modal pair terms, and thus the complexity is quadratic $O(N^2)$. In practice, this can be circumvented by the minibatch loss (i.e., $\mathcal{D}$ replaced by a minibatch $\mathcal{B} \subset \mathcal{D}$ in (\ref{eq:contrastive_loss})), however it still requires quadratic sampling complexity to go through entire cross-modal pairs. %, relevant or irrelevant, in training data. 

%\hl{two main drawbacks: quadratic sampling complexity, automatic irrelevance for different pairs}

%%%%%%%%%%%%%%%%%%%%%%%%%%%%%%%%%%%%%%%%%%%%%%%%%%%%%%%%%%%%%%%%%%%%%%%%%%%%%%%
%%%%%%%%%%%%%%%%%%%%%%%%%%%%%%%%%%%%%%%%%%%%%%%%%%%%%%%%%%%%%%%%%%%%%%%%%%%%%%%
\section{Our Approach: SwAMP}\label{sec:swamp}

Our idea is to introduce {\em (latent) semantic class labels} for data instances and use them to learn the feature extractors. The class labels supposedly decide the relevance of data instances from different modalities, that is, $x^A$ and $x^B$ are considered relevant if their class labels are the same, and vice versa. Obviously, the paired cross-modal instances in the training data must have the same class labels. But beyond this, instances from different pairs can also be deemed relevant if they belong to the same semantic class labels. The motivation is that if we estimate the class labels accurately, the feature extractor learning can be turned into a supervised classification problem. % of linear sampling complexity. 

More formally, we consider (unknown) class labels to be assigned to the data instances. Let $y^A, y^B \in \{1,\dots,K\}$ be the class labels for $x^A$ and $x^B$, respectively, where $K$ is chosen by the user. The relevance of %cross-modal data instances 
$x^A$ and $x^B$ is determined by their class labels: $x^A$ and $x^B$ are deemed relevant if $y^A=y^B$ and irrelevant if $y^A \neq y^B$. If we knew the class labels that bear such semantics in the training data, then training becomes supervised learning that can be done for each modality, which allows us to avoid pairwise terms in the loss function. %, leading to linear sampling complexity. 
However, we don't have class labels, and we optimize them (i.e., self-supervised learning) together with the feature extractors $\phi^{A}(\cdot)$ and $\phi^{B}(\cdot)$.
To this end, we build linear classifiers $p(y|x^A)$ and $p(y|x^B)$ on the extracted features. For each modality $M \in \{A,B\}$, %by attaching (linear) classification heads to the outputs of the feature extractors, 
%%%%
%\vspace{-0.5em}
\begin{align}
%\vspace{-0.5em}
p(y=j|x^M) = \frac{\exp(p_j^\top \phi^M(x) / \tau)}{\sum_{l}\exp(p_{l}^\top \phi^M(x) / \tau)}, %\ \ \ \ %M = A \textrm{ or } B
%M \in \{A,B\}
% , \ \ \ \
% p(y=j|x^B) = \frac{\exp(p_j^\top \phi^B(x) / \tau)}{\sum_{j'=1}^{K}\exp(p_{j'}^\top \phi^B(x) / \tau)}, 
\label{eq:class_predictive}
\end{align}
%%%%
where $P=\{p_1,\dots,p_K\}$ are trainable parameters that are shared between two modalities, and $\tau$ is the temperature in the softmax. We can regard each $p_j$ as the {\em  prototype vector} for class $j$ that lies in the shared feature space. % induced from $\phi^{A/B}(\cdot)$. 
Since we have classification models, the (supervised) cross-entropy loss minimization is a natural choice to optimize them. That is, letting $p_{true}(y|x^A)$ be the true conditional class distribution for modality $A$, we minimize $\mathbb{E}_{p_{true}(y|x^A)}[-\log p(y|x^A)]$ with respect to $P$ and the network parameters of $\phi^A(\cdot)$ (similarly for modality $B$). Since we cannot access $p_{true}(y|x^A)$, one may be tempted to use the model $p(y|x^A)$ in (\ref{eq:class_predictive}) instead. However, it can easily lead to a degenerate solution such as the one that puts all the probability mass on a particular single class all the time (thus attaining the optimal cross-entropy loss $0$). Moreover, this would make learning $\phi^A(\cdot)$ and $\phi^B(\cdot)$ nearly independent and less interacted with each other, %where two being interacted 
merely through the shared prototypes $P$. %, and we are unable to exploit the paired data supervision available in the training data. 

Instead, we form an optimization problem to estimate a surrogate of $p_{true}(y|x^A)$, which we denote by $q(y|x^A)$, using the information from the other modality $B$, while imposing additional constraints to avoid the degenerate solutions. More specifically, we optimize the surrogate $q(y|x^A)$ with the following two criteria. First, $q(y|x^A)$ needs to be well aligned with the current estimate $p(y|x^B)$ for $x^B$ that is paired with $x^A$. This is due to the aforementioned requirements for the class labels, where the class labels (more generally, their distributions) of the paired instances should match. Secondly, the marginal distribution $q(y)=\mathbb{E}_{x^A \sim \mathcal{D}}[q(y|x^A)]$ is constrained to be a {\em uniform} distribution\footnote{This means balanced clusters. Even when data exhibit imbalance in semantic classes (e.g., long-tail distributions), our clustering model can still handle it by learning semantically redundant multiple clusters, thus forming {\em super}-clusters while rendering others minor classes. See Sec.~\ref{sec:qualitative} for illustration.}. This constraint naturally arises from the symmetry of class labels, a reasonable assumption about the true class distribution, and successfully leaves out the degenerate solutions discussed above. 
To summarize, the following is the optimization problem for $q(y|x^A)$, where $Q^A$ is the $(N \times K)$ matrix with $Q^A_{iy}:=q(y|x_i^A)$. % for each and every $x^A\in\mathcal{D}$.
Recall that $\mathcal{D} = \{(x_i^A,x_i^B)\}_{i=1}^N$ is the training data of paired instances. 
%%%%
\begin{align}
\min_{Q^A} &\ \ \mathbb{E}_{i\sim\mathcal{D}}\big[ \mathbb{E}_{q(y|x^A_i)}[-\log p(y|x^B_i)]\big] \label{eq:ot_q} \\
\textrm{s.t.} &\ \ \mathbb{E}_{i \sim \mathcal{D}}[q(y|x_i^A)] = 1/K, \ \forall{y}. %\frac{1}{K}. 
\nonumber
\end{align}
%%%%
% %%%%
% %\vspace{-0.5em}
% \begin{align}
% %\vspace{-0.5em}
% \min_{Q^A} \ \ \mathbb{E}_{i\sim\mathcal{D}}\big[ \mathbb{E}_{q(y|x^A_i)}[-\log p(y|x^B_i)]\big] %\nonumber \\
% \ \ 
% \textrm{s.t.} \ \ \mathbb{E}_{i \sim \mathcal{D}}[q(y|x_i^A)] = 1/K, \ \forall{y}. %\frac{1}{K}. 
% \label{eq:ot_q}
% \end{align}
% %%%%

We perform similar optimization for $q(y|x^B)$ ($Q^B_{iy}:=q(y|x_i^B)$) to approximate $p_{true}(y|x^B)$ by exchanging the roles of $A$ and $B$. The optimal solutions (surrogates) are denoted by $q^A$ and $q^B$, where we use the superscript to distinguish the two modalities. Note that during the optimization of (\ref{eq:ot_q}) for $q^A$ and $q^B$, we fix the model parameters, that is, $P$ and the feature extractor networks. The overall optimization is {\em alternation} between: i) surrogate optimization (\ref{eq:ot_q}) with $P$, $\phi^A$, $\phi^B$ fixed, and ii) supervised (cross-entropy) loss minimization with $q^A$ and $q^B$ fixed, where the latter can be written as (subscript $s$ stands for SwAMP):
%%%%
\begin{align}
\min_{P,\phi^A,\phi^B} \mathcal{L}_{s} := \ & %(P,\phi^A,\phi^B) := 
\mathbb{E}_{i\sim\mathcal{D}}\big[ \mathbb{E}_{q^A(y|x_i^A)}[-\log p(y|x_i^A)]\big] \ + \nonumber \\
& \ \ \mathbb{E}_{i\sim\mathcal{D}}\big[
\mathbb{E}_{q^B(y|x_i^B)}[-\log p(y|x_i^B)]\big]
\label{eq:ce_surrogate}
\end{align}
%%%%

Now we discuss how to optimize (\ref{eq:ot_q}). It is essentially the optimal transport (OT)  problem~\citep{ot_villani,ot_cuturi},  specifically with the cost matrix $C_{iy} = -\log p(y|x_i^B)$ and the marginal constraints $\sum_i Q^A_{iy}=1/K, \forall y$ (and implicitly $\sum_y Q^A_{iy}=1/N, \forall i\in\mathcal{D}$). Although the OT is known to be an instance of the linear program (LP), conventional LP solvers are not suitable for large-scale problems. As is common practice, we relax the problem by augmenting the loss with the entropic regularizer for $q(y|x^A)$, namely $\frac{1}{\eta} \sum_{iy} Q^A_{iy}\log Q^A_{iy}$ added to the loss (thus, penalizing small entropy), which can be solved by the efficient Sinkhorn-Knopp (SK)  algorithm~\citep{ot_cuturi}. Here $\eta$ is the regularization trade-off hyperparameter. The SK algorithm finds the optimal solution as $Q^A = \textrm{Diag}(u) A \textrm{Diag}(v)$,  where $A_{iy}=e^{-\eta C_{iy}}$ and the vectors $u \in\mathbb{R}^N_+$ and $v\in\mathbb{R}^K_+$ are the fixed points of $u_i = \frac{1}{N} / (Av)_i$, $v_j = \frac{1}{K} / (A^\top u)_j$ for $i=1,\dots,N$, $j=1,\dots,K$. The fixed point iteration usually converges quickly after a few iterations. We denote the algorithm as:
%%%%
\begin{equation}
Q \leftarrow \textrm{SK}(\textrm{cost}=C, \textrm{reg}=\eta).
\end{equation}
%%%%

One challenge in optimizing (\ref{eq:ot_q}) with the SK, however, is that it involves the entire dataset $\mathcal{D}$ in the loss, which means that the model update (\ref{eq:ce_surrogate}) has to be deferred until $q$ is optimized for an entire data epoch. Simply replacing $\mathcal{D}$ with a minibatch might be dangerous since the population class marginal distributions are poorly covered by a minibatch. We need an even larger subset of $\mathcal{D}$ to roughly meet the (uniform) class constraint. 
To this end, we adopt the (FIFO) queues, where we accumulate the embeddings $\phi^A(x^A)$ and $\phi^B(x^B)$ from the latest minibatches into the queues. The optimization (\ref{eq:ot_q}) is then performed on the queue data ($\mathcal{D}$ replaced by the data in the queues). To have the uniform class constraint meaningful, we choose the queue size to be greater than $K$. 
%several multiples of the class cardinality $K$ (In our empirical study, we typically choose $K=1000$ and the queue size is $5000$).
Note that (\ref{eq:ot_q}) is solved by the SK algorithm, and thus no backprop is required, hence enlarging the queue size does not incur computational issue. Similar ideas were used in the self-supervised representation learning literature, e.g.,~\citep{he2019moco} and~\citep{swav}.
%The queues need to contain latest fresh features, that's why we used the FIFO queues. 
To have the queues filled with the latest features, we insert the features of the current minibatch into the queues, then perform the SK algorithm. Once (\ref{eq:ot_q}) is done, we can optimize (\ref{eq:ce_surrogate}) by gradient descent, but only the current minibatch portion of $q$ is used. %That is, we only use the portion of $q$'s that belong to the current batch. 
The final loss function is a combination of the SwAMP loss and the contrastive loss,
%%%%
\begin{equation}
\mathcal{L}(P,\phi^A,\phi^B) = \mathcal{L}_c(\phi^A,\phi^B) + \lambda \mathcal{L}_s(P,\phi^A,\phi^B),
\label{eq:loss_comb}
\end{equation}
%%%%
where $\lambda$ is the trade-off hyperparameter. % indicating impact of the SwAMP loss part. 
%\hl{Defend why adding the contrastive loss term is required and how it diminishes the claim of sub-quadratic complexity. Eg, contrastive loss alone requires quadratic sample complexity, but with the help of the swamp loss we reduce the sample complexity, as evidenced by empirical results.}

As we estimate the surrogate $q^A$ using the current classification model in modality $B$, and vice versa, the class assignment is {\em swapped}. %Therefore we name this approach {\em SwAMP} (Swapped Assignment of Multi-modal Pairs). 
The pseudo code of our algorithm is shown in Alg.~\ref{alg:swamp}.
The idea of optimizing class labels in the representation learning was previously introduced in~\citep{sela,swav}, however, they aimed for self-supervised representation learning as an instance discrimination pretext task with augmented data. In this paper, we deal with the {\em cross-modal retrieval} problem, where we estimate the class labels of instances in one modality using the features from the other modality. 
%Unlike SeLa, we aim to estimate the class labels for one modality (A), and with the estimated labels we learn the embedding of the other modality (B). 
%(X) Do we derive the JoAMP-no-mono-modal like formulation? ie, setting $y_i := y_i^A = y_i^B$ for relevant pairs? This will end up with the OT cost equal to the sum of negative class log-lik for two modalities $cost = -\log p(y_i|x_i^A) - \log p(y_i|x_i^B)$. \hl{If we use this, then the benefit is it becomes different from SeLa and SwAV, slightly different though}.
%
%Before we conclude the section, we emphasize another benefit of the SwAMP. 
Unlike the  contrastive loss, SwAMP does not make any assumption that instances from different pairs are automatically irrelevant. The OT class assignment finds similar instances from other pairs, and the feature extractor is trained in such a way that the same-class instances, even in different pairs, are well aligned. This feature of aligning instances in different pairs is hardly exploited in the contrastive loss.

%\hl{Our final loss is combination of the contrastive and swamp losses, by the trade-off parameter lambda.}

%%%%
\newcommand\inlineeqno{\stepcounter{equation}\ (\theequation)}
\newcommand{\INDSTATE}[1][1]{\STATE\hspace{#1\algorithmicindent}}
%%%%
%\vspace{-1.5em}
\begin{algorithm}[t!]
\caption{SwAMP Training. %\hl{Modify this to PyTorch style??}
}
\label{alg:swamp}
\begin{small}
\begin{algorithmic}
%%%
%\STATE {\bfseries [Meta training]}
\STATE \textbf{Input:} Class cardinality $K$,
queue size, %$|Q^A|=|Q^B|$, the 
%softmax 
temp.~$\tau$, %trade-off 
$\eta$ in SK.
\STATE \textbf{Initialize:} %Prototypes
$P=\{p_k\}_{k=1}^K$, 
%feature extractors 
$\phi^A$, $\phi^B$. %$\phi^A(\cdot)$, $\phi^B(\cdot)$.
Empty queue $\mathcal{Q}$. % and $\mathcal{Q}^B$. 
\STATE \textbf{Output:} Trained model  $\{P,\phi^A(\cdot),\phi^B(\cdot)\}$.
\STATE \textbf{Repeat until convergence:}
    \INDSTATE[1] 1. Sample a minibatch of paired data $\mathcal{B} = \{(x^A_i,x^B_i)\}$.
    \INDSTATE[1] 2. Evaluate $\phi^A(x^A_i)$ and $\phi^B(x^B_i)$ for $i\in\mathcal{B}$ (forward pass). 
    \INDSTATE[1] 3. Insert $\{(\phi^A(x^A_i), \phi^B(x^B_i))\}_{i\in\mathcal{B}}$ into the queue $\mathcal{Q}$.  %$\mathcal{Q}^A \leftarrow \{\phi^A(x^A)\}_{x^A\in\mathcal{B}}$, $\mathcal{Q}^B \leftarrow \{\phi^B(x^B)\}_{x^B\in\mathcal{B}}$.
    \INDSTATE[1] 4. Solve (\ref{eq:ot_q}) for modality $A$ and $B$: \\
    \ \ \ \ \ \ \ $\{q^A(y|i)\}_{i\in \mathcal{Q}} \leftarrow $ SK(cost=$\{-\log p(y|x^B_i)\}_{i\in \mathcal{Q}}$, reg$=\eta$)\\
    %\INDSTATE[1] 5. Similarly, solve it for modality $B$: 
    \ \ \ \ \ \ \ $\{q^B(y|i)\}_{i\in \mathcal{Q}} \leftarrow $ SK(cost=$\{-\log p(y|x^A_i)\}_{i\in \mathcal{Q}}$, reg$=\eta$)
    \INDSTATE[1] 5. %Take the portions of $q^A(y|i)$ and $q^B(y|i)$ that correspond to the current minibatch $i \in \mathcal{B}$, 
    Take the minibatch portions $\{q^A(y|i),q^B(y|i)\}_{i \in \mathcal{B}}$; %, and %\\
    %\ \ \ \ \ \ \ \ \ \ 
    \\ \ \ \ \ \ \ \ \ \ \ Do SGD update with $\mathcal{L}$ in (\ref{eq:loss_comb}).
\end{algorithmic}
\end{small}
\end{algorithm}
%\vspace{-1.5em}
%%%%

%%%%%%%%%%%%%%%%%%%%%%%%%%%%%%%%%%%%%%%%%%%%%%%%%%%%%%%%%%%%%%%%%%%%%%%%%%%%%%%
%%%%%%%%%%%%%%%%%%%%%%%%%%%%%%%%%%%%%%%%%%%%%%%%%%%%%%%%%%%%%%%%%%%%%%%%%%%%%%%
\section{Related Work}\label{sec:related}

\textbf{Cross-modal retrieval.} It is beyond the scope of the paper to enumerate all previous works on cross-modal retrieval, and we refer the readers to recent survey papers such as~\citep{cm_survey}. Recently, the most interesting cross-modal tasks involve, among others, video-text~\citep{colab_exp19,mmt,patrick20,Miech_2021_CVPR,howto100m,howto100m_mil,t2vlad,mm_cluster}, image-text~\citep{coco,vsepp,scan,pcme,vsr,caan}, and sketch-photo~\citep{doodle2search,Sain_2021_CVPR}. For the training data of relevant pairs, most approaches commonly rely on the idea of contrastive learning~\citep{contrastive1,contrastive2}.  Beyond the intuitive triplet forms~\citep{triple_1,triple_2}, more sophisticated losses were introduced in~\citep{n_pair,lifted_str,ranked_list,ms} to deal with a positive and multiple negative pairs as well as hard examples. To reduce the  super-linear time computational overhead, several sophisticated sampling strategies were proposed~\citep{sampling_matters,smart_mining,hard_aware}. 
As discussed in Sec.~\ref{sec:setup}, there are broadly two different ways to define the similarity metric between instances of different modalities: modality-wise feature representation and cross-modal attention. The main benefit of the former is the computational efficiency, scalable to billions of instances at training/test time, thanks to the efficient dot-product. The latter directly computes the similarity score without having modality-wise representation~\citep{scan,ca1,ca2,ca3} using the transformer-like attentive neural networks which aim to capture interactions between local features in the instances from different modalities. 
Although they can capture cross-modal interactions between local features of data instances from different modalities, they are computationally demanding and very slow due to the quadratic complexity in the number of local features. 
%MOVE TO RELATED WORK???
%Oftentimes, each data instance $x$ is formed as a set of local elements (e.g., an image as a set of local patches/features, and a text as a collection of words). In the cross-modal attention, they model interaction between local elements of an instance in one modality with those in the other modality in the attention/transformer fashion to define the final similarity score. Although it captures interactions between every pair of cross-modal local features, it is computationally demanding as the complexity is quadratic in the number of local features. 
In~\citep{Miech_2021_CVPR}, a hybrid of the two is introduced, which retains the two models, but performs re-ranking/distillation at test time for speed-up.

\textbf{Clustering-based approaches.} %\hl{for self-sup representation learning, not for cross-modal retrieval..} 
There were previous attempts to cluster (group) data instances, or equivalently self-labeling, to improve saliency in representation learning. Some approaches perform offline K-means clustering for every epoch~\citep{selfsup_kmeans,xdc}, which can make training slow. The idea of optimizing class labels in the representation learning was previously introduced in~\citep{sela,swav}. However, all these previous approaches aimed for self-supervised representation learning as an instance discrimination pretext task with augmented data. On the other hand, we perform simultaneous learning of class labels and the feature extraction networks for the cross-modal retrieval setting. More recently 
\citep{mm_cluster} proposed a clustering-based cross-modal retrieval method based on the semantic similarity. However, our approach is mainly different from it in that we adopt the OT-based class label assignment forming a joint feature-label optimization, instead of simple fusion of multi-modal features for clustering as in~\citep{mm_cluster}.

%, where we estimate the class labels of instances in one modality using the features from the other. 

%All these they mainly aimed for self-supervised representation learning for video-text multi-modal data, not intended for cross-modal retrieval as we did. 
%Another self-supervised representation learning, not directly clustering-based though, NN-Aug~\citep{nn_aug} considers nearest neighbors are positive examples instead of augmented samples. 

%\textbf{Proxy-based loss.} Beyond the pair-based contrastive loss, we can form a loss function based on the semantic class labels of data instances, known as proxy-based loss. Analogous to that we learn class prototypes in our SwAMP loss, the proxy-based methods introduce learnable proxy vectors (class representatives), one for each class, and pull together data instances that belong to the same class toward the proxies. As the loss function is defined solely with distances between data instances and proxy vectors without pairwise distances, it reduces the sampling complexity to linear. The idea was developed in the deep metric learning such as proxy-NCA~\citep{proxy_nca}, SoftTriple~\citep{soft_triple}, and the   proxy-anchor~\citep{pa_cvpr20}. However, unlike our SwAMP, they deal with the  supervised setup where the ground-truth semantic class labels are provided. % in the training data.

%%%%%%%%%%%%%%%%%%%%%%%%%%%%%%%%%%%%%%%%%%%%%%%%%%%%%%%%%%%%%%%%%%%%%%%%%%%%%%%
%%%%%%%%%%%%%%%%%%%%%%%%%%%%%%%%%%%%%%%%%%%%%%%%%%%%%%%%%%%%%%%%%%%%%%%%%%%%%%%
\section{Experimental Results}\label{sec:expmts}

We test the proposed SwAMP loss on several different types of real-world cross-modal retrieval problems. %: text-based video retrieval (Sec.~\ref{sec:video_text}), sketch-based photo image retrieval (Sec.~\ref{sec:sketch}), and image-text retrieval (Sec.~\ref{sec:img_txt}). 
For each problem/dataset, we choose the most popular and successful method in the literature, and replace its loss function (mostly contrastive loss) with the proposed SwAMP loss to demonstrate the performance improvement. To this end, for fair comparison, we faithfully follow the same optimization strategy and hyperparameters as the baseline methods. % (detailed in the related sections). % and Supplement). %The performance metrics (e.g., recall, average precision) are slightly different from problem to problem, and we also follow the metrics used in the baseline methods.
\subsection{Text-based Video Retrieval}\label{sec:video_text}

%For the real-world problems, 
We first consider the text-to-video retrieval task where the goal is to find the most relevant video clip for a given natural language text query. We consider three datasets for this task: i) \textbf{YouCook2}~\citep{yc2} of cooking videos and instructions, ii) \textbf{MSR-VTT}~\citep{msrvtt} of generic videos and captions from YouTube, and iii) \textbf{LSMDC}~\citep{lsmdc} of movie clips and subtitles. All these datasets provide pairs of video clip and text description, forming a multi-modal paired data format $(text,video)$ which conforms to our SwAMP framework. 

%%%% 
\begin{table}[t!]
%\vspace{-1.5em}
\caption{%Performance improvement achieved by the proposed SwAMP against the contrastive learning 
Text-video retrieval results on YouCook2. %constrastiv learning~\citep{howto100m} (denoted by Contrastive). We also include the other approaches.
%The improved scores of SwAMP over contrastive are boldfaced. % for the two training strategies, No PT and PT. 
}
\vspace{+0.3em}
\centering
%\vspace{-0.8em}
%\vskip 0.05in
\begin{scriptsize}
%\begin{footnotesize}
%\begin{small}
%\begin{sc}
\centering
%\scalebox{0.95}{
\begin{tabular}{lcccc}
\toprule
Methods & R@1 $\uparrow$ & R@5 $\uparrow$ & R@10 $\uparrow$ & Med-R $\downarrow$
\\ \hline
Random\Tstrut & 0.03 & 0.15 & 0.3 & 1675 \\
FV-CCA%~\citep{fv_cca} 
& 4.6 & 14.3 & 21.6 & 75 \\
\hline
%
% MIL~\citep{howto100m_mil}\Tstrut & 6.1 & 17.3 & 24.8 & 46 \\
% \hline
%
Contrastive (No PT)\Tstrut & 4.2 & 13.7 & 21.5 & 65 \\
%
%Contrastive (No PT) $+$ 
SwAMP (No PT) & ${\bf 4.8}$ & ${\bf 14.5}$ & ${\bf 22.5}$ & ${\bf 57}$ \\
\hline
Contrastive (PT)\Tstrut & 8.2 & 24.5 & 35.3 & 24 \\
%
%Contrastive (PT) $+$ 
SwAMP (PT) & ${\bf 9.4}$ & ${\bf 24.9}$ & 35.3 & ${\bf 22}$ \\
\bottomrule
\end{tabular}
%}
%\end{sc}
%\end{footnotesize}
\end{scriptsize}
%\end{small}
\label{tab:yc2}
%\vspace{-1.5em}
\end{table}
%%%%

For the raw text/video features and the feature extractor networks, as well as the training/test protocols, we follow the methods in~\citep{howto100m}, and the details are described in Appendix (Sec.~C). %The raw video features are outputs of pre-trained 2D/3D CNNs while the raw text features are GoogleNews pre-trained word embeddings. Then the raw features pass through the sigmoid-gated linear transform where the gating functions are two-layer linear networks~\citep{gated_linear}. The details are described in Supplement (Sec.~C). 
%Whereas the details of the datasets and experimental setups are described in the subsequent sections, the  features are specifically built by the following procedures.
%
% First, the raw features are obtained by the pretrained networks: (a) raw video features (4096D) are  concatenation of frame-level and video-level features extracted from the pretrained 2D/3D CNNs (the ImageNet pre-trained Resnet-152~\citep{resnet152} for 2D features and the Kinetics~\citep{kinetics} pre-trained ResNeXt-101 16-frame model~\citep{resnext101} for 3D features), (b) raw text features (4096D) are the GoogleNews pre-trained
% word2vec embeddings~\citep{word2vec} for the pre-processed transcribed video narrations with the common stop words removed. Then the feature extractor networks $\phi^{video}(\cdot)$ and $\phi^{text}(\cdot)$ transform these raw features into 4096D features by the sigmoid-gated linear transform where the gating functions are two-layer linear networks~\citep{gated_linear}. We fix the raw features and train only the latter sigmoid-gated networks ($\sim 67M$ parameters). %, which comprise about 67M parameters. 
%
%Given a textual description, the goal is to retrieve representative video clips from a large pool of videos. We evaluate our learned embedding using the standard recall metrics R@1, R@5, R@10 and the median rank (Median R). We provide experimental results for the following domain-specific video description datasets.
%
Following~\citep{howto100m}, there are two training strategies: i) No-pretraining (No-PT) where the feature extraction networks are randomly initialized, and the training is done on the training split of the dataset, and ii) Pretraining (PT) where the feature extractors are first pretrained on the large-scale HowTo100M dataset~\citep{howto100m}, and finetuned on the target dataset. In~\citep{howto100m}, they adopt the contrastive (triplet) loss for training the feature extractors. Although we also compare our approach with the state-of-the-arts, 
% \footnote{Although there was more recent approach from the same authors~\citep{howto100m_mil}, we have not included it since they formed a zero-shot transfer problem, simply applying the HowTo100M-pretrained model without model fine-tuning on the target dataset. % (aka zero-shot transfer), thus 
% That is, the performance is dominantly
% influenced by the enormous dataset (HowTo100M) rather than the loss function employed.}, 
the main focus in this experiment is to demonstrate the performance improvement achieved by the proposed SwAMP loss against vanilla contrastive learning. The SwAMP hyperparameter $\lambda$, the weight/impact of the SwAMP loss against the contrastive loss in (\ref{eq:loss_comb}) is chosen as $\lambda=0.25$ for all three datasets, except the LSMDC-PT case for which  $\lambda=0.1$. 
 %empirically chosen. % (weight for the contrastive loss is always 1.0). We use $\lambda_{swamp}=0.25$ for all three datasets, except the LSMDC PT case for which  $\lambda_{swamp}=0.1$. 
We also choose temperature in softmax $\tau=0.25$, entropic regularization trade-off in SK $\eta=5.0$, the number of classes $K=500$, and the queue size $2,048$ for the SwAMP. The other learning hyperparameters common in SwAMP and contrastive losses are not changed from~\citep{howto100m}, and summarized in Appendix (Sec.~C).

%\st{Unlike the synthetic data experiment, training the model with the SwAMP loss alone did not learn the model properly. This can be originated from the high dimensionality of the embedded space and the overfitted feature extractor: if the (initial) class prediction is incorrect, the model  tends to adapt to it faithfully anyway, learning embeddings that conform to the wrong clustering of data points. %if the initial embeddings are poor, then class predictions are incorrect, leading to wrong clustering of data points, which exacerbate the situation preventing the models from learning better embeddings, the vicious cycle. 
%To alleviate this issue, we combine the SwAMP loss and the contrastive loss by the weighted sum, where the latter enforces direct inner product alignment in the latent space, and helps preventing the model from being adapted to wrong clustering.}

%%%% 
\begin{table*}[t!]
%\vspace{-1.0em}
\caption{%Performace improvement achieved by SwAMP 
Text-Video retrieval results on MSRVTT and LSMDC.}
\vspace{+0.3em}
\centering
%\vspace{-0.8em}
%\vskip 0.05in
\begin{scriptsize}
%\begin{footnotesize}
%\begin{small}
%\begin{sc}
\centering
%\scalebox{0.95}{
\begin{tabular}{lcccccccc}
\toprule
\multirow{2}{*}{Methods} & \multicolumn{4}{c}{MSRVTT} & \multicolumn{4}{c}{LSMDC} \\
\cmidrule(lr){2-5} \cmidrule(lr){6-9}
& \ R@1 $\uparrow$ & R@5 $\uparrow$ & R@10 $\uparrow$ & Med-R $\downarrow$ \ 
& \ R@1 $\uparrow$ & R@5 $\uparrow$ & R@10 $\uparrow$ & Med-R $\downarrow$ \ 
\\ \hline
Random\Tstrut & 0.1 & 0.5 & 1.0 & 500 & 0.1 & 0.5 & 1.0 & 500 \\
C+LSTM+SA+FC7%~\citep{c_lstm} 
& 4.2 & 12.9 & 19.9 & 55 & 4.3 & 12.6 & 18.9 & 98 \\
VSE-LSTM%~\citep{vse_lstm} 
& 3.8 & 12.7 & 17.1 & 66 & 3.1 & 10.4 & 16.5 & 79 \\
SNUVL%~\citep{snuvl} 
& 3.5 & 15.9 & 23.8 & 44 & 3.6 & 14.7 & 23.9 & 50 \\
Temporal Tessellation%~\citep{temporal_tessellation} 
& 4.7 & 16.6 & 24.1 & 41 & 4.7 & 15.9 & 23.4 & 64 \\
CT-SAN%~\citep{ct-san} 
& 4.4 & 16.6 & 22.3 & 35 & 4.5 & 14.1 & 20.9 & 67 \\
JSFusion%~\citep{jsfusion} 
& 10.2 & 31.2 & 43.2 & 13 & 9.1 & 21.2 & 34.1 & 36 \\
\hline
%
% MIL~\citep{howto100m_mil}\Tstrut & 4.0 & 9.8 & 14.0 & 137 \\
% \hline
%
Contrastive (No PT)\Tstrut & 12.1 & 35.0 & 48.0 & 12 & 7.2 & 18.3 & 25.0 & 44 \\
%
%Contrastive (No PT) $+$ 
SwAMP (No PT) & ${\bf 15.0}$ & ${\bf 38.5}$ & ${\bf 50.3}$ & ${\bf 10}$ & ${\bf 7.7}$ & ${\bf 19.3}$ & ${\bf 27.7}$ & ${\bf 40}$ \\
\hline
Contrastive (PT)\Tstrut & 14.9 & 40.2 & 52.8 & 9 & 7.1 & 19.6 & 27.9 & 40 \\
%
%Contrastive (w/ PT) $+$ 
SwAMP (PT) & ${\bf 19.0}$ & ${\bf 42.4}$ & ${\bf 55.2}$ & ${\bf 8}$ & ${\bf 8.3}$ & ${\bf 20.0}$ & ${\bf 28.9}$ & ${\bf 37}$ \\
\bottomrule
\end{tabular}
%}
%\end{sc}
%\end{footnotesize}
\end{scriptsize}
%\end{small}
\label{tab:msrvtt_lsmdc}
%\vspace{-2.5em}
\end{table*}
%%%%

% %%%% 
% \begin{table}%[b!]
% %\vspace{-2.5em}
% \caption{%Performance improvement achieved by SwAMP 
% Text-video retrieval results on MSRVTT. 
% }
% %\vspace{+0.3em}
% \centering
% \vspace{-0.8em}
% %\vskip 0.05in
% \begin{scriptsize}
% %\begin{footnotesize}
% %\begin{small}
% %\begin{sc}
% \centering
% %\scalebox{0.95}{
% \begin{tabular}{lcccc}
% \toprule
% Methods & R@1 $\uparrow$ & R@5 $\uparrow$ & R@10 $\uparrow$ & Med-R $\downarrow$
% \\ \hline
% Random\Tstrut & 0.1 & 0.5 & 1.0 & 500 \\
% %
% C+LSTM+SA+FC7%~\citep{c_lstm} 
% & 4.2 & 12.9 & 19.9 & 55 \\
% %
% VSE-LSTM%~\citep{vse_lstm} 
% & 3.8 & 12.7 & 17.1 & 66 \\
% %
% SNUVL%~\citep{snuvl} 
% & 3.5 & 15.9 & 23.8 & 44 \\
% %
% Temporal Tessellation%~\citep{temporal_tessellation} 
% & 4.7 & 16.6 & 24.1 & 41 \\
% %
% CT-SAN%~\citep{ct-san} 
% & 4.4 & 16.6 & 22.3 & 35 \\
% %
% JSFusion%~\citep{jsfusion} 
% & 10.2 & 31.2 & 43.2 & 13 \\
% \hline
% %
% %Contrastive (zero-shot transfer)
% % MIL~\citep{howto100m_mil}\Tstrut & 7.5 & 21.2 & 29.6 & 38 \\
% % \hline
% %
% Contrastive (No PT)\Tstrut & 12.1 & 35.0 & 48.0 & 12 \\
% %
% %Contrastive (w/o PT) $+$ 
% SwAMP (No PT) & ${\bf 15.0}$ & ${\bf 38.5}$ & ${\bf 50.3}$ & ${\bf 10}$ \\
% \hline
% %
% Contrastive (PT)\Tstrut & 14.9 & 40.2 & 52.8 & 9 \\
% %
% %Contrastive (w/ PT) $+$ 
% SwAMP (PT) & ${\bf 19.0}$ & ${\bf 42.4}$ & ${\bf 55.2}$ & ${\bf 8}$ \\
% %
% \bottomrule
% \end{tabular}
% %}
% %\end{sc}
% %\end{footnotesize}
% \end{scriptsize}
% %\end{small}
% \label{tab:msrvtt}
% \vspace{-0.5em}
% \end{table}
% %%%%

%%%%%%%%%%%%%%
%\subsubsection{YouCook2} %~\citep{yc2}}
\textbf{YouCook2.} 
%cooking videos in YouCook2~\citep{yc2}
This cooking video dataset collected from YouTube, contains 89 recipes and 14K video clips %that are 
annotated with textual descriptions from paid human workers. The test data are formed by taking 3.5K clips from the validation set, and the test set comprises of $3,350$ pairs. The retrieval performance metrics are recall-at-$k$ (R@k) with $k=1,5,10$ and the median rank (Med-R). Hence, the random guess attains R@1$=0.03\%$ Med-R=$1,675$. 
%YC2: temperature in softmax $\tau=0.25$, entropic regularization trade-off in SK $\lambda=5.0$, impact of the SwAMP loss is $0.25$, and $K=500$, queue size is 2048. 
%Number of test samples??
The results are summarized in Table~\ref{tab:yc2}. In the bottom four rows, we see the performance improvement achieved by the proposed SwAMP against the contrastive loss~\citep{howto100m}. For both training strategies, No PT (random model initialization) and PT (initialized with the HowTo100M-pretrained model), our SwAMP improves the retrieval performance significantly (e.g., about $12\%$ reduction in Median Rank for the No PT case). SwAMP also outperform the %As also reported in~\citep{howto100m}, the 
CCA baseline FV-CCA~\citep{fv_cca}. %, %with the features described as above are compared. 
\textbf{MSRVTT.}  
This %generic video-text 
dataset~\citep{msrvtt} collected from YouTube contains videos of specific categories including music, sports, and movie. There are 200K video-caption pairs obtained by human annotation. We follow the retrieval training/test protocol of~\citep{jsfusion,howto100m}. The test set consists of 1K pairs. 
%MSRVTT: (the same as YC2) temperature in softmax $\tau=0.25$, entropic regularization trade-off in SK $\lambda=5.0$, impact of the SwAMP loss is $0.25$, and $K=500$, queue size is 2048. 
As reported in Table~\ref{tab:msrvtt_lsmdc}, our SwAMP loss improves the performance over the contrastive learning significantly for both no-pretraining and pretraining cases: about $24\%$ in R@1 in the No PT case, and $27\%$ in the PT case. Furthermore, the SwAMP outperforms with large margin the state-of-the-arts: C+LSTM+SA+FC7~\citep{c_lstm}, VSE-LSTM~\citep{vse_lstm}, Temporal Tessellation~\citep{temporal_tessellation}, CT-SAN~\citep{ct-san}, and JSFusion~\citep{jsfusion}.  %fusion approach~\citep{jsfusion} with large margin. 

%%%% 
\begin{table*}[t!]
%\vspace{-0.5em}
\caption{%Performance improvement achieved by SwAMP 
Sketch-based image retrieval results. % results on three sketch datasets. 
The contrastive-learning-based Doodle2Search~\citep{doodle2search} (denoted by D2S) is compared with the proposed SwAMP learning. % (denoted by SwAMP). %We also include the other approaches ZSIH~[CITE] and CVAE~[CITE]
}
\vspace{+0.3em}
\centering
%\vspace{-0.8em}
%\vskip 0.05in
\begin{scriptsize}
%\begin{footnotesize}
%\begin{small}
%\begin{sc}
\centering
%\scalebox{0.95}{
\begin{tabular}{lccccccccc}
\toprule
\multirow{2}{*}{Methods / Datasets} & 
\multicolumn{3}{c}{Sketchy%~\citep{sketchy}
} & \multicolumn{3}{c}{TU-Berlin%~\citep{tu-berlin}
} & \multicolumn{3}{c}{QuickDraw%~\citep{doodle2search}
}
\\ \cmidrule(lr){2-4} \cmidrule(lr){5-7} \cmidrule(lr){8-10}
& mAP & mAP@200 & P@200 & mAP\Tstrut & mAP@200 & P@200 & mAP\Tstrut & mAP@200 & P@200 \\
\hline
ZSIH~\citep{zsih}\Tstrut & 25.40 & - & - & 22.00 & - & - & - & - & - \\
CVAE~\citep{cvae_sketch} & 19.59 & 22.50 & 33.30 & 0.50 & 0.90 & 0.30 & 0.30 & 0.60 & 0.30 \\
\hline
D2S~\citep{doodle2search}\Tstrut & 36.91 & 46.06 & 37.04 & 10.94 & 15.68 & 12.08 & 7.52 & 9.01 & 6.75 \\
%\hline
%
%D2S $+$ 
SwAMP & ${\bf 40.32}$ & ${\bf 51.94}$ & ${\bf 40.81}$ & ${\bf 17.63}$ & ${\bf 24.49}$ & ${\bf 19.75}$ & 
${\bf 8.19}$ & ${\bf 11.62}$ & ${\bf 9.10}$ \\
%${\bf 7.94}$ & ${\bf 10.98}$ & ${\bf 9.00}$ \\
%
\bottomrule
\end{tabular}
%}
%\end{sc}
%\end{footnotesize}
\end{scriptsize}
%\end{small}
\label{tab:sketch}
%\vspace{-2.0em}
\end{table*}
%%%%

%%%%%%%%%%%%%%
%\subsubsection{LSMDC}
\textbf{LSMDC.} 
%The LSMDC~\citep{lsmdc}\footnote{\url{https://sites.google.com/site/describingmovies/lsmdc-2016/movieretrieval}} is a 
This dataset of movie video clips is comprised of 101K video-caption pairs. The captions are collected either from the movie scripts or the audio descriptions. The test set contains 1K pairs. For this dataset, %we use the SwAMP hyperparameter (impact of the SwAMP loss against the contrastive loss) 
$\lambda = 0.1$ (impact of the SwAMP loss against contrastive) for the PT case. The results are shown in Table~\ref{tab:msrvtt_lsmdc}. Similar to the other two datasets, our SwAMP is consistently better than the contrastive learning (about $7\sim 9\%$ in Median Rank).

%%%%%%%%%%%%%%%%%%%%%%%%%%%%%%%%%%%%%%%%%%%%%%%%%%%%%%%%%%%%%%%%%%%%%%%%%%%%%%%
\subsection{Sketch-based Image Retrieval}\label{sec:sketch}

%We next test our SwAMP  on the sketch-based image retrieval task. %\hl{As shown in the figure}, 
In sketch-based image retrieval, 
the model takes a user's sketch (quick drawing) of an object as input query, and retrieves the photo images that correspond to the same object category as query's. %Sketch-to-image retrieval gains much  attention these days due to the pervasive availability of touch screens or similar drawing devices. 
We follow the recent %practical 
framework of~\citep{doodle2search} which  reports the state-of-the-art performance on the three large-scale sketch-image benchmarks: Sketchy-Extended~\citep{sketchy}, TU-Berlin-Extended~\citep{tu-berlin}, and QuickDraw-Extended~\citep{doodle2search}. The datasets roughly consist of 100--200 object classes with hundreds to thousands of sketch/photo images for each class. For all these datasets, we have {\em zero-shot} setting,  %learning tasks, 
meaning that %the training/validation/test
training/test splits have %images and sketches
instances from disjoint object categories.

In this experiment we aim to show the improvement in the retrieval performance when our SwAMP loss is augmented to the existing loss function. To this end, we follow the same embedding networks for images and sketches, as well as the same loss function as the Doodle2Search. %To briefly describe the loss function of the Doodle2Search, it 
The loss function consists of three losses: {\em Triplet loss} is the conventional triplet loss, {\em Domain loss} uses an adversarial domain classifier to penalize misalignment between embedding distributions of photo images and sketches, and {\em Semantic loss} urges the embeddings of the photo images and sketches to reconstruct the pretrained word embedding of the corresponding object word. We also use the same attention-based embedding networks for photo and sketch modalities.
Then, we add our SwAMP loss to the Doodle2Search's loss with the impact $\lambda=0.1$ for all three datasets. We use the queue size $1000$ ($2000$ for QuickDraw-Extended) and class cardinality $K=500$, softmax temperature $\tau=0.25$, entropic regularization impact $\eta=5.0$. The retrieval performances on the three datasets are summarized in Table~\ref{tab:sketch}. The performance metrics are mean average precision (mAP), mAP@200, and the precision-at-200 (P@200). As shown, our SwAMP loss when added to the existing contrastive-based loss, significantly improves the retrieval performance (about $9\%$ in mAP for Sketchy and about $60\%$ for TU-Berlin).

%\hl{Add retrieval examples/figures, compared with existing approaches.}

%%%% 
\begin{table*}[t!]
\vspace{-1.0em}
\caption{%Performance improvement achieved by SwAMP 
Image-text retrieval results on Flickr30K. %(Report R@1 only??)
}
\vspace{+0.3em}
\centering
%\vspace{-1.0em}
%\vskip 0.05in
\begin{scriptsize}
%\begin{footnotesize}
%\begin{small}
%\begin{sc}
\centering
\begin{tabular}{lcccccc}
\toprule
\multirow{2}{*}{Methods} & 
\multicolumn{3}{c}{Image $\to$ Text} &  \multicolumn{3}{c}{Text $\to$ Image}
\\ \cmidrule(lr){2-4} \cmidrule(lr){5-7} 
& R@1 & R@5 & R@10 & R@1 & R@5 & R@10 \\
\hline
DAN~\citep{dan}\Tstrut & 55.0 & 81.8 & 89.0 & 39.4 & 69.2 & 79.1 \\
DPC~\citep{dpc} & 55.6 & 81.9 & 89.5 & 39.1 & 69.2 & 80.9 \\
VSE++~\citep{vsepp} & 52.9 & - & 87.2 & 39.6 & - & 79.5 \\
SCO~\citep{sco} & 55.5 & 82.0 & 89.3 & 41.1 & 70.5 & 80.1 \\
\hline
%
%SCAN i-t AVG\Tstrut & ${\bf 67.9}$ & 89.0 & 94.4 & 43.9 & 74.2 & 82.8 \\
SCAN i-t AVG\Tstrut & $67.9$ & 89.0 & 94.4 & 43.9 & 74.2 & 82.8 \\
SCAN t-i AVG & 61.8 & 87.5 & 93.7 & 45.8 & 74.4 & 83.0 \\
%
%SCAN t-i AVG + i-t LSE & 67.4 & ${\bf 90.3}$ & ${\bf 95.8}$ & 48.6 & ${\bf 77.7}$ & ${\bf 85.2}$ \\
SCAN t-i AVG + i-t LSE & 67.4 & $90.3$ & $95.8$ & 48.6 & $77.7$ & $85.2$ \\
\hline
Contrastive-PAR\Tstrut & 65.7 & 86.8 & 92.4 & 48.2 & 75.8 & 84.2 \\
%SwAMP-PAR & 67.8 & 88.5 & 94.0 & ${\bf 49.1}$ & 76.1 & 83.7 \\
SwAMP-PAR & ${\bf 67.8}$ & ${\bf 88.5}$ & ${\bf 94.0}$ & ${\bf 49.1}$ & ${\bf 76.1}$ & ${\bf 83.7}$ \\
\bottomrule
\end{tabular}
%\end{sc}
%\end{footnotesize}
\end{scriptsize}
%\end{small}
\label{tab:flickr}
\end{table*}
%%%%
%
% %%%% 
% \begin{table}%[b!]
% %\vspace{+0.2em}
% \caption{%Performance improvement achieved by SwAMP 
% Image-text retrieval results (R@1) on Flickr30K. %(Report R@1 only??)
% }
% %\vspace{+0.3em}
% \centering
% \vspace{-0.6em}
% %\vskip 0.05in
% %\begin{scriptsize}
% %\begin{footnotesize}
% \begin{small}
% %\begin{sc}
% \centering
% %\scalebox{0.95}{
% \begin{tabular}{lcc}
% \toprule
% Methods & 
% I $\to$ T & T $\to$ I \\
% \hline
% %
% DAN~\citep{dan}\Tstrut & 55.0 & 39.4 \\
% %
% DPC~\citep{dpc} & 55.6 & 39.1 \\
% %
% VSE++~\citep{vsepp} & 52.9 & 39.6 \\
% %
% SCO~\citep{sco} & 55.5 & 41.1 \\
% %
% \hline
% %
% %SCAN i-t AVG\Tstrut & ${\bf 67.9}$ & 89.0 & 94.4 & 43.9 & 74.2 & 82.8 \\
% SCAN i-t AVG\Tstrut & $67.9$ & 43.9 \\
% %
% SCAN t-i AVG & 61.8 & 45.8 \\
% %
% %SCAN t-i AVG + i-t LSE & 67.4 & ${\bf 90.3}$ & ${\bf 95.8}$ & 48.6 & ${\bf 77.7}$ & ${\bf 85.2}$ \\
% SCAN t-i AVG + i-t LSE & 67.4 & 48.6 \\
% \hline
% %
% Contrastive-PAR\Tstrut & 65.7 & 48.2 \\
% %SwAMP-PAR & 67.8 & 88.5 & 94.0 & ${\bf 49.1}$ & 76.1 & 83.7 \\
% SwAMP-PAR & ${\bf 67.8}$ & ${\bf 49.1}$ \\
% %
% \bottomrule
% \end{tabular}
% %}
% %\end{sc}
% %\end{footnotesize}
% %\end{scriptsize}
% \end{small}
% \label{tab:flickr}
% %\vspace{-0.7em}
% \end{table}
% %%%%
%
%%%% 
\begin{table*}[t!]
\vspace{-1.0em}
\caption{%Performance improvement achieved by SwAMP 
Image-text retrieval results on MS-COCO. %(Report R@1 only??)
}
\vspace{+0.3em}
\centering
%\vspace{-1.0em}
%\vskip 0.05in
\begin{scriptsize}
%\begin{footnotesize}
%\begin{small}
%\begin{sc}
\centering
\begin{tabular}{lcccccccccccc}
\toprule
\multirow{3}{*}{Methods} & \multicolumn{6}{c}{5-fold (1K test images)} & \multicolumn{6}{c}{Entire (5K test images)}
\\ \cmidrule(lr){2-7} \cmidrule(lr){8-13} 
& \multicolumn{3}{c}{Image $\to$ Text} &  \multicolumn{3}{c}{Text $\to$ Image} & \multicolumn{3}{c}{Image $\to$ Text} &  \multicolumn{3}{c}{Text $\to$ Image}
\\ \cmidrule(lr){2-4} \cmidrule(lr){5-7} \cmidrule(lr){8-10} \cmidrule(lr){11-13}
& R@1 & R@5 & R@10 & R@1 & R@5 & R@10 & R@1 & R@5 & R@10 & R@1 & R@5 & R@10 \\
\hline
DPC~\citep{dpc}\Tstrut & 65.6 & 89.8 & 95.5 & 47.1 & 79.9 & 90.0 & 41.2 & 70.5 & 81.1 & 25.3 & 53.4 & 66.4 \\
VSE++~\citep{vsepp} & 64.6 & - & 95.7 & 52.0 & - & 92.0 & 41.3 & - & 81.2 & 30.3 & - & 72.4\\
GXN~\citep{gxn} & 68.5 & - & 97.9 & 56.6 & - & 94.5 & 42.0 & - & 84.7 & 31.7 & - & 74.6 \\
SCO~\citep{sco} & 69.9 & 92.9 & 97.5 & 56.7 & 87.5 & 94.8 & 42.8 & 72.3 & 83.0 & 33.1 & 62.9 & 75.5 \\
PCME~\citep{pcme} & 68.8 & - & - & 54.6 & - & - & 44.2 & - & - & 31.9 & - & - \\
\hline
SCAN i-t\Tstrut & 69.2 & 93.2 & 97.5 & 54.4 & 86.0 & 93.6 & 46.4 & 77.4 & 87.2 & 34.4 & 63.7 & 75.7 \\
SCAN t-i + i-t & 72.7 & 94.8 & 98.4 & 58.8 & 88.4 & 94.8 & 50.4 & 82.2 & 90.0 & 38.6 & 69.3 & 80.4 \\
\hline
Contrastive-PAR\Tstrut & 71.8 & 94.3 & 97.9 & 56.8 & 86.9 & 93.8 & 48.4 & 78.1 & 88.1 & 34.3 & 64.4 & 76.2 \\
%SwAMP-PAR & 72.6 & 94.6$ & 98.0$ & 57.4 & 87.6 & 94.1 & 49.7 & 79.1 & 88.3 & 35.0 & 65.1 & 76.6 \\
SwAMP-PAR & ${\bf 72.6}$ & ${\bf 94.6}$ & ${\bf 98.0}$ & ${\bf 57.4}$ & ${\bf 87.6}$ & ${\bf 94.1}$ & ${\bf 49.7}$ & ${\bf 79.1}$ & ${\bf 88.3}$ & ${\bf 35.0}$ & ${\bf 65.1}$ & ${\bf 76.6}$ \\
\bottomrule
\end{tabular}
%\end{sc}
%\end{footnotesize}
\end{scriptsize}
%\end{small}
\label{tab:coco}
%\vspace{-2.0em}
\end{table*}
%%%%

% %%%% 
% \begin{table}%[b!]
% %\vspace{-2.5em}
% \caption{%Performance improvement achieved by SwAMP 
% Image-text retrieval results (R@1) on MS-COCO. %(Report R@1 only??)
% }
% %\vspace{+0.3em}
% \centering
% \vspace{-0.6em}
% %\vskip 0.05in
% %\begin{scriptsize}
% %\begin{footnotesize}
% \begin{small}
% %\begin{sc}
% \centering
% %\scalebox{0.95}{
% \begin{tabular}{lcccc}
% \toprule
% \multirow{2}{*}{Methods} & \multicolumn{2}{c}{1K test} & \multicolumn{2}{c}{5K test}
% \\ \cmidrule(lr){2-3} \cmidrule(lr){4-5} 
% & I $\to$ T &  T $\to$ I & I $\to$ T &  T $\to$ I \\
% \hline
% %
% DPC\Tstrut%~\citep{dpc} 
% & 65.6 & 47.1 & 41.2 & 25.3 \\
% %
% VSE++%~\citep{vsepp} 
% & 64.6 & 52.0 & 41.3 & 30.3 \\
% %
% %GXN~\citep{gxn} & 68.5 & 56.6 & 42.0 & 31.7 \\
% %
% SCO%~\citep{sco} 
% & 69.9 & 56.7 & 42.8 & 33.1 \\
% %
% PCME%~\citep{pcme} 
% & 68.8 & 54.6 & 44.2 & 31.9 \\
% %
% \hline
% %
% SCAN i-t\Tstrut & 69.2 & 54.4 & 46.4 & 34.4 \\
% %
% SCAN t-i + i-t & 72.7 & 58.8 & 50.4 & 38.6 \\
% \hline
% %
% Contrastive-PAR\Tstrut & 71.8 & 56.8 & 48.4 & 34.3 \\
% %SwAMP-PAR & 72.6 & 94.6$ & 98.0$ & 57.4 & 87.6 & 94.1 & 49.7 & 79.1 & 88.3 & 35.0 & 65.1 & 76.6 \\
% SwAMP-PAR & ${\bf 72.6}$ & ${\bf 57.4}$ & ${\bf 49.7}$ & ${\bf 35.0}$ \\
% %
% \bottomrule
% \end{tabular}
% %}
% %\end{sc}
% %\end{footnotesize}
% %\end{scriptsize}
% \end{small}
% \label{tab:coco}
% %\vspace{-1.5em}
% \end{table}
% %%%%

%%%%%%%%%%%%%%%%%%%%%%%%%%%%%%%%%%%%%%%%%%%%%%%%%%%%%%%%%%%%%%%%%%%%%%%%%%%%%%%
\subsection{Image-Text Retrieval}\label{sec:img_txt}

For the image-text cross-modal retrieval task, we follow the features and protocols from the well-known {\em stacked cross attention network} (SCAN)~\citep{scan}. In their framework, each image is represented by a set of local features $V=\{v_1,\dots,v_k\}$, where $v_i\ (\in\mathbb{R}^D) = W_v f_i + b_v$ and $f_i$'s are the CNN features extracted from salient image regions detected by the Faster-R-CNN model~\citep{faster_rcnn}. The raw features $f_i$'s are fixed and $\{W_v, b_v\}$ are learnable parameters. 
The text (sentence) is also treated as a set of word features $E=\{e_1,\dots,e_n\}$, where $e_i\ (\in\mathbb{R}^D) = (h^{lr}_i + h^{rl}_i)/2$ and $h^{lr/rl}_i$ are the outputs of the bi-directional GRU~\citep{gru,bidir_rnn} with the sequence of word embeddings as input. Both the word embeddings and GRU parameters are learnable. These image/text features contain rich local information, however, one challenge is that both representations are {\em sets}, hence the number of elements ($k$ and $n$) can vary from instance to instance. 

In~\citep{scan}, they proposed a cross-modal attention model, where each local feature from one modality is transformed by the attention~\citep{transformer} with the set of local features in the other modality; e.g., $v_i$ is transformed to $attn(v_i; \{e_j\}_{j=1}^n) = $ the weighted sum of {\em values} $\{e_j\}_{j=1}^n$ with $v_i$ as a {\em query} and $\{e_j\}_{j=1}^n$ as {\em keys} (this denoted by i-t, while the other attention direction t-i can be used alternatively). Then the similarity score between image $V$ and text $E$ is defined as $pool(\{cos(v_i,attn(v_i; \{e_j\}_{j=1}^n))\}_{i=1}^K)$, where $cos(a,b)$ is the cosine similarity and $pool$ is the pooling operation, either of $AVG$ (average) or $LSE$ (log-sum-exp). Then the triplet contrastive loss of (\ref{eq:contrastive_loss}) is employed. %For the details, please refer to~\citep{scan}.
%
%Note that in the SCAN, there is no succinct modality-wise embedding vector representation, but 
%That is, in the SCAN, 
Although the cross-attention %means that the similarity score between instances of two modalities is computed by highly complex attention operations. Although this 
is useful for capturing interaction between local features, computing the similarity score takes quadratic time in the number of local features in the instances. This is time consuming compared to the simple dot-product of the modality-wise embedding vectors (See Table~\ref{tab:image_text_time} for wall-clock times). % for the actual running times). % compared with the approaches based on modality-wise feature representation).
%Moreover, it is not applicable to our SwAMP approach since we need to predict the class labels for each modality from modality-wise representation $\phi^{image}(V)$, $\phi^{text}(E)$. 

To have modality-wise succinct representation instead (for SwAMP), we adopt the {\em induced-set attention} idea from Set-Transformer~\citep{set_transformer}. Specifically, we introduce $p$ learnable prototype (query) vectors $\{q_j\}_{j=1}^p$, $q_j\in \mathbb{R}^D$. Then we compute the attention for each query with $V$ (or $E$), i.e., $z_j = attn(q_j; \{v_i\}_{i=1}^k)$. We define $\phi^{image}(V) = concat(z_1,\dots,z_p)$, similarly for $\phi^{text}(E)$, where $concat$ refers to concatenation. 
%Thus the parameters for $\phi^{image}()$ are $\{W_v, b_v\}$ and $\{q_j\}_{j=1}^p$, and the parameters for $\phi^{text}()$ are the word embeddings, GRU parameters, and $\{q_j\}_{j=1}^p$. 
We share the same $\{q_j\}_{j=1}^p$ for both modalities. We also have multi-head extension. % by computing these features multiple times and concatenating them. 
%the simple prototype-based attention module, where we essentially maintain a fixed number (denoted by $p$) of prototype vectors (of the same dimensionality as that of the embedding), then we use the prototype vectors as queries and set data $x^A$ or $x^B$ as keys/values in the attention. The transformed queries are concatenated as a ($p \cdot d$-dim vector, which is the succinct vector representation for a set instance. 
We call these modality-wise features as {\em prototype attention representation} (PAR). Note that computing PAR features has linear complexity in the number of local features ($p$ assumed constant), and the cross-modal similarity is simply dot-product of PAR features, and can be computed in linear time (See also Table~\ref{tab:image_text_time} for comparison with SCAN's cross-modal attention).

% %%%%
% \begin{itemize}
% %
% \item SCAN
% %
% \item CosSim-PAR: Contrastive loss with PAR vector representation of a image/text set instance. 
% %
% \item CosSim-ST: Contrastive loss with Set-Transformer representation of a image/text set instance. 
% %
% \item SwAMP-PAR: SwAMP loss with PAR
% %
% \item SwAMP-ST: SwAMP loss with ST
% %
% \end{itemize}
% %%%%

%%%%%%%%%%%%%%
%\subsubsection{Datasets and Results}

We test our approach on the popular image-text retrieval datasets, MS-COCO and Flickr30K. 
The details of the datasets and training/test protocols are described in Appendix (Sec.~D). 
%There are 31K images and five captions for each image in Flickr30K. MS-COCO contains $123,287$ images, where each image is annotated with five text descriptions. Following the widely-used split~\citep{karpathy,vsepp}, for the Flickr30K, we have 1K images for validation, 1K images for testing, and the rest for training. For MS-COCO, there are 5K test images (and 25K captions, five captions for each image). We also follow two standard protocols for measuring the test retrieval performance for MS-COCO: 1) using the entire 5K test images or 2) splitting the test set into 5 folds and report the average retrieval performance over the 5 folds. 
The results are summarized in Table~\ref{tab:flickr} %(Flickr) and
and Table~\ref{tab:coco}. % (MS-COCO). % for R@1 (R@5 and R@10 reported in Supplement). 
We specifically highlight the comparison between the contrastive loss and our SwAMP loss with the modality-wise feature representation (Contrastive-PAR vs.~SwAMP-PAR). For the PAR features, we choose the number of prototypes $p=20$, attention weight temperature $T=0.5$, and the number of heads $H=1$ for Flickr, and $p=10, T=0.5, H=2$ for MS-COCO. For the SwAMP hyperparameters, we use the impact of SwAMP loss $\lambda=1.0$, softmax temperature $\tau=0.025$, the number of classes $K=1,000$, queue size $1,280$ for both datasets. 
%As shown, the SwAMP loss 
SwAMP performs consistently better than the contrastive loss and outperforms 
several state-of-the-arts including the recent sophisticated probabilistic embedding strategy (PCME)~\citep{pcme}.

%%%% 
\begin{table}[t!]
\vspace{-0.7em}
\caption{Running times (seconds) % of SCAN %(cross-modal attention) 
%and SwAMP-PAR, 
measured on (Core i7 3.50GHz CPU / 128GB RAM / 1 %GeForce 
RTX-2080Ti GPU). %We report per-batch times for training, and entire retrieval times for test. 
Per-batch times for training, entire times for test. 
For MS-COCO test, %entire times for 
times for 5K test images (1K test in parentheses). 
%the running times for 5K test images are reported, where times for 1K test images averaged over 5 folds are shown in the parentheses. 
%For SCAN, when we use features in both directions (e.g., t-i AVG + i-t LSE), the running times are roughly doubled.
}
\vspace{+0.3em}
\centering
%\vspace{-1.0em}
%\vskip 0.05in
\begin{scriptsize}
%\begin{footnotesize}
%\begin{small}
%\begin{sc}
\centering
%\scalebox{0.95}{
\begin{tabular}{lcccc}
\toprule
\multirow{2}{*}{Methods} & 
\multicolumn{2}{c}{\ \ \ \ Flickr30K \ \ \ \ } &  \multicolumn{2}{c}{MS-COCO}
\\ \cmidrule(lr){2-3} \cmidrule(lr){4-5} 
& Train & Test & Train & Test \\
\hline
SCAN i-t AVG\Tstrut & 
0.35 %0.3470 
& 336.9 %336.9181 
& 0.33 %0.3280 
& 9352.0 (350.3)
%9351.9934 (350.2627) 
\\
\hline
SwAMP-PAR\Tstrut & 
0.09 %0.0968 
& 3.8 %3.7806 
& 0.08 %0.0819 
& 25.9 (16.3)
%25.8951 (16.2914) 
\\
\bottomrule
\end{tabular}
%}
%\end{sc}
%\end{footnotesize}
\end{scriptsize}
%\end{small}
\label{tab:image_text_time}
%\vspace{-1.5em}
\end{table}
%%%%

When compared with the computationally expensive SCAN, SwAMP mostly outperforms SCAN except for the SCAN's best attention direction/combination choices. 
%Note that SwAMP uses the simple feature aggregation strategy (PAR) to have fast and succinct modality-wise feature representation, whereas SCAN relies on the cross-modal attention similarity scoring model, which is computationally expensive. 
To see the computational advantage of SwAMP-PAR, we compare the actual training/test times for the two approaches in Table~\ref{tab:image_text_time}, measured on the same machine with a single GPU (RTX 2080 Ti) and Core i7 3.50GHz CPU. %, and 128 GB RAM. 
Our SwAMP-PAR is about 4 times faster than SCAN for training on both datasets, while the difference becomes even more pronounced during test; SwAMP-PAR is about two orders of magnitude faster than the cross-modal attention model.

%\hl{Also add a few retrieval examples/figures...}

%Hyperparameters: We used the SwAMP loss alone, but it turned out it didn't work well (because of the chicken-egg problem, ie, initial embedding is inaccurate, and the class prediction using the inaccurate embedding is inaccurate, so vicious cycle). Instead, we combine the SwAMP and CosSim losses together with the balancing parameters $\lambda = 1.0$ for the impact of the CosSim-loss. 

% \hl{May put the following benefit over cross-modal attention?}
% %%%%
% \begin{enumerate}
% %
% \item Benefit-1: SwAMP loss requires a linear sampling complexity (unlike quadratic for the contrastive cossim loss), which results in faster convergence during training (??), and makes \hl{the SwAMP loss lead to better retrieval performance than the contrastive cossim loss.}
% %
% \item Benefit-2: On image-text retrieval benchmarks, the SwAMP loss with a simple feature aggregation leads to retrieval performance (sometimes better) comparable to that of the complex cross-modal attention similarity scoring model (SCAN). That is, \hl{the SwAMP loss with simple modality-wise feature representation has retrieval performance comparable to the complex SCAN's joint similarity scoring model, with significantly faster test retrieval time (about two orders of magnitude).}
% %
% \end{enumerate}
% %%%%

%%%%%%%%%%%%%%%%%%%%%%%%%%%%%%%%%%%%%%%%%%%%%%%%%%%%%%%%%%%%%%%%%%%%%%%%%%%%%%%
\subsection{Ablation Study}\label{sec:ablation}

%In our experiments, we chose the hyperparameters by cross validation with grid search. 
We perform empirical study on the impact of two important hyperparameters in our model: the number of classes $K$ and SwAMP loss trade-off $\lambda$. 

%%%%
\begin{figure}[t!]
%\vspace{-1.3em}
\begin{center}
%
%\begin{subfigure}[b]{0.9\textwidth}
\centering
\includegraphics[trim = 10mm 5mm 7mm 4mm, clip, scale=0.243]{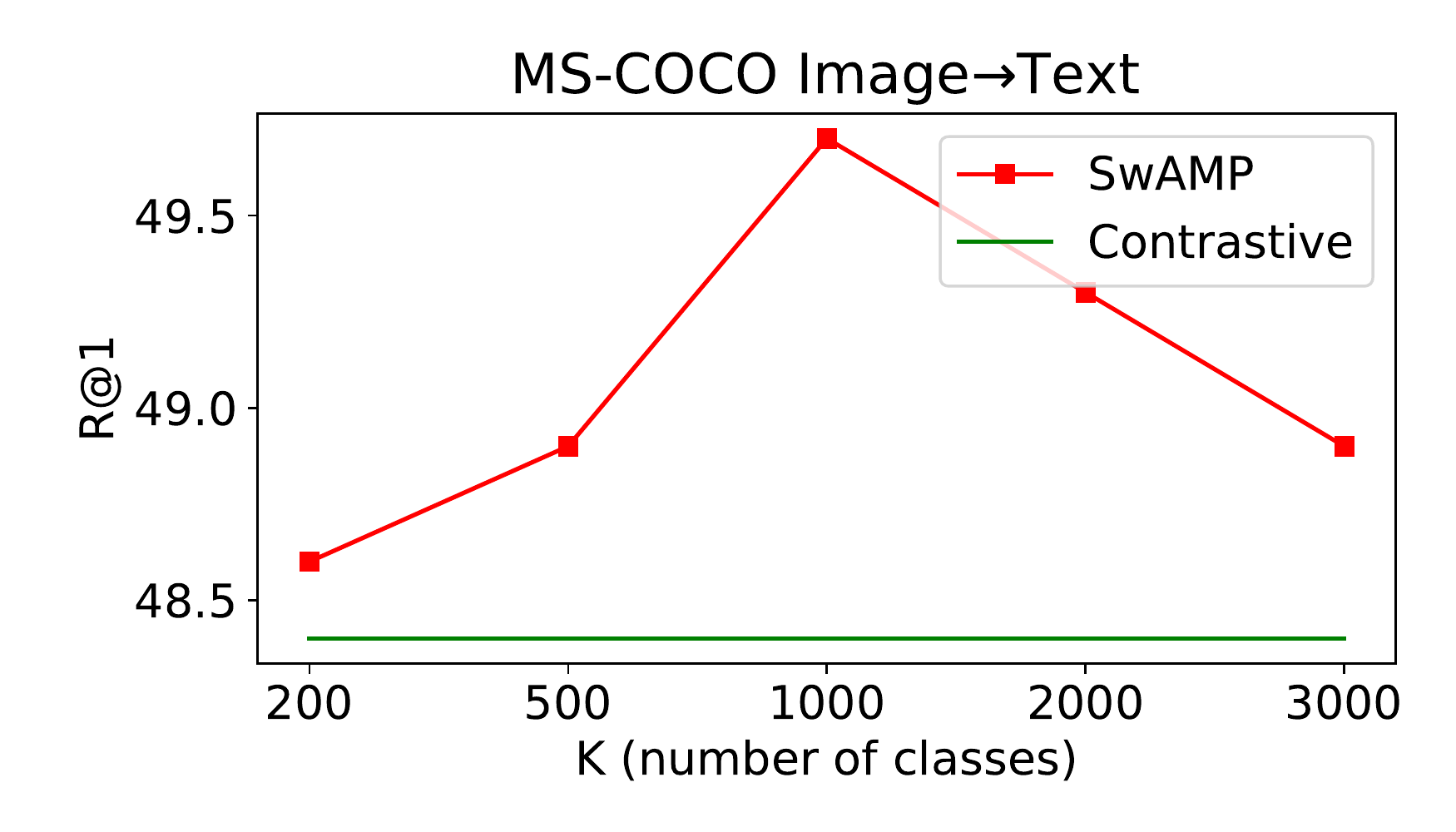} \  \includegraphics[trim = 10mm 5mm 7mm 4mm, clip, scale=0.243]{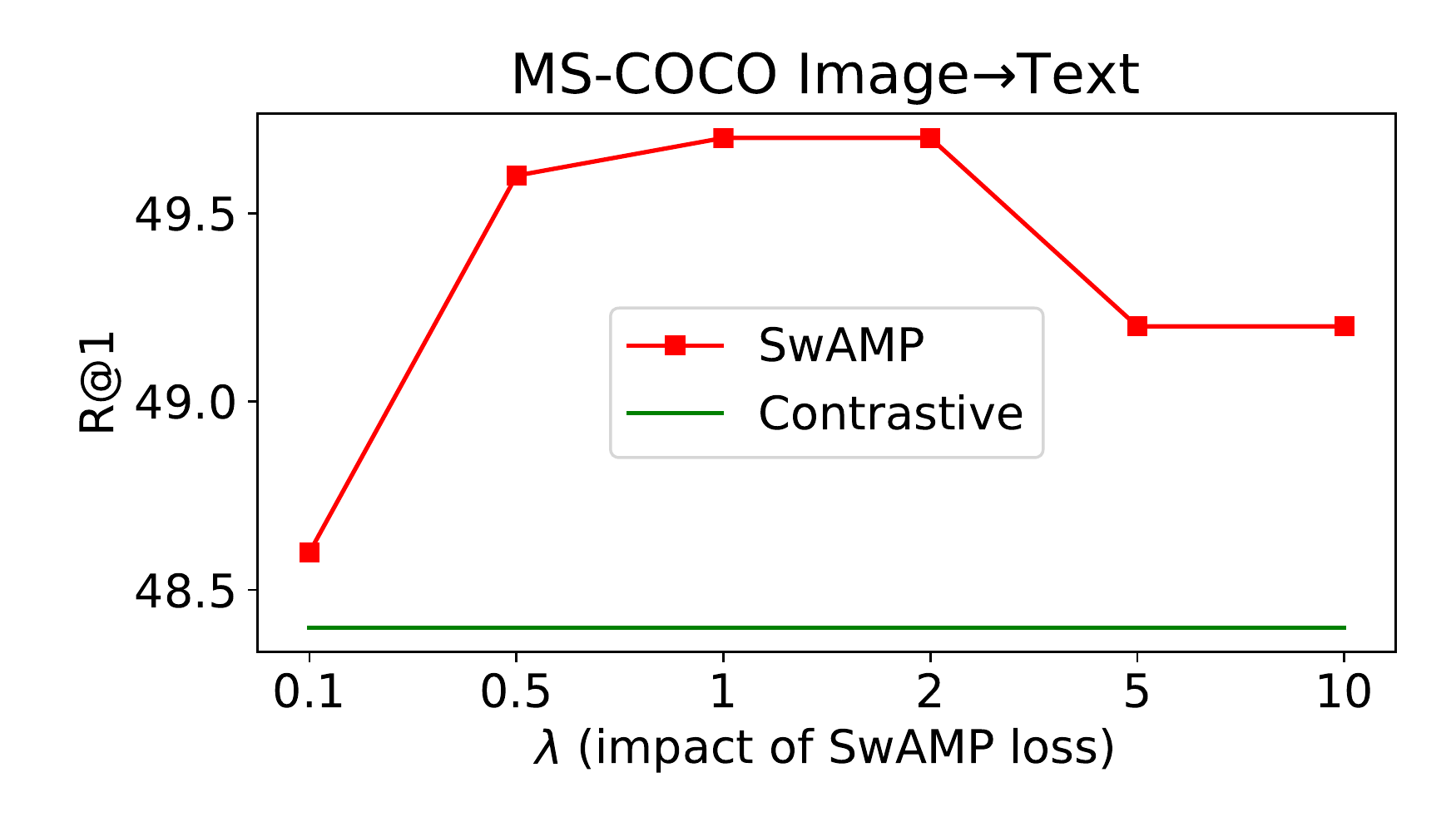}
\end{center}
\vspace{-1.3em}
\caption{Impact of $K$ (the number of classes) and $\lambda$. 
}
\vspace{-1.0em}
\label{fig:impact_of_K}
\end{figure}
%%%%

\textbf{Number of classes ($K$).} 
Recall that the best $K$ values we chose were: $K=1000$ for the image-text retrieval datasets and $K=500$ for text-based video retrieval. To see how the retrieval performance is affected by other choices of $K$, we conduct experiments by varying $K$ around the optimal values. The results on MS-COCO ($I\to T$) and YouCook2 tasks are shown in Fig.~\ref{fig:impact_of_K} (Left). (More results on other datasets can be found in Appendix (Fig.~4--8, Sec.~B).) 
%from $\{200,500,1000,2000,3000\}$, and record the R@1 scores for both pair and class based error types for our SwAMP model. 
Clearly, very small $K$ has low retrieval performance (R@1), and increasing $K$ leads to improvement. However, beyond certain points, there is no benefit of increasing $K$ and we even see performance degradation, which agrees with the observations from previous work~\citep{sela,swav}. This is perhaps due to the difficulty of assigning meaningful cluster labels in optimal transport. Overall, with properly chosen $K$, SwAMP outperforms contrastive learning, signifying that SwAMP's grouping/clustering of similar instances is more effective than vanilla instance discrimination. The fact that the optimal $K$ values are different in two tasks (image-text and video-text) implies that the best cardinality of semantic clusters is highly dependent on the dataset characteristics (e.g., size and semantic diversity). 

%%%%

% %%%%
% \begin{figure}[t!]
% %\vspace{-1.5em}
% \begin{center}
% %
% %\begin{subfigure}[b]{0.9\textwidth}
% \centering
% \includegraphics[trim = 5mm 2mm 5mm 4mm, clip, scale=0.285]{figs/coco_impact_of_lambda_i2t.pdf} \ \  
% % \includegraphics[trim = 5mm 2mm 5mm 4mm, clip, scale=0.285]{figs/msrvtt_impact_of_lambda.pdf}
% %
% \end{center}
% \vspace{-1.5em}
% \caption{Sensitivity to the SwAMP loss trade-off ($\lambda$). 
% }
% \vspace{-1.5em}
% \label{fig:impact_of_lambda}
% \end{figure}
% %%%%

\textbf{SwAMP impact ($\lambda$).} 
%We perform sensitivity analysis on $\lambda$, the strength of the SwAMP loss in (\ref{eq:loss_comb}). 
The sensitivity to $\lambda$ is shown in Fig.~\ref{fig:impact_of_K} (Right), and more results and further discussions are %results 
in Appendix~(Fig.~9--13, Sec.~B).

\begin{figure}[t!]
%\vspace{-1.0em}
\begin{center}
%
%\begin{subfigure}[b]{0.9\textwidth}
\centering
\includegraphics[trim = 5mm 3mm 3mm 4mm, clip, scale=0.292]{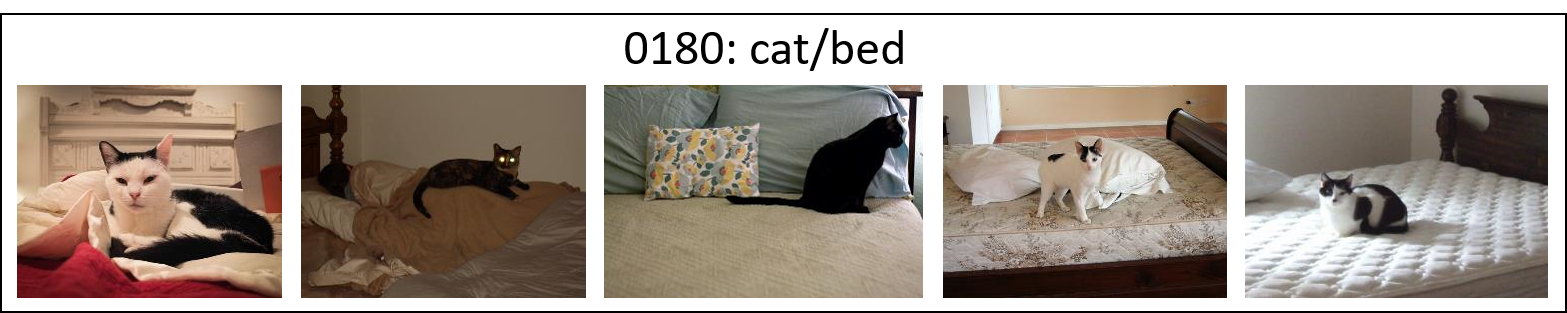} \ \ 
\includegraphics[trim = 5mm 5mm 3mm 5mm, clip, scale=0.266]{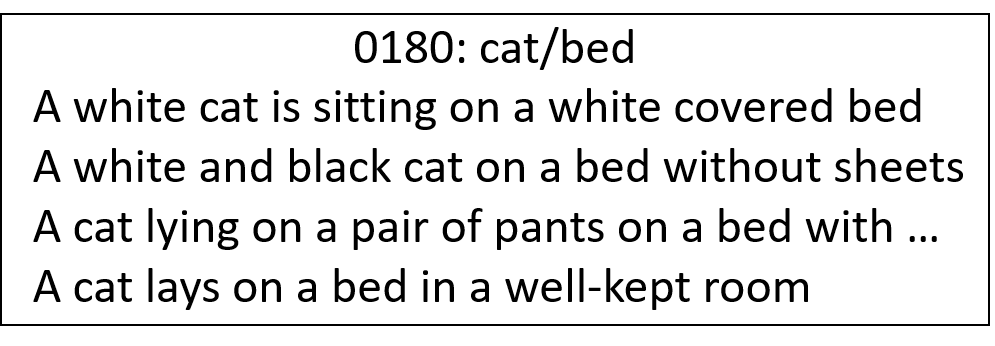} \\ \vspace{+0.5em}
\includegraphics[trim = 5mm 3mm 3mm 4mm, clip, scale=0.292]{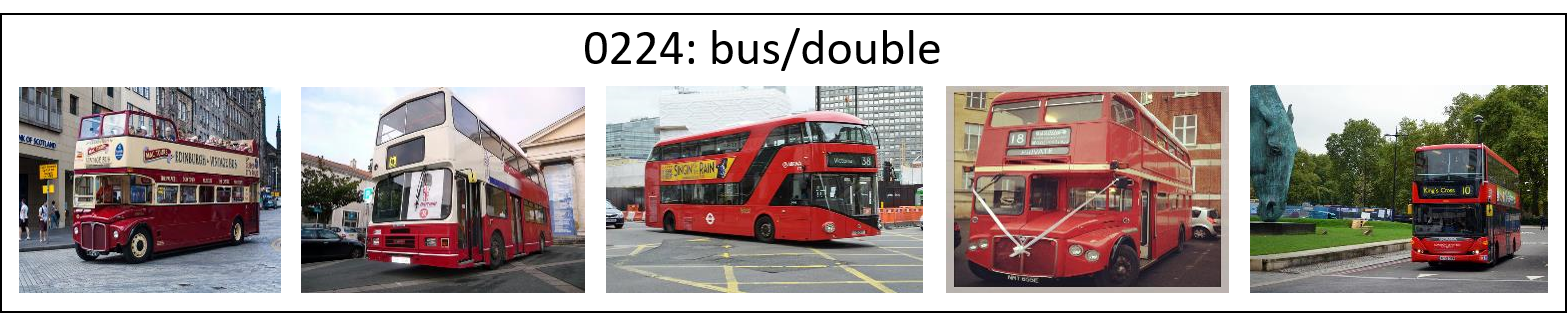} \ \ 
\includegraphics[trim = 5mm 5mm 3mm 5mm, clip, scale=0.266]{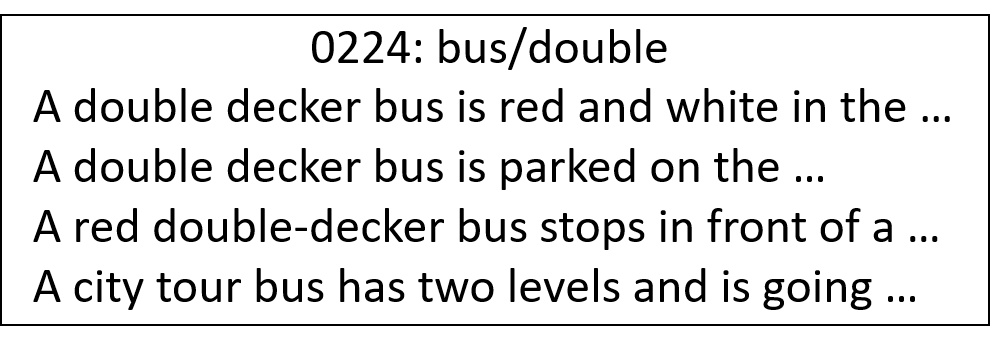} \\ \vspace{+0.5em}
\includegraphics[trim = 5mm 3mm 3mm 4mm, clip, scale=0.292]{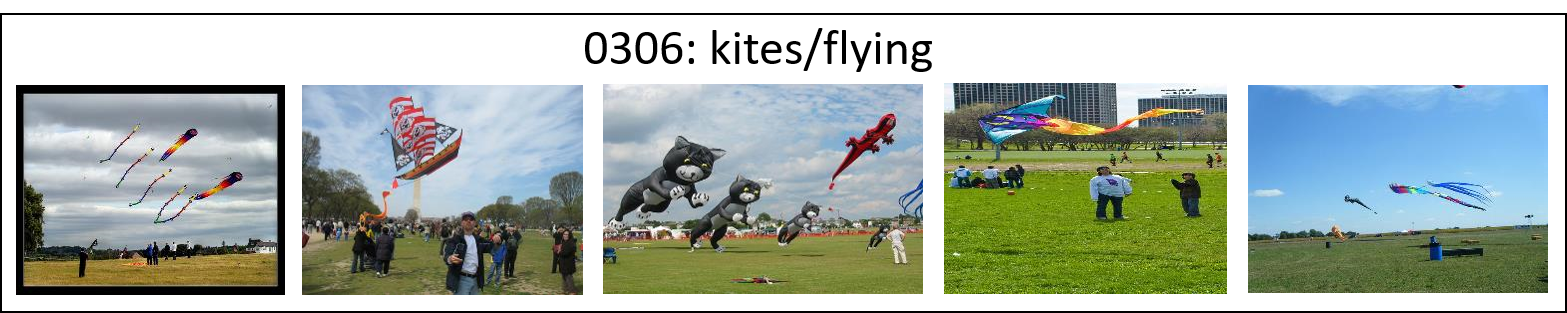} \ \ 
\includegraphics[trim = 5mm 5mm 3mm 5mm, clip, scale=0.266]{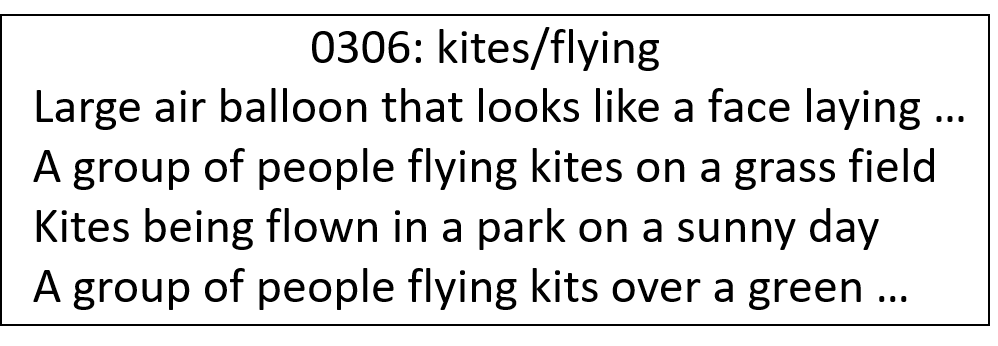} \\ \vspace{+0.5em}
\includegraphics[trim = 5mm 3mm 3mm 4mm, clip, scale=0.292]{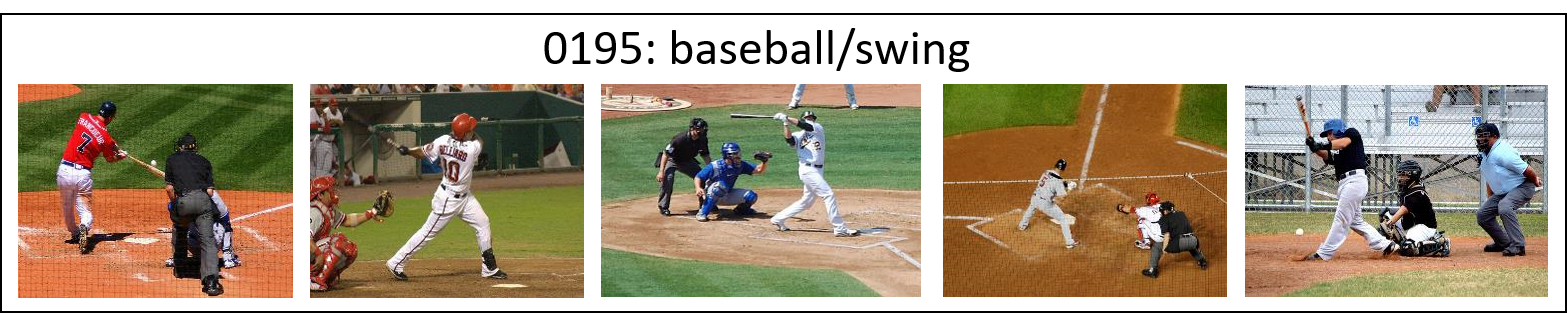} \ \ 
\includegraphics[trim = 5mm 5mm 3mm 5mm, clip, scale=0.266]{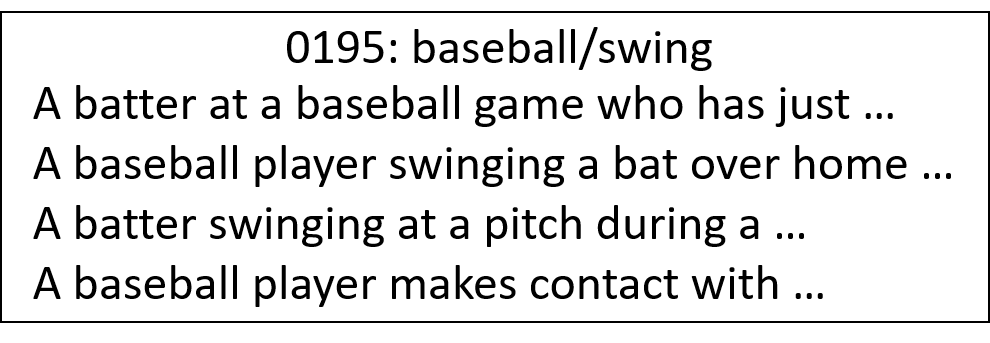} \\ \vspace{+0.5em}
\includegraphics[trim = 5mm 3mm 3mm 4mm, clip, scale=0.292]{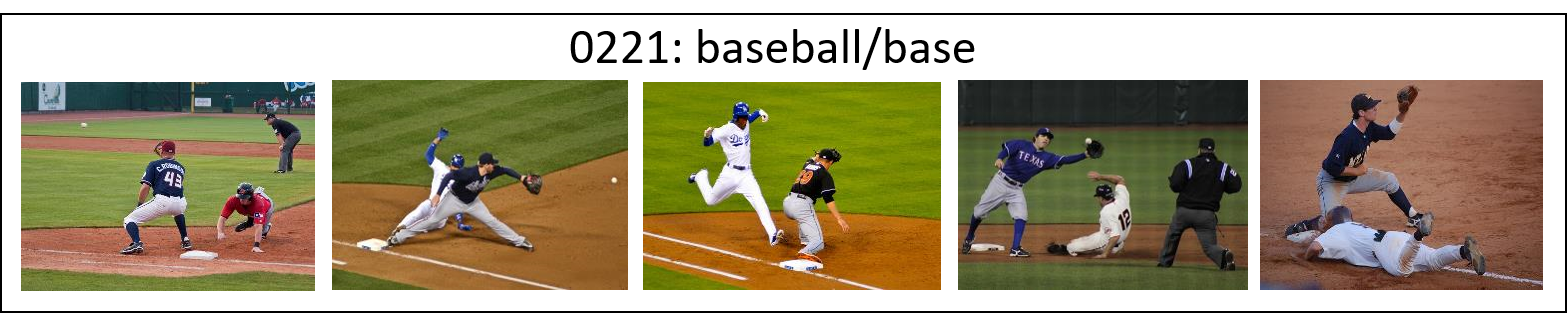} \ \ 
\includegraphics[trim = 5mm 5mm 3mm 5mm, clip, scale=0.266]{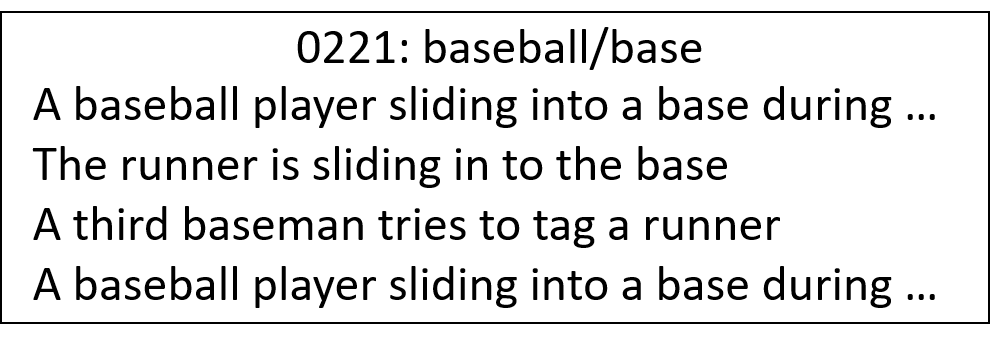} \\ \vspace{+0.5em}
\includegraphics[trim = 5mm 3mm 3mm 4mm, clip, scale=0.292]{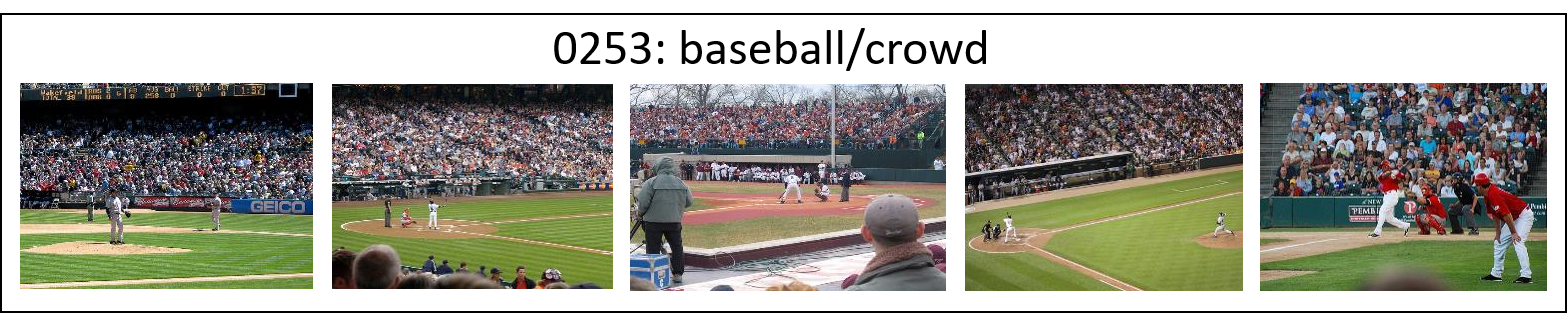} \ \ 
\includegraphics[trim = 5mm 5mm 3mm 5mm, clip, scale=0.266]{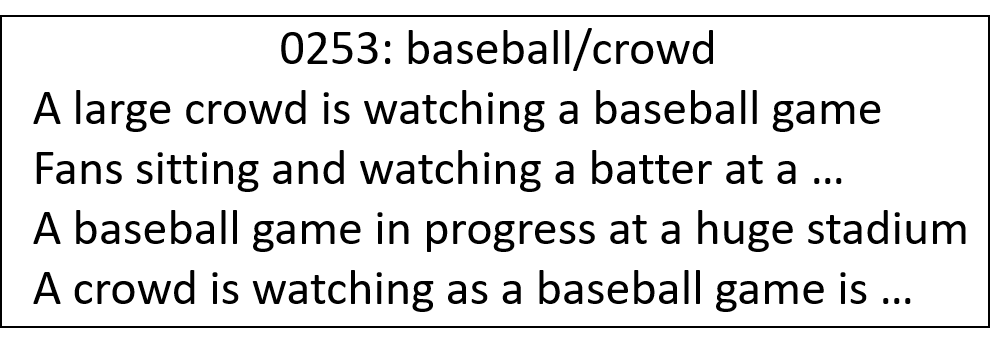}
\end{center}
\vspace{-1.5em}
\caption{%For MS-COCO trained model. 
Some randomly selected clusters with images and texts that belong to them. Each cluster, titled by {\em ID: keywords}, shows randomly chosen 5 images and 4 texts. 
}
\vspace{-0.5em}
\label{fig:clusters_examples}
\end{figure}
%%%%

%%%%%%%%%%%%%%%%%%%%%%%%%%%%%%%%%%%%%%%%%%%%%%%%%%%%%%%%%%%%%%%%%%%%%%%%%%%%%%%
\subsection{Visualization of Learned Clusters}\label{sec:qualitative}

As qualitative analysis, we visualize the learned clusters to see if they capture meaningful semantic information. On MS-COCO (trained with the number of classes $K=1000$), we organize images and texts by their assigned cluster labels using the learned prototype classification model (i.e., (\ref{eq:class_predictive})). 
We first visually inspect individual clusters, images and texts that belong to each cluster. As we show a few examples in Fig.~\ref{fig:clusters_examples} (more in Appendix (Fig.~2,3, Sec.~A)), each cluster contains semantically coherent data samples. Then we inspect texts (captions) in each cluster, and select a few keywords, those words that appear the most frequently in the texts. These keywords for each cluster consist of objects (noun) and/or actions (verb) that faithfully describe the cluster and data samples that belong to it. The full list is shown in Appendix (Fig.~1, Sec.~A), but to enumerate a few of them ({\em cluster ID: keywords}), for instance, 
{\em 0014: giraffe/feeding},
{\em 0169: soccer/playing}, 
{\em 0283: bus/parked}, 
%{\em 0385: bed/laying}, 
{\em 0405: pizza/oven}, 
{\em 0597: vase/flowers}, 
%{\em 0604: train/traveling}, 
{\em 0713: dog/ball}, 
{\em 0818: kite/flying}, 
{\em 0956: parking/meter}.
% 0001: airplane/fly
% 0014: giraffe/feeding
% 0141: cake/cutting
% 0169: soccer/playing
% 0276: table/eating
% 0283: bus/parked
% 0340: bus/double/decker
% 0385: bed/laying
% 0405: pizza/oven
% 0491: cat/chair
% 0550: bird/perched
% 0597: vase/flowers
% 0604: train/traveling
% 0713: dog/ball
% 0818: kite/flying
% 0956: parking/meter

Although the last three clusters in Fig.~\ref{fig:clusters_examples} all have the semantic meaning of {\em baseball}, they have different details in either activity or focus/scene: {\em swing}, {\em base playing}, and {\em crowd scene}. This means that SwAMP finds clusters based on the whole contents (objects, acitivities, and scenes), % from both modalities, 
instead of doing merely object-based clustering. 
%However, even though the semantics are similar there are subtle detailed separations. (baseball, pizza), hierarchical clustering. 
Although we have roughly equal numbers of samples per cluster, we found that some clusters are overlapped with others in terms of semantic meaning (redundant clusters in Appendix (Fig.~1, Sec.~A)), constituting larger {\em super}-clusters. These clusters are related to dominant data samples (e.g., cat, dog, tennis, baseball). This implies that the SwAMP can effectively deal with imbalance of semantic classes that can reside in data. 

% %%%%
% \begin{figure}[t!]
% %\vspace{-1.5em}
% \begin{center}
% %
% %\begin{subfigure}[b]{0.9\textwidth}
% \centering
% \includegraphics[trim = 5mm 6mm 5mm 2mm, clip, scale=0.292]{figs/vis/clusters_keywords.png}
% %
% \end{center}
% \vspace{-2.3em}
% \caption{1000 clusters trained with MS-COCO. For each cluster, we depict a few most frequent keywords in the captions that belong to the cluster.
% }
% %\vspace{-1.0em}
% \label{fig:clusters_keywords}
% \end{figure}
% %%%%

%%%%%%%%%%%%%%%%%%%%%%%%%%%%%%%%%%%%%%%%%%%%%%%%%%%%%%%%%%%%%%%%%%%%%%%%%%%%%%%
%%%%%%%%%%%%%%%%%%%%%%%%%%%%%%%%%%%%%%%%%%%%%%%%%%%%%%%%%%%%%%%%%%%%%%%%%%%%%%%
\section{Conclusion}\label{sec:conclusion}

We have proposed a novel clustering-based loss function for cross-modal retrieval. The swapped class assignment over the modalities enables improved feature alignment with increased flexibility, while discovering meaningful latent semantic classes. %, while it helps reducing the sampling complexity significantly. 
The efficacy of our approach was demonstrated on several real-world cross-modal retrieval problems in diverse modalities, text-video, sketch-photo, and image-text, where our method achieved significant performance improvement over the contrastive learning for all these tasks.

%\clearpage
\bibliographystyle{plainnat}
\bibliography{main}

\clearpage
\appendix

\onecolumn
\aistatstitle{Appendix
}

%%%%
\begin{itemize}
\item Visualization of Learned Clusters (Sec.~\ref{appsec:qualitative})
\item Ablation Study (Sec.~\ref{appsec:ablation})
\item (Detailed) Text-based Video Retrieval (Sec.~\ref{appsec:video_text})
%
%\item (Detailed) Sketch-based Image Retrieval (Sec.~\ref{appsec:sketch})
%
\item (Detailed) Image-Text Retrieval (Sec.~\ref{appsec:img_txt})
\item (Extra Experiments) Synthetic Data (Sec.~\ref{appsec:synth})
\vspace{+2.0em}
\end{itemize}
%%%%

%%%%%%%%%%%%%%%%%%%%%%%%%%%%%%%%%%%%%%%%%%%%%%%%%%%%%%%%%%%%%%%%%%%%%%%%%%%%%%%
%%%%%%%%%%%%%%%%%%%%%%%%%%%%%%%%%%%%%%%%%%%%%%%%%%%%%%%%%%%%%%%%%%%%%%%%%%%%%%%
\section{Visualization of Learned Clusters}\label{appsec:qualitative}

\begin{figure}[t!]
%\vspace{-1.5em}
\begin{center}
%
%\begin{subfigure}[b]{0.9\textwidth}
\centering
\includegraphics[trim = 5mm 6mm 5mm 2mm, clip, scale=0.363]{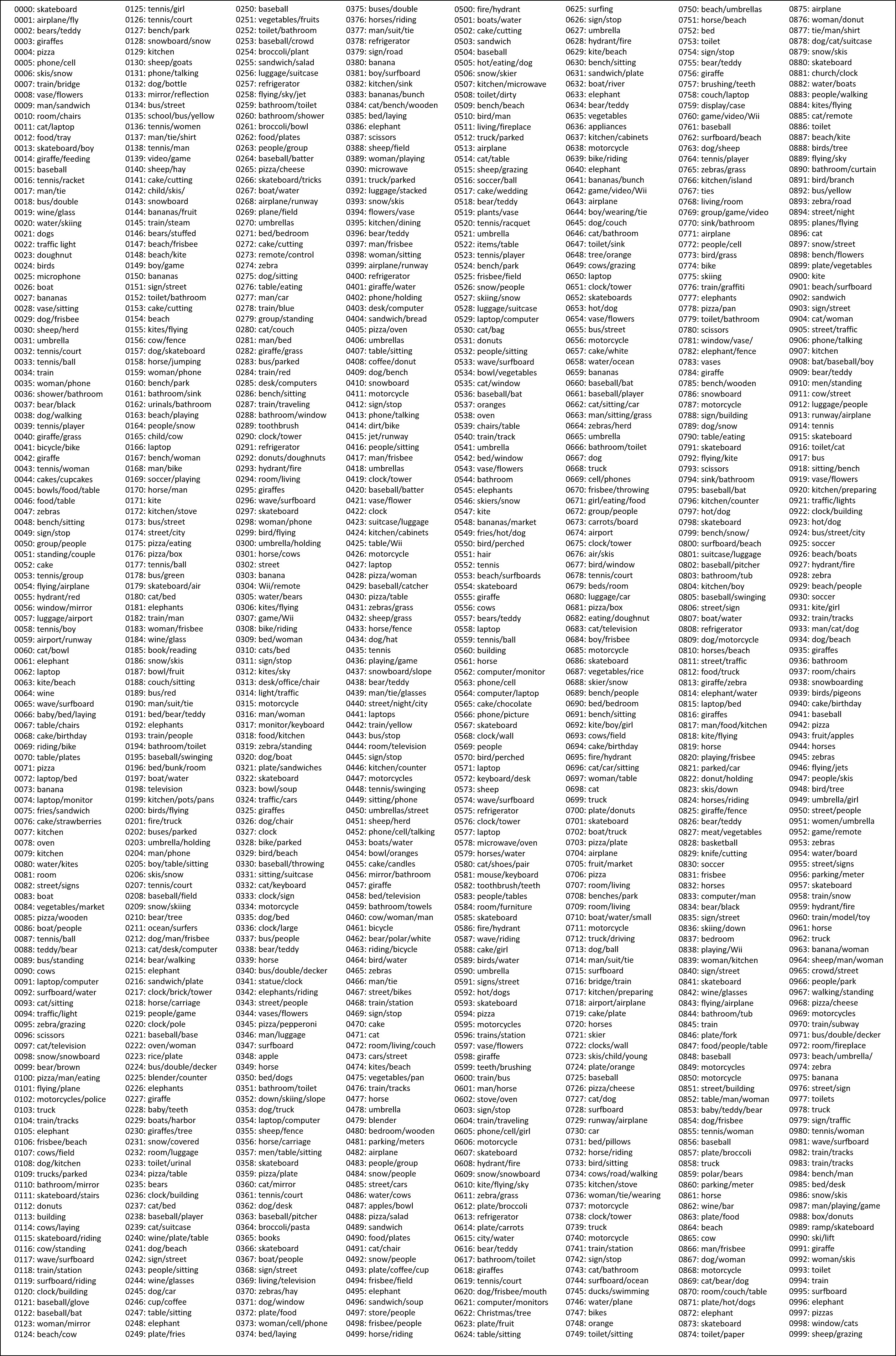}
\end{center}
\vspace{-1.5em}
\caption{1000 clusters trained with MS-COCO. For each cluster, we depict a few most frequent keywords in the captions that belong to the cluster.
}
%\vspace{-1.0em}
\label{appappfig:clusters_keywords}
\end{figure}

\begin{figure}[t!]
\vspace{-0.5em}
\begin{center}
%
%\begin{subfigure}[b]{0.9\textwidth}
\centering
\includegraphics[trim = 5mm 3mm 5mm 4mm, clip, scale=0.342]{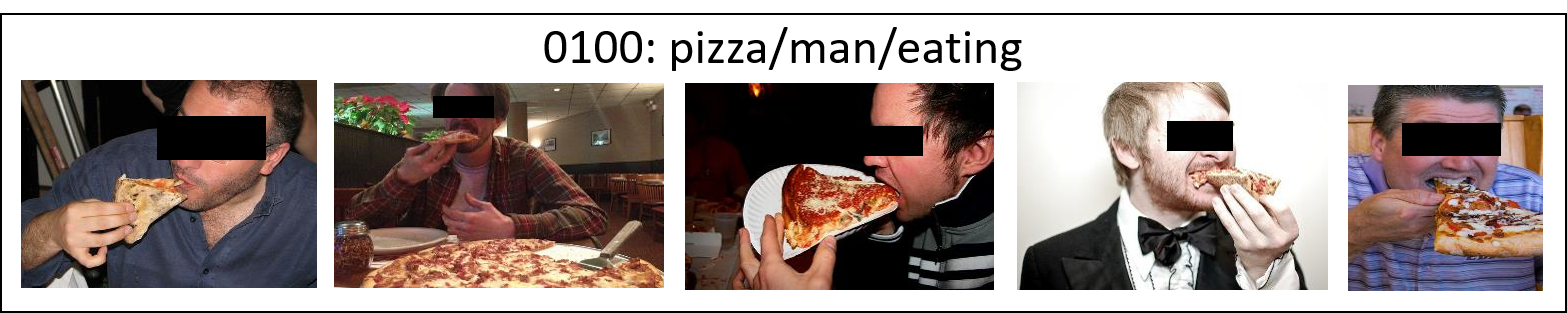} \ \ 
\includegraphics[trim = 5mm 5mm 5mm 5mm, clip, scale=0.312]{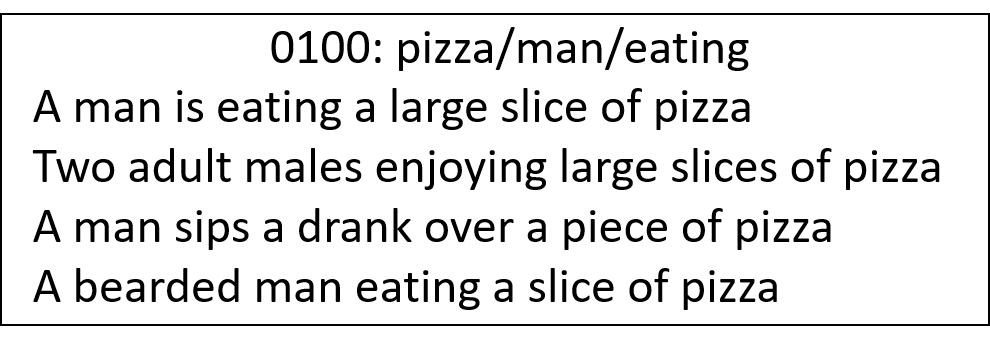} \\ \vspace{+0.2em}
\includegraphics[trim = 5mm 3mm 5mm 4mm, clip, scale=0.342]{figs/vis/0180_cat_bed_img.png} \ \ 
\includegraphics[trim = 5mm 5mm 5mm 5mm, clip, scale=0.312]{figs/vis/0180_cat_bed_txt.png} \\ \vspace{+0.2em}
\includegraphics[trim = 5mm 3mm 5mm 4mm, clip, scale=0.342]{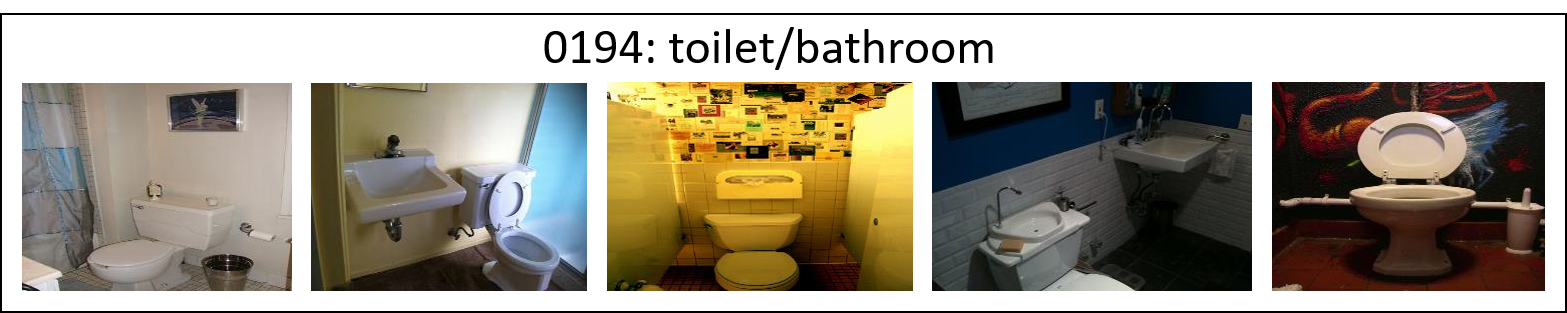} \ \ 
\includegraphics[trim = 5mm 5mm 5mm 5mm, clip, scale=0.312]{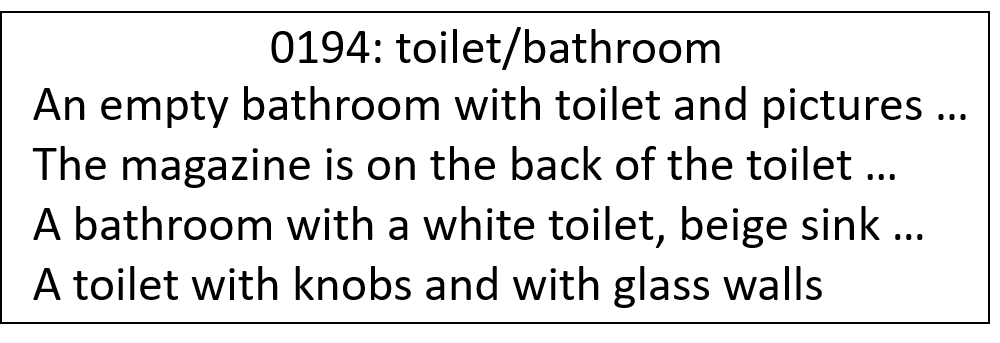} \\ \vspace{+0.2em}
\includegraphics[trim = 5mm 3mm 5mm 4mm, clip, scale=0.342]{figs/vis/0195_baseball_swing_img.png} \ \ 
\includegraphics[trim = 5mm 5mm 5mm 5mm, clip, scale=0.312]{figs/vis/0195_baseball_swing_txt.png} \\ \vspace{+0.2em}
\includegraphics[trim = 5mm 3mm 5mm 4mm, clip, scale=0.342]{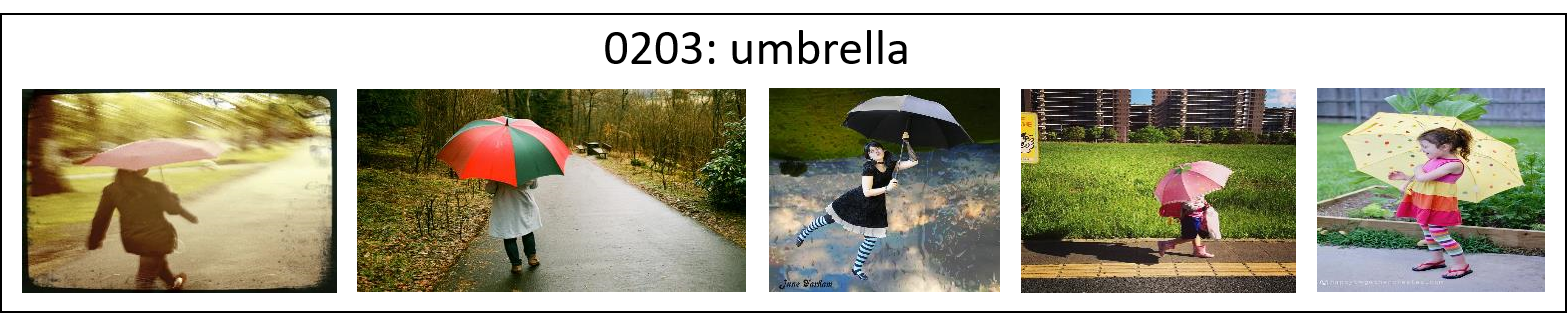} \ \ 
\includegraphics[trim = 5mm 5mm 5mm 5mm, clip, scale=0.312]{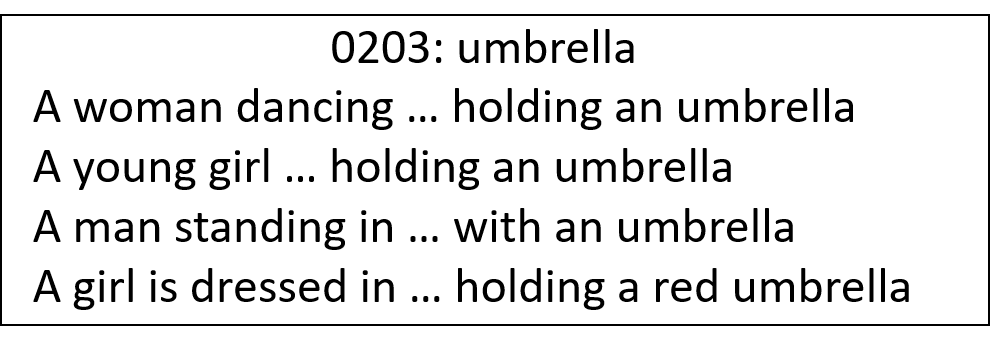} \\ \vspace{+0.2em}
\includegraphics[trim = 5mm 3mm 5mm 4mm, clip, scale=0.342]{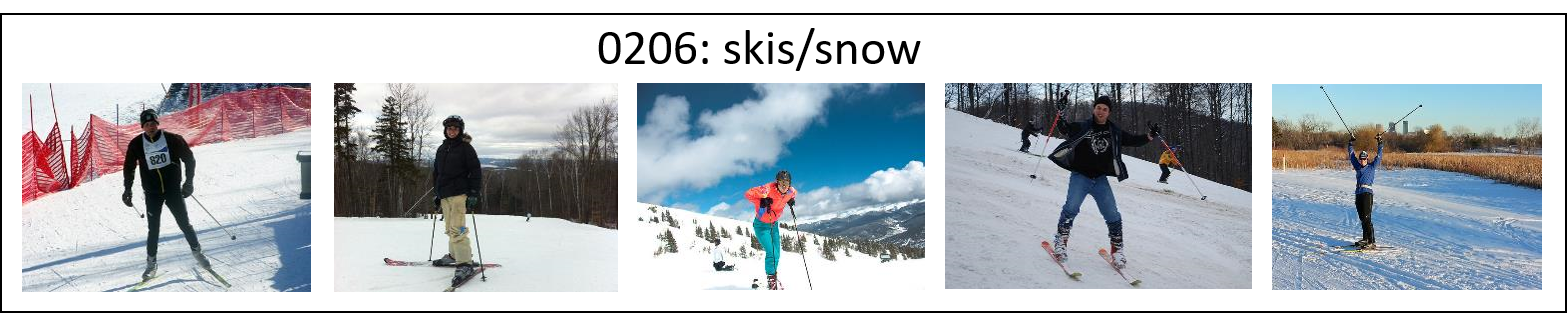} \ \ 
\includegraphics[trim = 5mm 5mm 5mm 5mm, clip, scale=0.312]{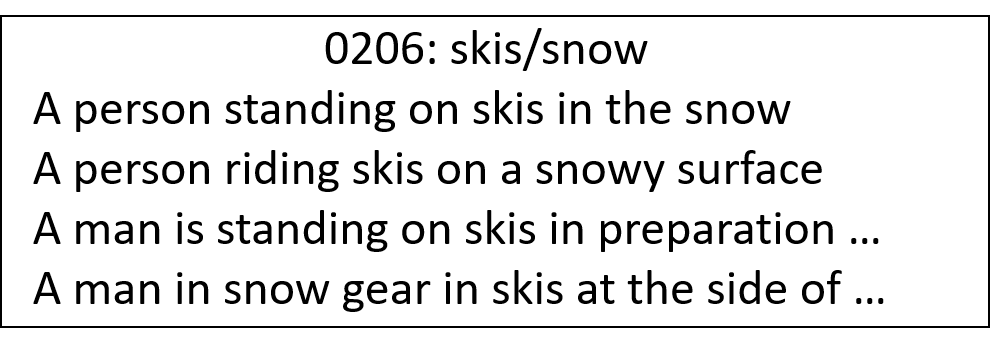} \\ \vspace{+0.2em}
\includegraphics[trim = 5mm 3mm 5mm 4mm, clip, scale=0.342]{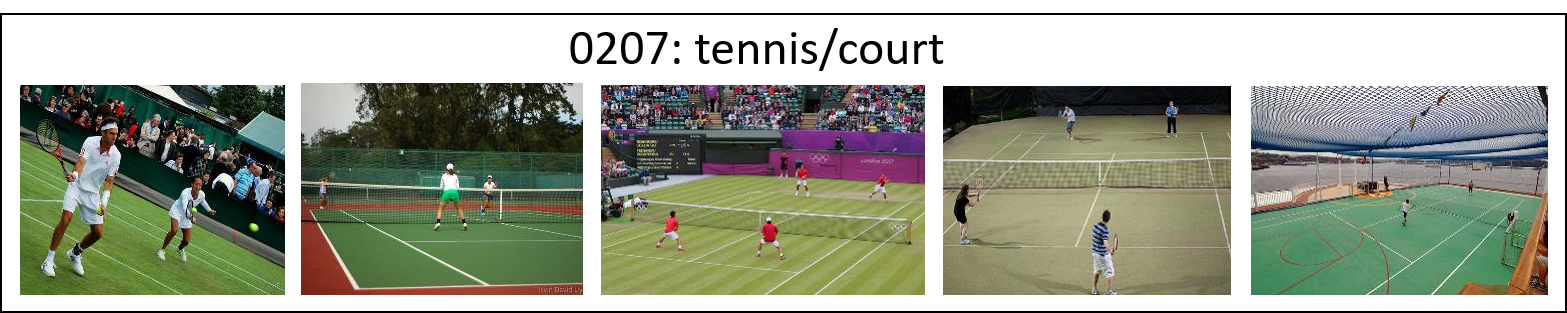} \ \ 
\includegraphics[trim = 5mm 5mm 5mm 5mm, clip, scale=0.312]{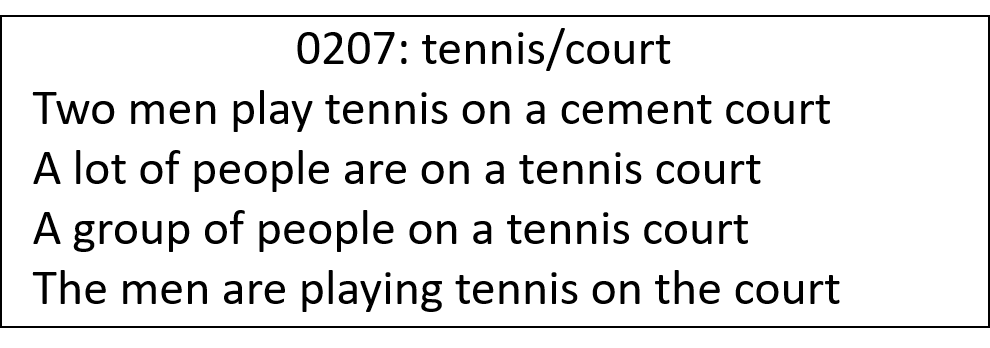} \\ \vspace{+0.2em}
\includegraphics[trim = 5mm 3mm 5mm 4mm, clip, scale=0.342]{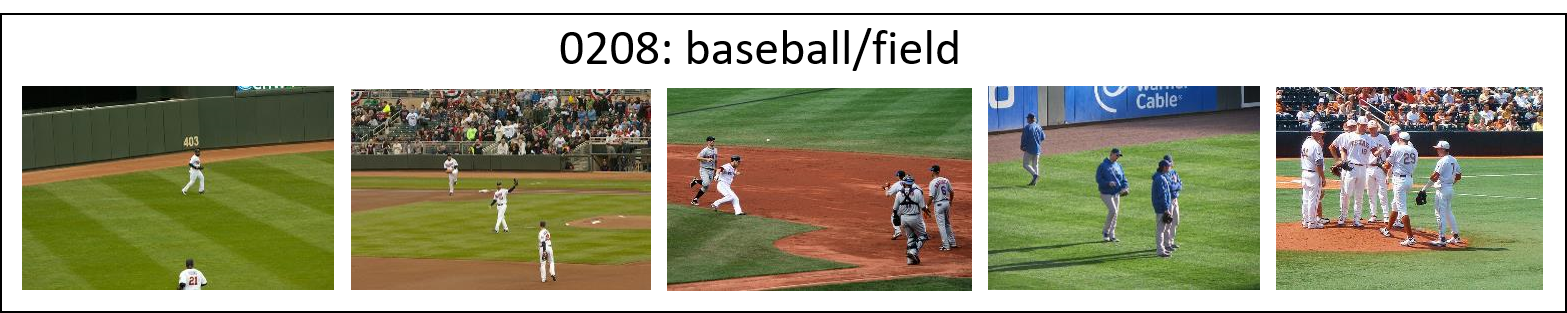} \ \ 
\includegraphics[trim = 5mm 5mm 5mm 5mm, clip, scale=0.312]{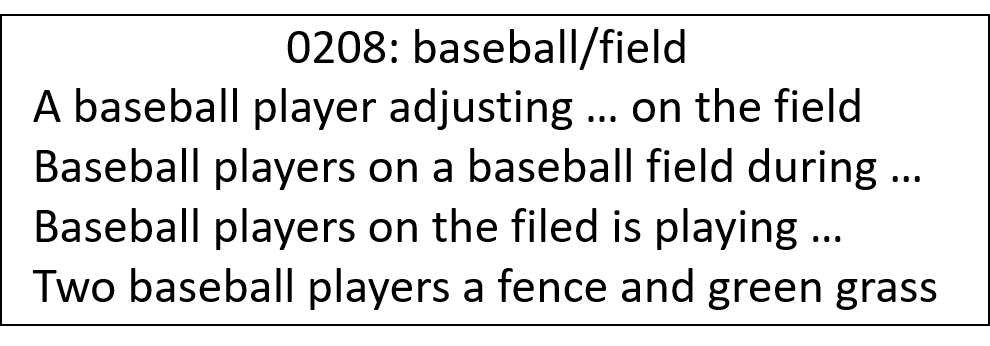} \\ \vspace{+0.2em}
\includegraphics[trim = 5mm 3mm 5mm 4mm, clip, scale=0.342]{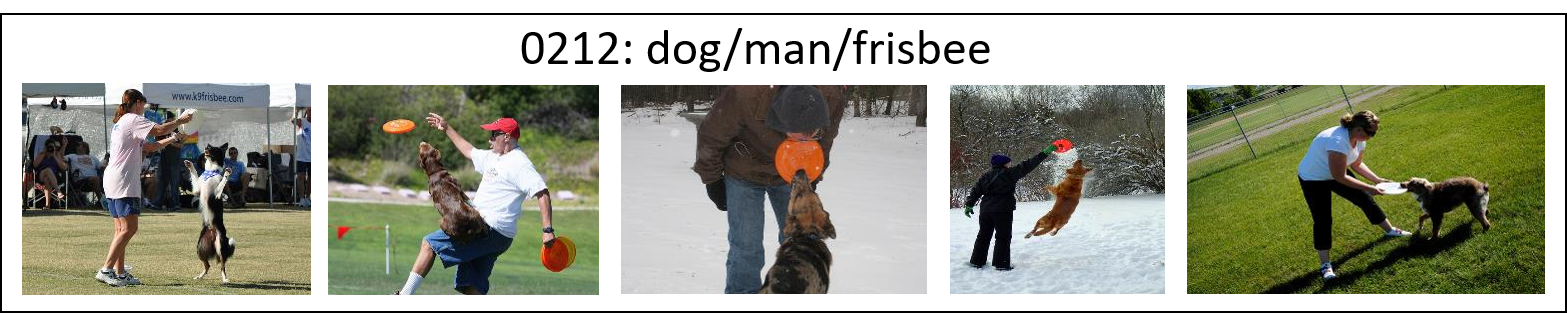} \ \ 
\includegraphics[trim = 5mm 5mm 5mm 5mm, clip, scale=0.312]{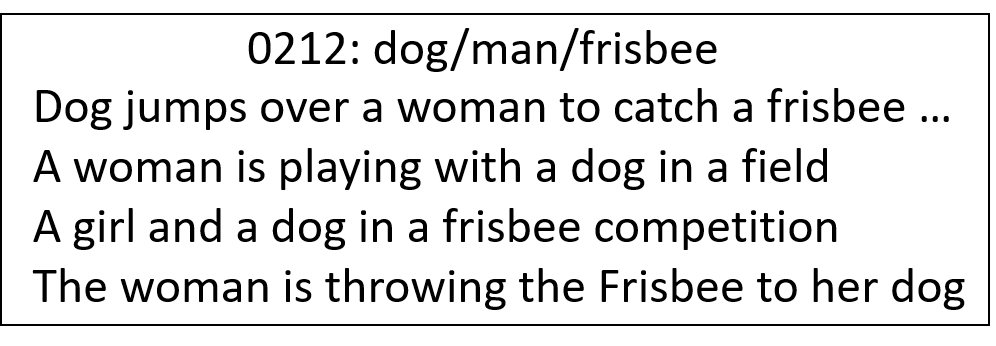} \\ \vspace{+0.2em}
\includegraphics[trim = 5mm 3mm 5mm 4mm, clip, scale=0.342]{figs/vis/0221_baseball_base_img.png} \ \ 
\includegraphics[trim = 5mm 5mm 5mm 5mm, clip, scale=0.312]{figs/vis/0221_baseball_base_txt.png} \\ \vspace{+0.2em}
\includegraphics[trim = 5mm 3mm 5mm 4mm, clip, scale=0.342]{figs/vis/0224_bus_double_img.png} \ \ 
\includegraphics[trim = 5mm 5mm 5mm 5mm, clip, scale=0.312]{figs/vis/0224_bus_double_txt.png} \\ \vspace{+0.2em}
\includegraphics[trim = 5mm 3mm 5mm 4mm, clip, scale=0.342]{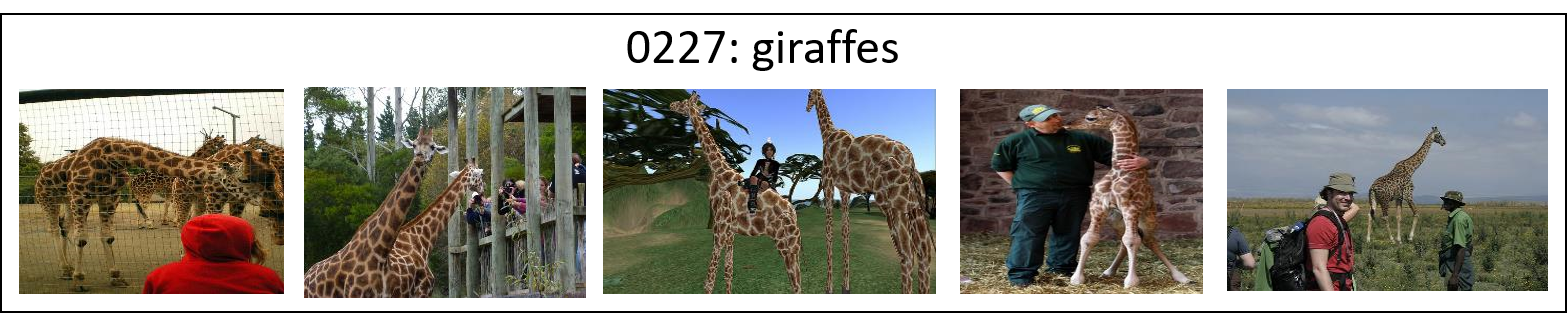} \ \ 
\includegraphics[trim = 5mm 5mm 5mm 5mm, clip, scale=0.312]{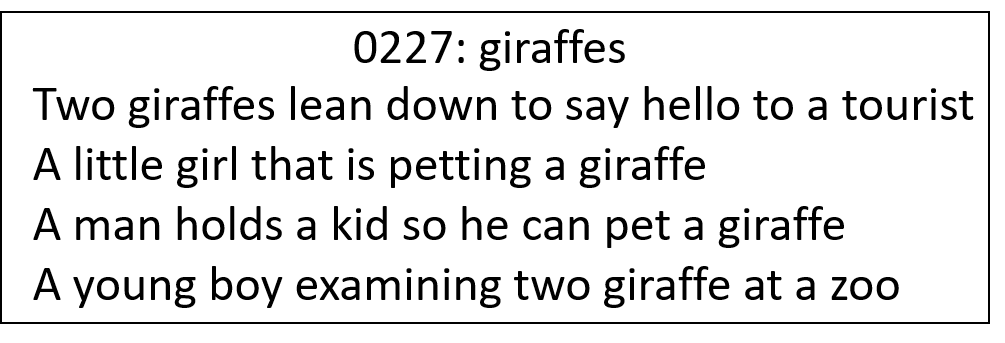} \\ \vspace{+0.2em}
\includegraphics[trim = 5mm 3mm 5mm 4mm, clip, scale=0.342]{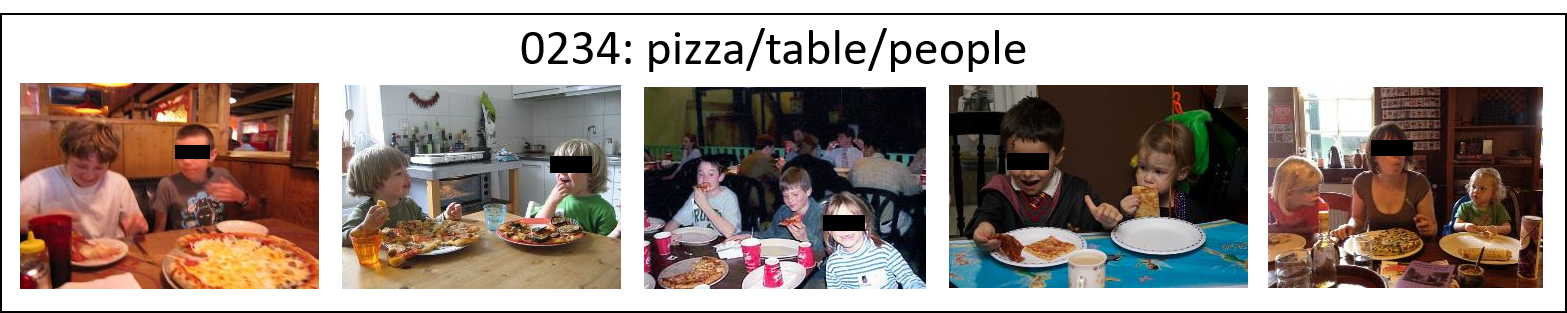} \ \ 
\includegraphics[trim = 5mm 5mm 5mm 5mm, clip, scale=0.312]{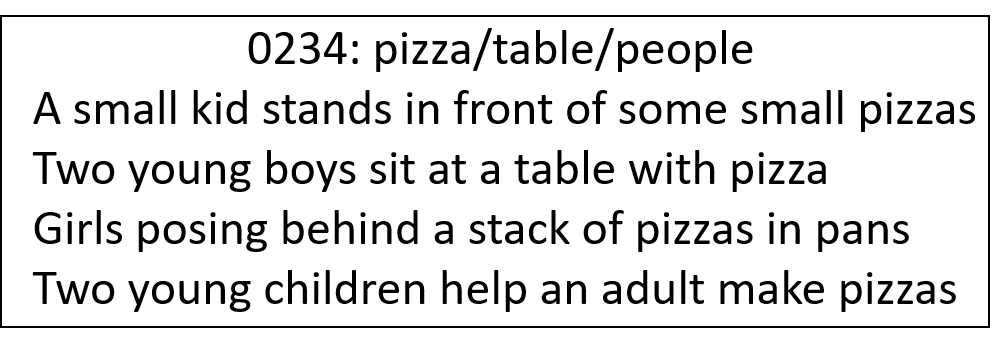}
\end{center}
\vspace{-1.0em}
\caption{Some randomly selected clusters with images and texts that belong to them. Each cluster, titled by {\em ID: keywords}, shows randomly chosen 5 images and 4 texts. 
}
%\vspace{-1.0em}
\label{appappfig:clusters_examples_1}
\end{figure}

\begin{figure}[t!]
\vspace{-0.5em}
\begin{center}
%
%\begin{subfigure}[b]{0.9\textwidth}
\centering
\includegraphics[trim = 5mm 3mm 5mm 4mm, clip, scale=0.342]{figs/vis/0253_baseball_crowd_img.png} \ \ 
\includegraphics[trim = 5mm 5mm 5mm 5mm, clip, scale=0.312]{figs/vis/0253_baseball_crowd_txt.png} \\ \vspace{+0.2em}
\includegraphics[trim = 5mm 3mm 5mm 4mm, clip, scale=0.342]{figs/vis/0306_kites_flying_img.png} \ \ 
\includegraphics[trim = 5mm 5mm 5mm 5mm, clip, scale=0.312]{figs/vis/0306_kites_flying_txt.png} \\ \vspace{+0.2em}
\includegraphics[trim = 5mm 3mm 5mm 4mm, clip, scale=0.342]{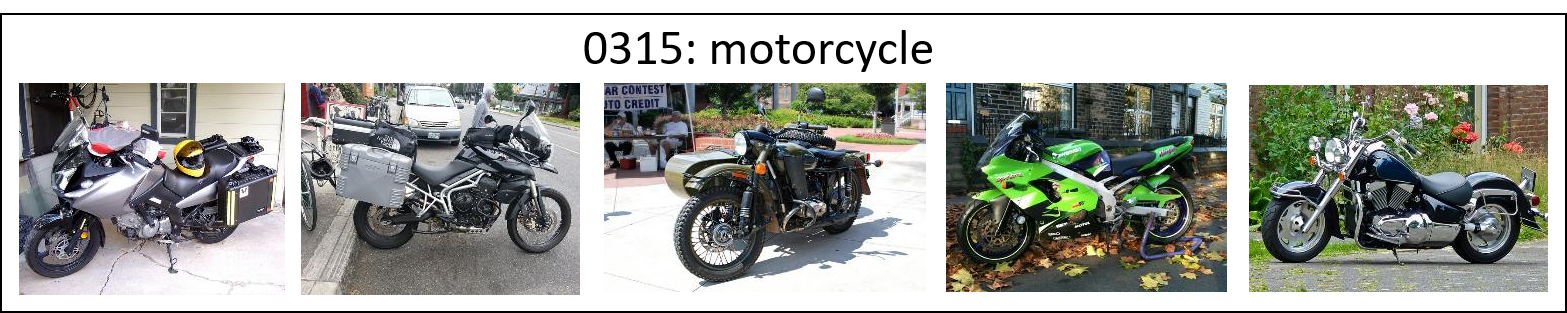} \ \ 
\includegraphics[trim = 5mm 5mm 5mm 5mm, clip, scale=0.312]{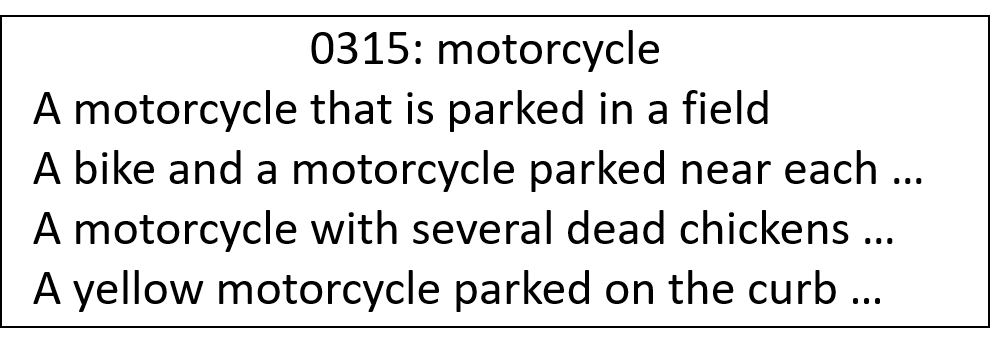} \\ \vspace{+0.2em}
\includegraphics[trim = 5mm 3mm 5mm 4mm, clip, scale=0.342]{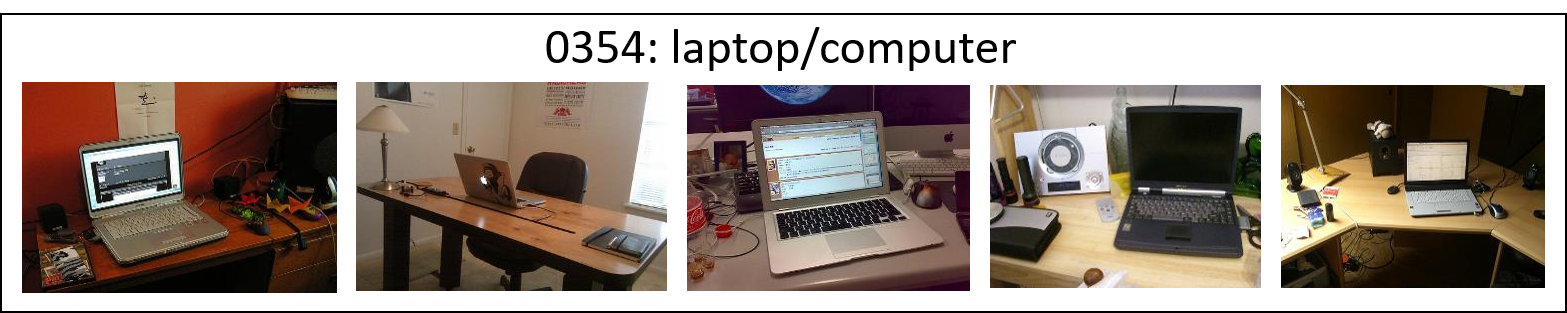} \ \ 
\includegraphics[trim = 5mm 5mm 5mm 5mm, clip, scale=0.312]{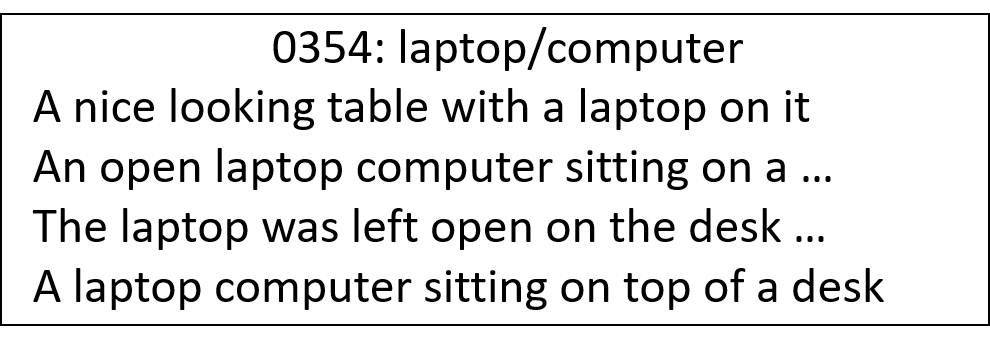} \\ \vspace{+0.2em}
\includegraphics[trim = 5mm 3mm 5mm 4mm, clip, scale=0.342]{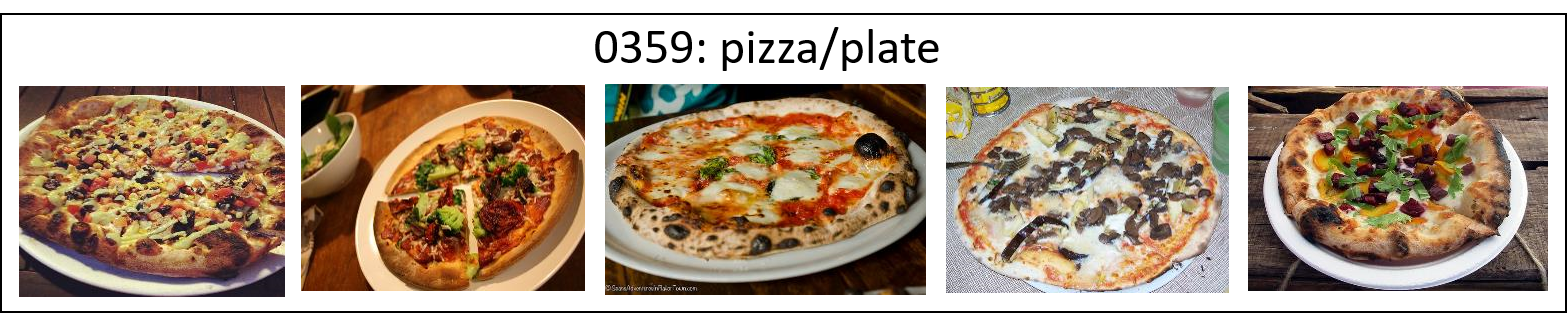} \ \ 
\includegraphics[trim = 5mm 5mm 5mm 5mm, clip, scale=0.312]{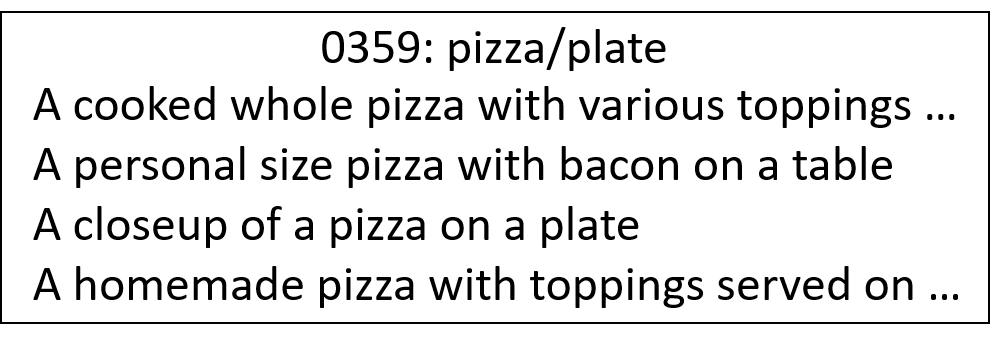} \\ \vspace{+0.2em}
\includegraphics[trim = 5mm 3mm 5mm 4mm, clip, scale=0.342]{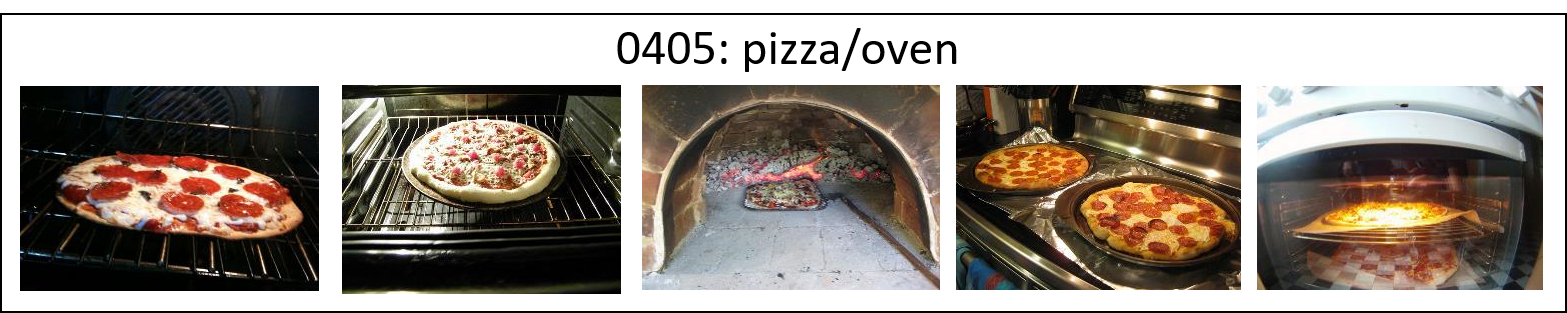} \ \ 
\includegraphics[trim = 5mm 5mm 5mm 5mm, clip, scale=0.312]{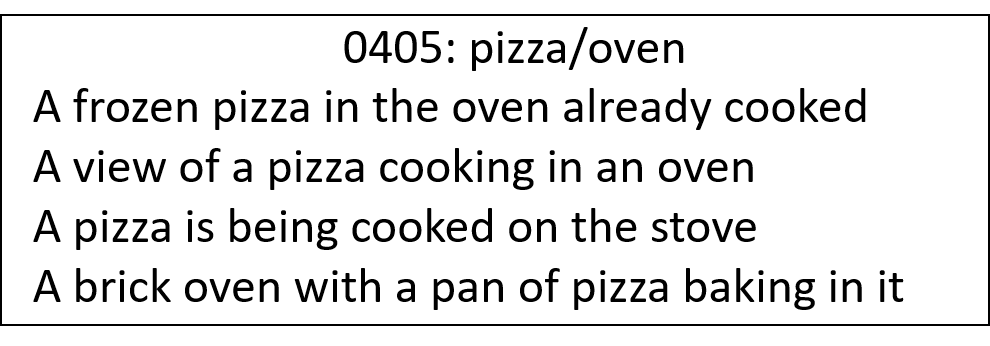} \\ \vspace{+0.2em}
\includegraphics[trim = 5mm 3mm 5mm 4mm, clip, scale=0.342]{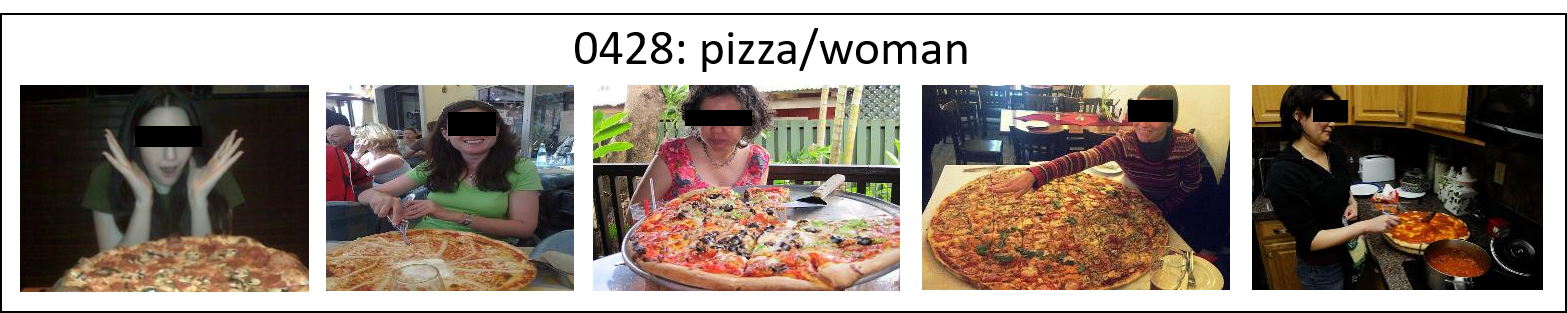} \ \ 
\includegraphics[trim = 5mm 5mm 5mm 5mm, clip, scale=0.312]{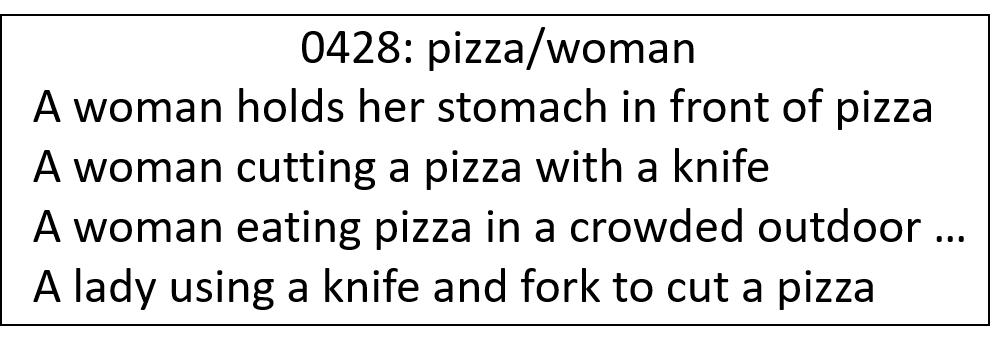} \\ \vspace{+0.2em}
\includegraphics[trim = 5mm 3mm 5mm 4mm, clip, scale=0.342]{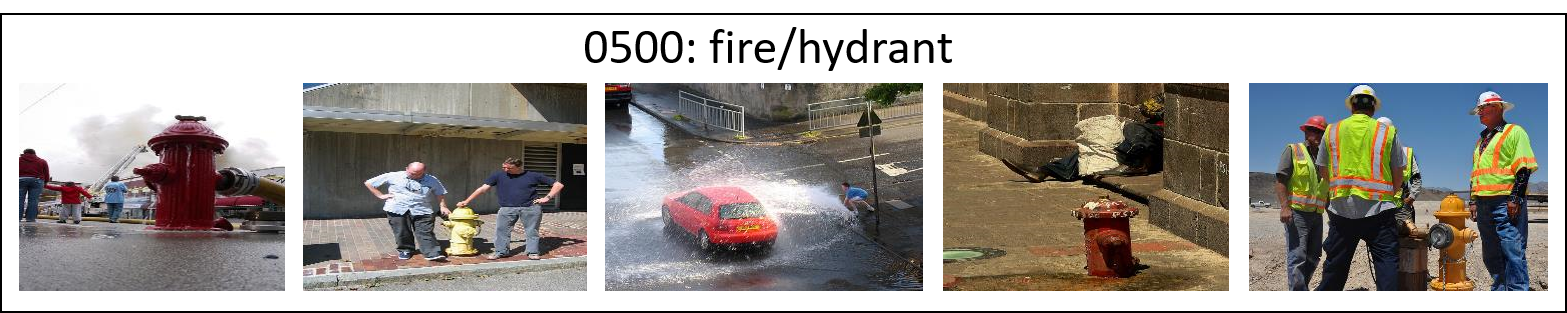} \ \ 
\includegraphics[trim = 5mm 5mm 5mm 5mm, clip, scale=0.312]{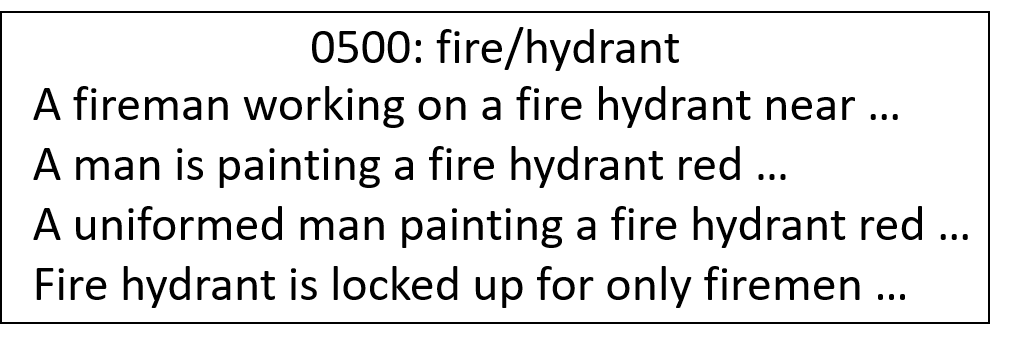} \\ \vspace{+0.2em}
\includegraphics[trim = 5mm 3mm 5mm 4mm, clip, scale=0.342]{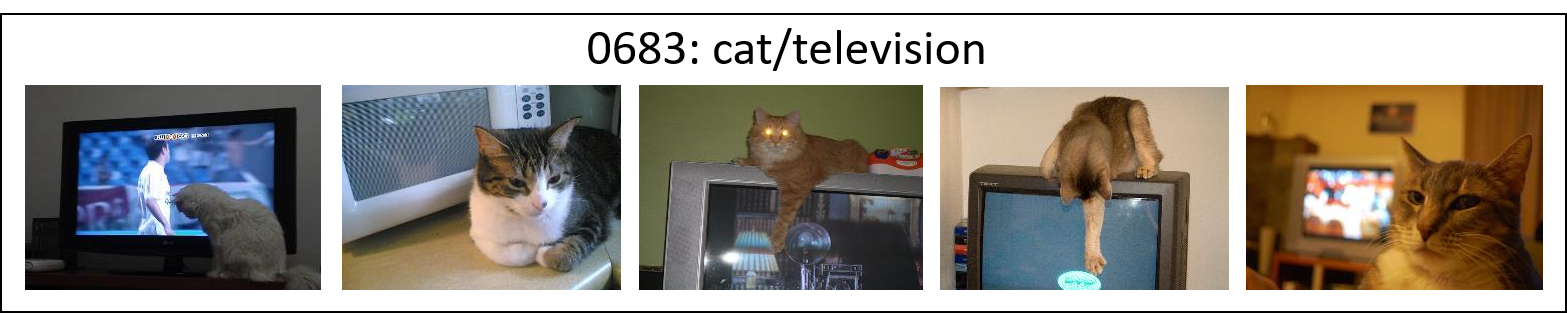} \ \ 
\includegraphics[trim = 5mm 5mm 5mm 5mm, clip, scale=0.312]{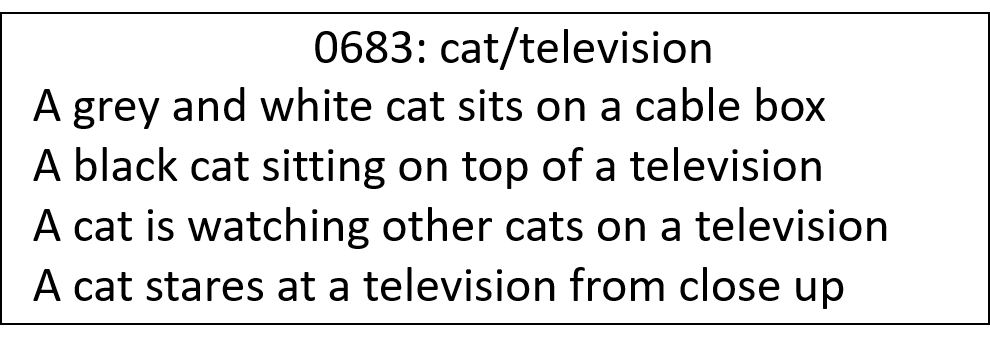} \\ \vspace{+0.2em}
\includegraphics[trim = 5mm 3mm 5mm 4mm, clip, scale=0.342]{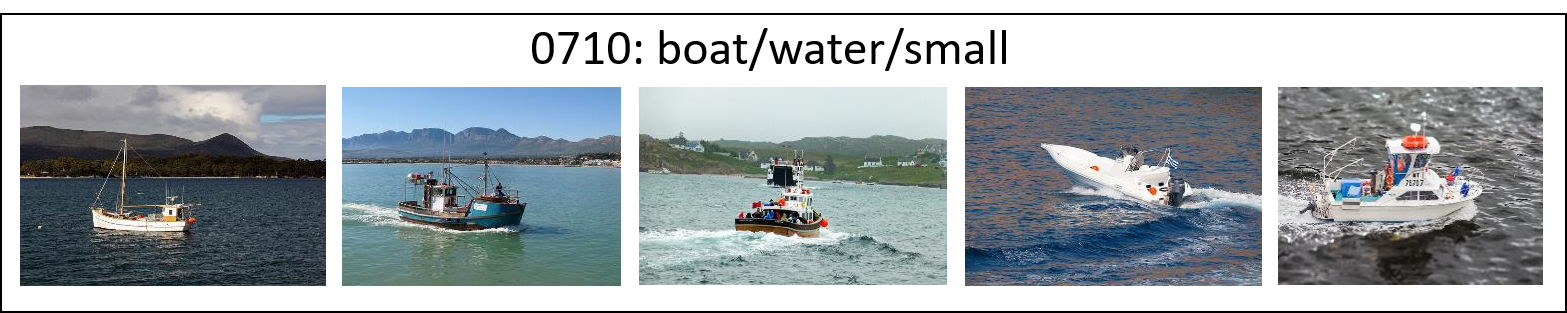} \ \ 
\includegraphics[trim = 5mm 5mm 5mm 5mm, clip, scale=0.312]{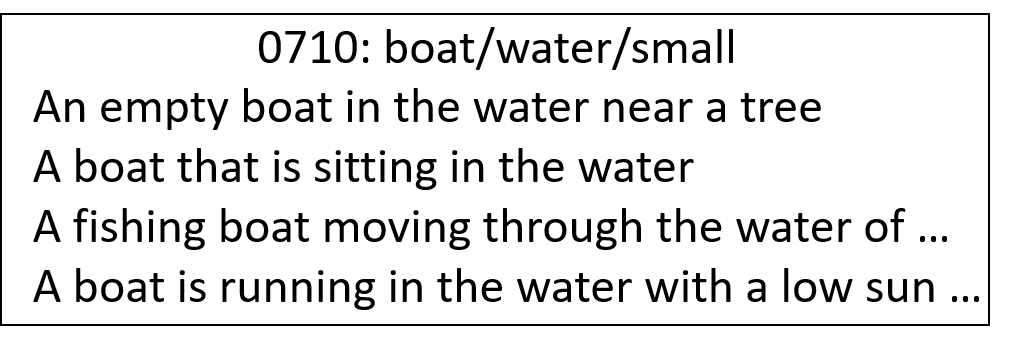} \\ \vspace{+0.2em}
\includegraphics[trim = 5mm 3mm 5mm 4mm, clip, scale=0.342]{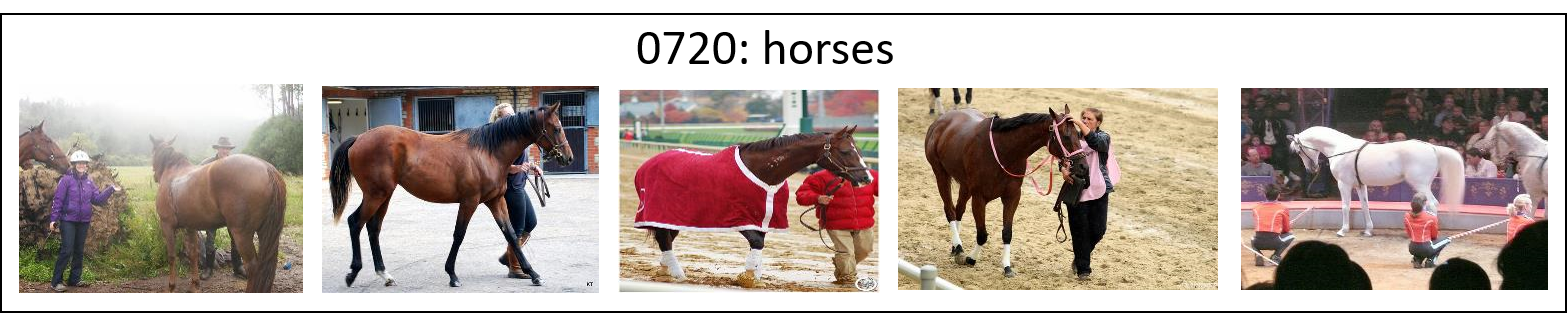} \ \ 
\includegraphics[trim = 5mm 5mm 5mm 5mm, clip, scale=0.312]{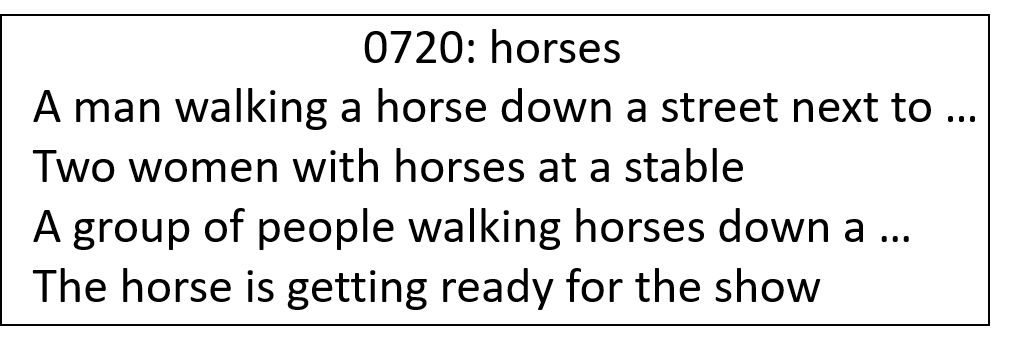} \\ \vspace{+0.2em}
\includegraphics[trim = 5mm 3mm 5mm 4mm, clip, scale=0.342]{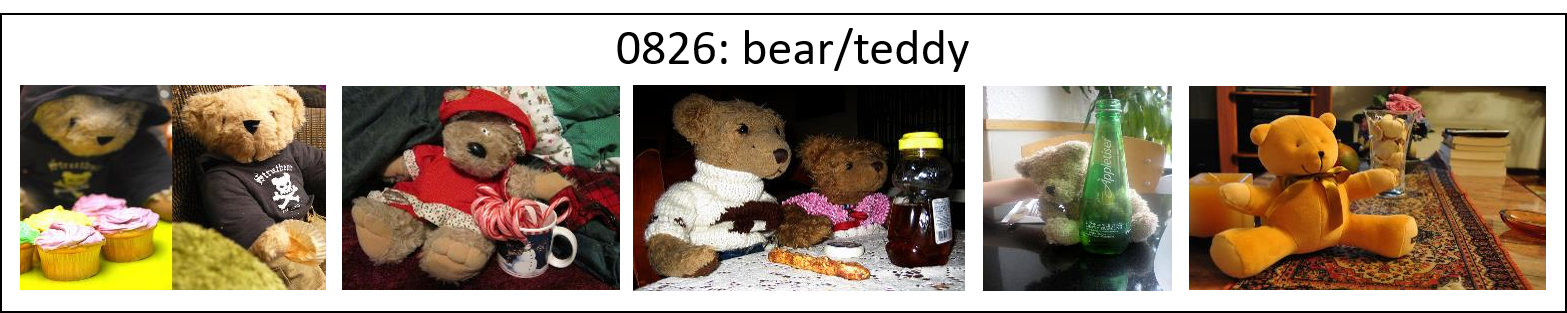} \ \ 
\includegraphics[trim = 5mm 5mm 5mm 5mm, clip, scale=0.312]{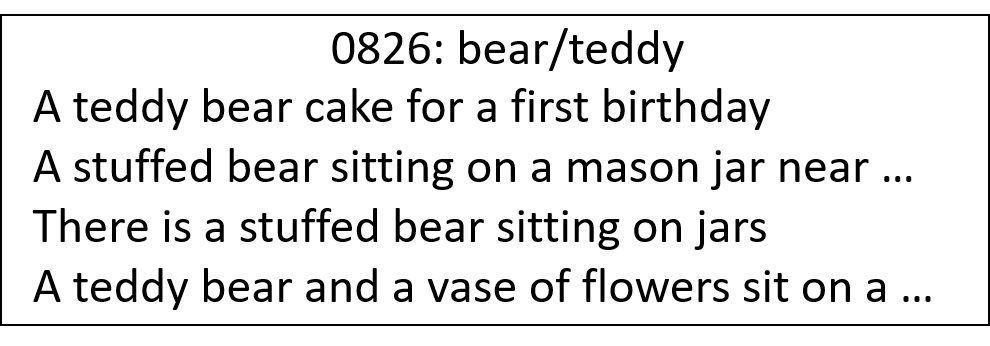} \\ \vspace{+0.2em}
\includegraphics[trim = 5mm 3mm 5mm 4mm, clip, scale=0.342]{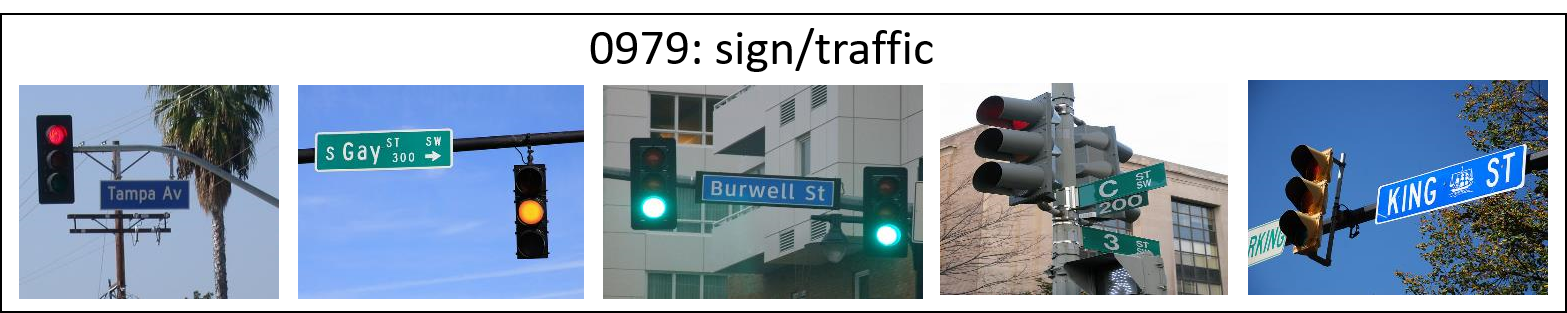} \ \ 
\includegraphics[trim = 5mm 5mm 5mm 5mm, clip, scale=0.312]{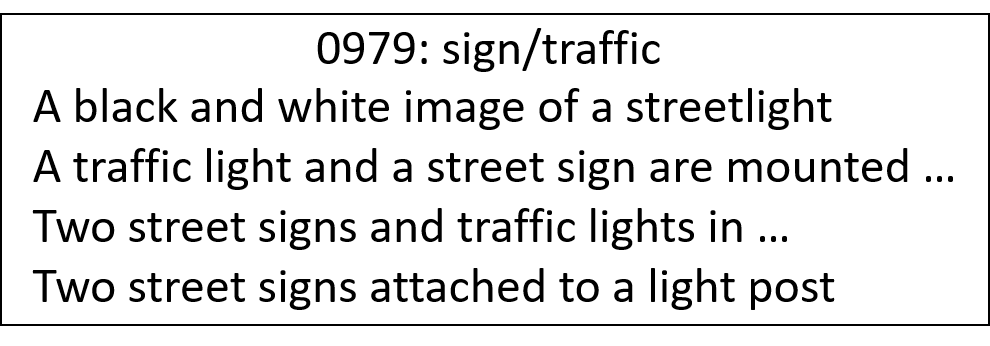}
\end{center}
\vspace{-1.0em}
\caption{More clusters (continued from Fig.~\ref{appfig:clusters_examples_1}).
}
%\vspace{-1.0em}
\label{appappfig:clusters_examples_2}
\end{figure}

As qualitative analysis, we visualize the learned clusters to see if they capture meaningful semantic information. On MS-COCO (trained with the number of classes $K=1000$), we organize images and texts by their assigned cluster labels using the learned prototype classification model,
%%%%
%\vspace{-0.5em}
\begin{align}
%\vspace{-0.5em}
p(y=j|x^M) = \frac{\exp(p_j^\top \phi^M(x) / \tau)}{\sum_{l}\exp(p_{l}^\top \phi^M(x) / \tau)}, \ \ \ \ %M = A \textrm{ or } B
M \in \{A,B\}
% , \ \ \ \
% p(y=j|x^B) = \frac{\exp(p_j^\top \phi^B(x) / \tau)}{\sum_{j'=1}^{K}\exp(p_{j'}^\top \phi^B(x) / \tau)}, 
\label{appappeq:class_predictive}
\end{align}
%%%%
in our SwAMP.
We first visually inspect individual clusters, images and texts that belong to each cluster. Some examples are shown in Fig.~\ref{appappfig:clusters_examples_1} and Fig.~\ref{appappfig:clusters_examples_2}, each cluster contains semantically coherent data samples. 
Then we inspect texts (captions) in each cluster, and select a few keywords, those words that appear the most frequently in the texts. These keywords for each cluster consist of objects (noun) and/or actions (verb) that faithfully describe the cluster and data samples that belong to it. The full list is provided in Fig.~\ref{appappfig:clusters_keywords}. 
%%%%
%%%%

\subsection{Clustering based on Whole Contents}
In Fig.~\ref{appfig:clusters_examples_1} and Fig.~\ref{appfig:clusters_examples_2}, looking at clusters: 
\begin{itemize}
\item Group-A $=(0100, 0234, 0359, 0405, 0428)$
\item Group-B $=(0195, 0208, 0221, 0253)$
\item Group-C $=(0180, 0683)$ 
\end{itemize}
the clusters within these groups are all related to the semantic meaning of {\em pizza}, {\em baseball}, and {\em cat}, respectively. 
However, they have different details in either activity or focus/scene: 
{\em man eating}, {\em people and table}, {\em on plate}, {\em oven}, and {\em with woman} for Group-A; {\em swing}, {\em on field}, {\em base playing}, and {\em crowd scene} for Group-B; {\em on bed} and {\em television} for Group-C. This means that SwAMP finds clusters based on the whole contents (objects, acitivities, and scenes), % from both modalities,
instead of doing merely object-based clustering. 
%However, even though the semantics are similar there are subtle detailed separations. (baseball, pizza), hierarchical clustering. 

\subsection{Class Imbalance}
Although we have roughly equal numbers of samples per cluster, we can see many redundant/repeated clusters in Fig.~\ref{appfig:clusters_keywords}. This means that some clusters are overlapped with others in terms of semantic meaning, constituting larger {\em super}-clusters. 
These clusters are related to dominant data samples (e.g., cat, dog, tennis, baseball). This implies that the SwAMP can effectively deal with imbalance of semantic classes that can reside in data. 

%%%%
%%%%

%%%%
%%%%

%%%%%%%%%%%%%%%%%%%%%%%%%%%%%%%%%%%%%%%%%%%%%%%%%%%%%%%%%%%%%%%%%%%%%%%%%%%%%%%
%%%%%%%%%%%%%%%%%%%%%%%%%%%%%%%%%%%%%%%%%%%%%%%%%%%%%%%%%%%%%%%%%%%%%%%%%%%%%%%
\section{Ablation Study}\label{appsec:ablation}

In our experiments, we chose the hyperparameters by cross validation with grid search. We perform empirical study on the impact of two important hyperparameters in our model: the number of classes $K$ and SwAMP loss trade-off $\lambda$.

\textbf{Number of classes ($K$).} 
Recall that the best $K$ values we chose were: $K=1000$ for the image-text retrieval datasets and $K=500$ for text-based video retrieval. To see how the retrieval performance is affected by other choices of $K$, we conduct experiments by varying $K$ around the optimal values. The results are shown in Fig.~\ref{appfig:flickr_impact_of_K}, Fig.~\ref{appfig:coco_impact_of_K}, Fig.~\ref{appfig:yc2_impact_of_K}, Fig.~\ref{appfig:msrvtt_impact_of_K}, and Fig.~\ref{appfig:lsmdc_impact_of_K}. 
%from $\{200,500,1000,2000,3000\}$, and record the R@1 scores for both pair and class based error types for our SwAMP model. 
Clearly, very small $K$ has low retrieval performance (R@1), and increasing $K$ leads to improvement. However, beyond certain points, there is no benefit of increasing $K$ and we even see performance degradation, which agrees with the observations from previous work~\citep{sela,swav}. This is perhaps due to the difficulty of assigning meaningful cluster labels in optimal transport. Overall, with properly chosen $K$, SwAMP outperforms contrastive learning, signifying that SwAMP's grouping/clustering of similar instances is more effective than vanilla instance discrimination. The fact that the optimal $K$ values are different in two tasks (image-text and video-text) implies that the best cardinality of semantic clusters is highly dependent on the dataset characteristics (e.g., size and semantic diversity). 

%%%%

\textbf{SwAMP loss trade-off ($\lambda$).} 
We perform sensitivity analysis on $\lambda$, the strength of the SwAMP loss. For different values of $\lambda$, the retrieval scores (R@1) are shown in Fig.~\ref{appfig:flickr_impact_of_lambda}, Fig.~\ref{appfig:coco_impact_of_lambda}, Fig.~\ref{appfig:yc2_impact_of_lambda}, Fig.~\ref{appfig:msrvtt_impact_of_lambda}, and Fig.~\ref{appfig:lsmdc_impact_of_lambda}.  Our model remains better than contrastive learning for large intervals of different $\lambda$'s, and the performance is not very sensitive to $\lambda$.

%%%%
\begin{figure}[t!]
%\vspace{-1.5em}
\begin{center}
%
%\begin{subfigure}[b]{0.9\textwidth}
\centering
\includegraphics[trim = 5mm 2mm 5mm 4mm, clip, scale=0.34]{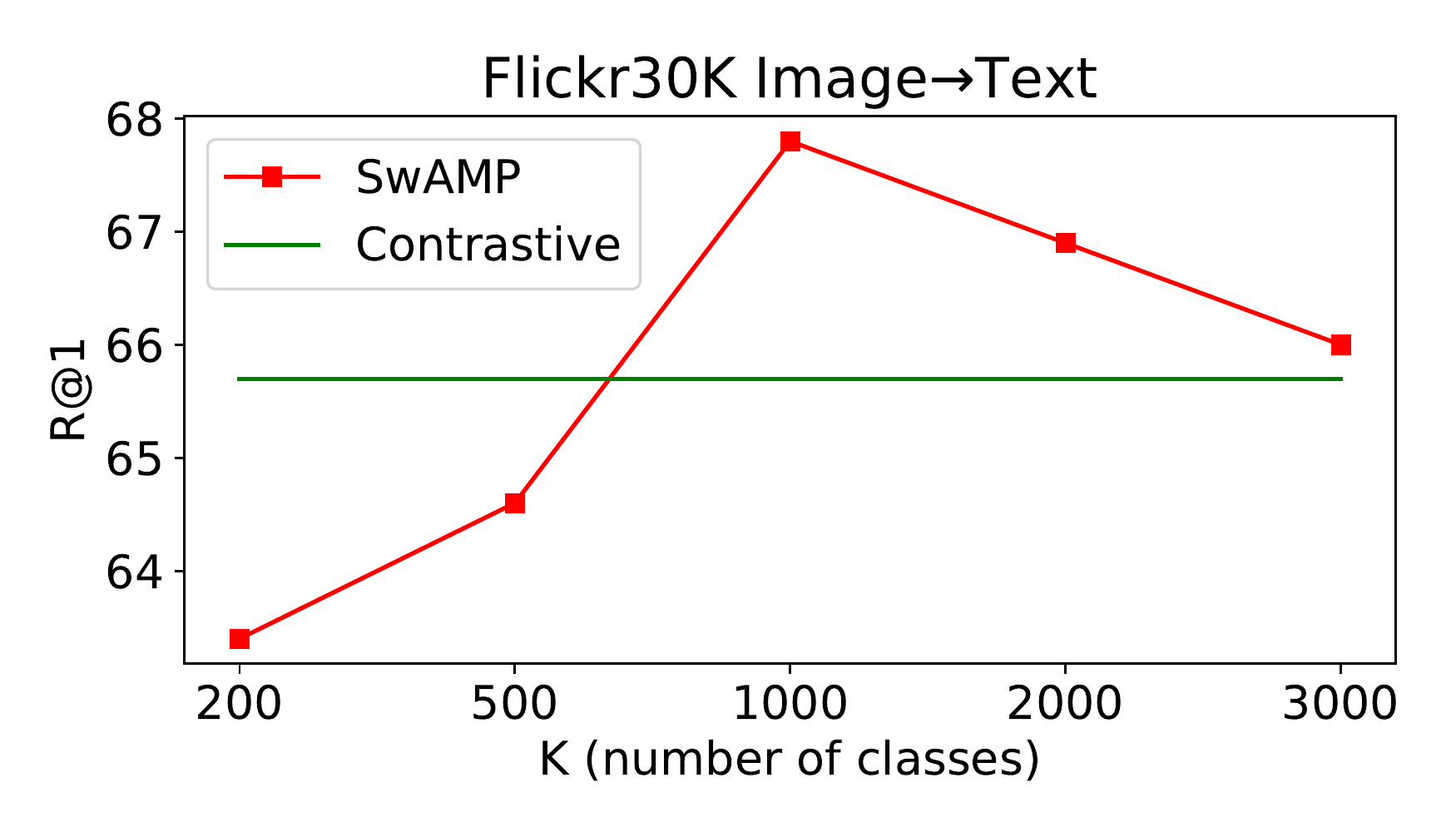} \ \  
\includegraphics[trim = 5mm 2mm 5mm 4mm, clip, scale=0.34]{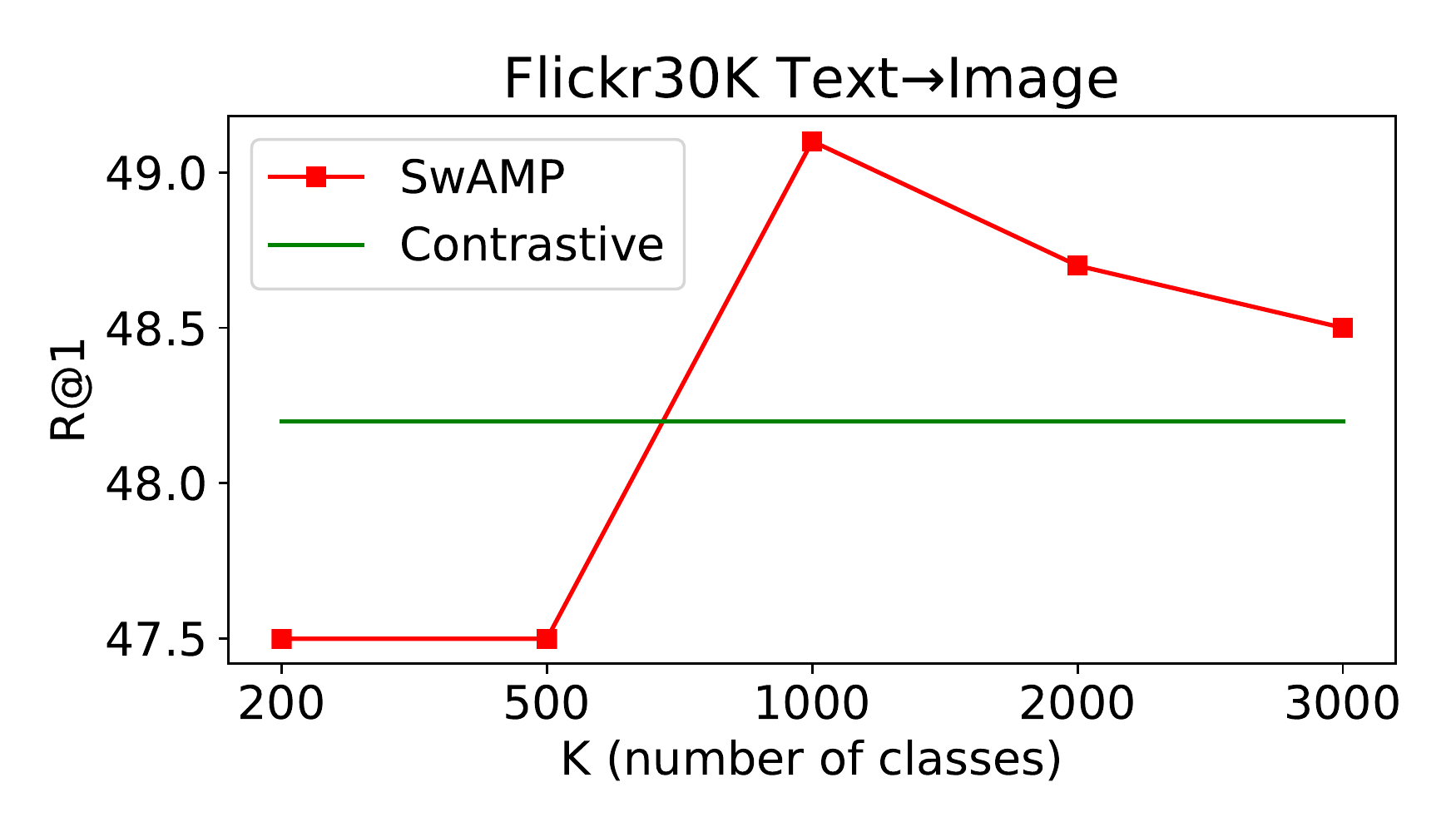}
\end{center}
\vspace{-1.5em}
\caption{(Flickr30K) Impact of the number of classes ($K$). 
}
%\vspace{-1.0em}
\label{appappfig:flickr_impact_of_K}
\end{figure}
%%%%

%%%%
\begin{figure}[t!]
%\vspace{-1.5em}
\begin{center}
%
%\begin{subfigure}[b]{0.9\textwidth}
\centering
\includegraphics[trim = 5mm 2mm 5mm 4mm, clip, scale=0.34]{figs/coco_impact_of_K_i2t.pdf} \ \  
\includegraphics[trim = 1mm 2mm 5mm 4mm, clip, scale=0.34]{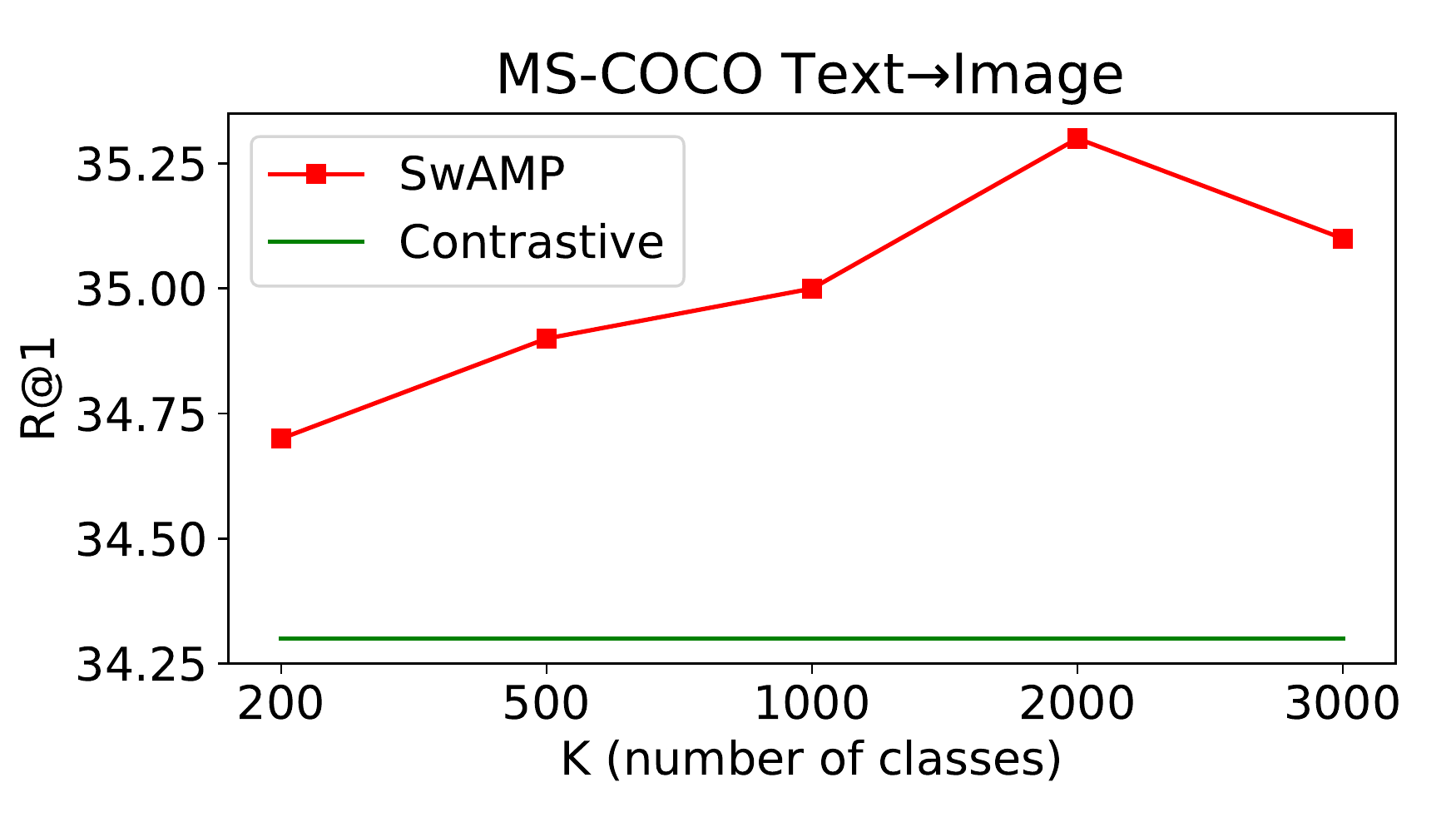}
\end{center}
\vspace{-1.5em}
\caption{(MS-COCO) Impact of the number of classes ($K$). 
}
%\vspace{-1.0em}
\label{appappfig:coco_impact_of_K}
\end{figure}
%%%%

%%%%
\begin{figure}[t!]
%\vspace{-1.5em}
\begin{center}
%
%\begin{subfigure}[b]{0.9\textwidth}
\centering
\includegraphics[trim = 5mm 2mm 5mm 4mm, clip, scale=0.34]{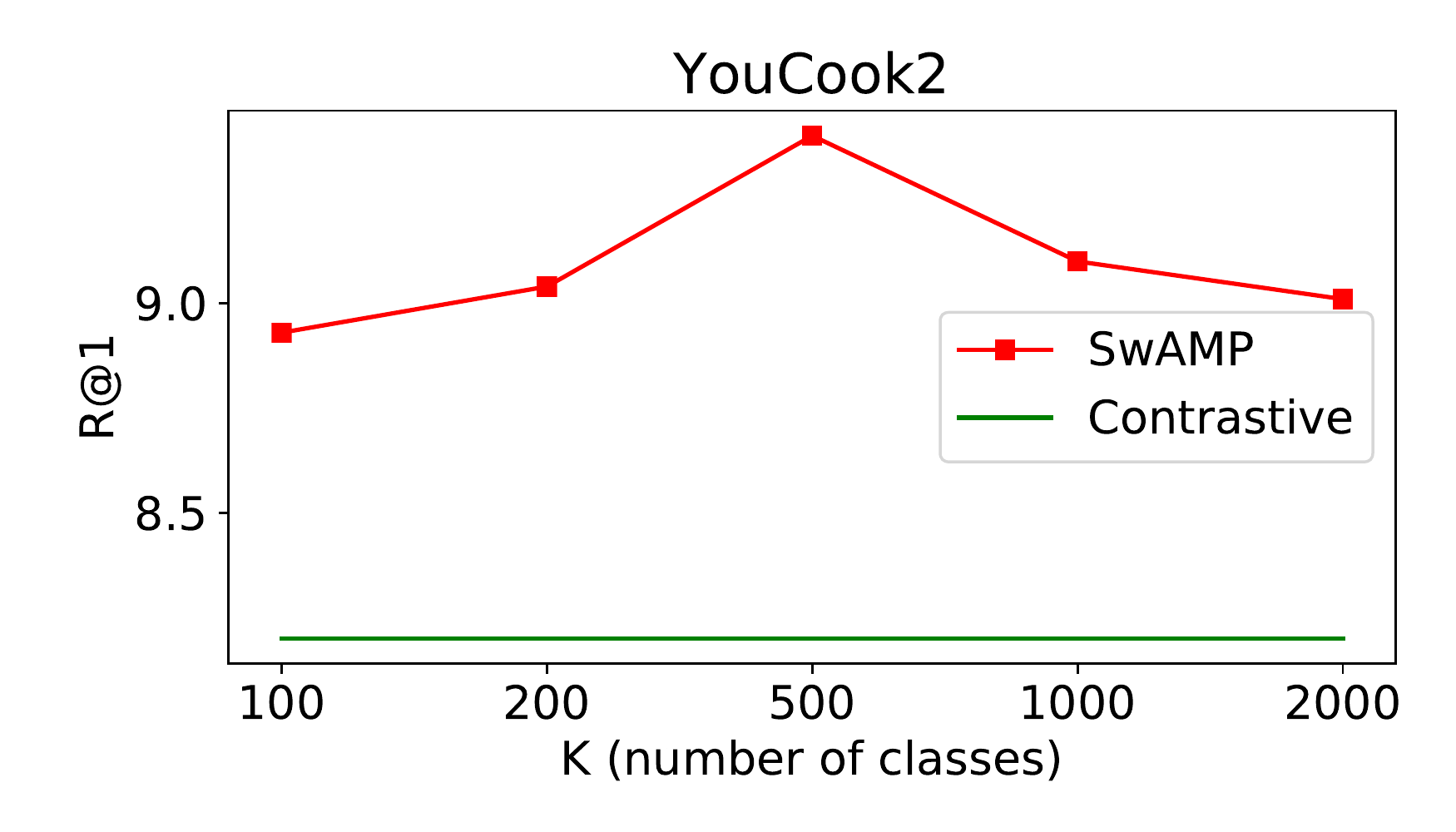}
\end{center}
\vspace{-1.5em}
\caption{(YouCook2) Impact of the number of classes ($K$). 
}
%\vspace{-1.0em}
\label{appfig:yc2_impact_of_K}
\end{figure}
%%%%

%%%%
\begin{figure}[t!]
%\vspace{-1.5em}
\begin{center}
%
%\begin{subfigure}[b]{0.9\textwidth}
\centering
\includegraphics[trim = 5mm 2mm 5mm 4mm, clip, scale=0.34]{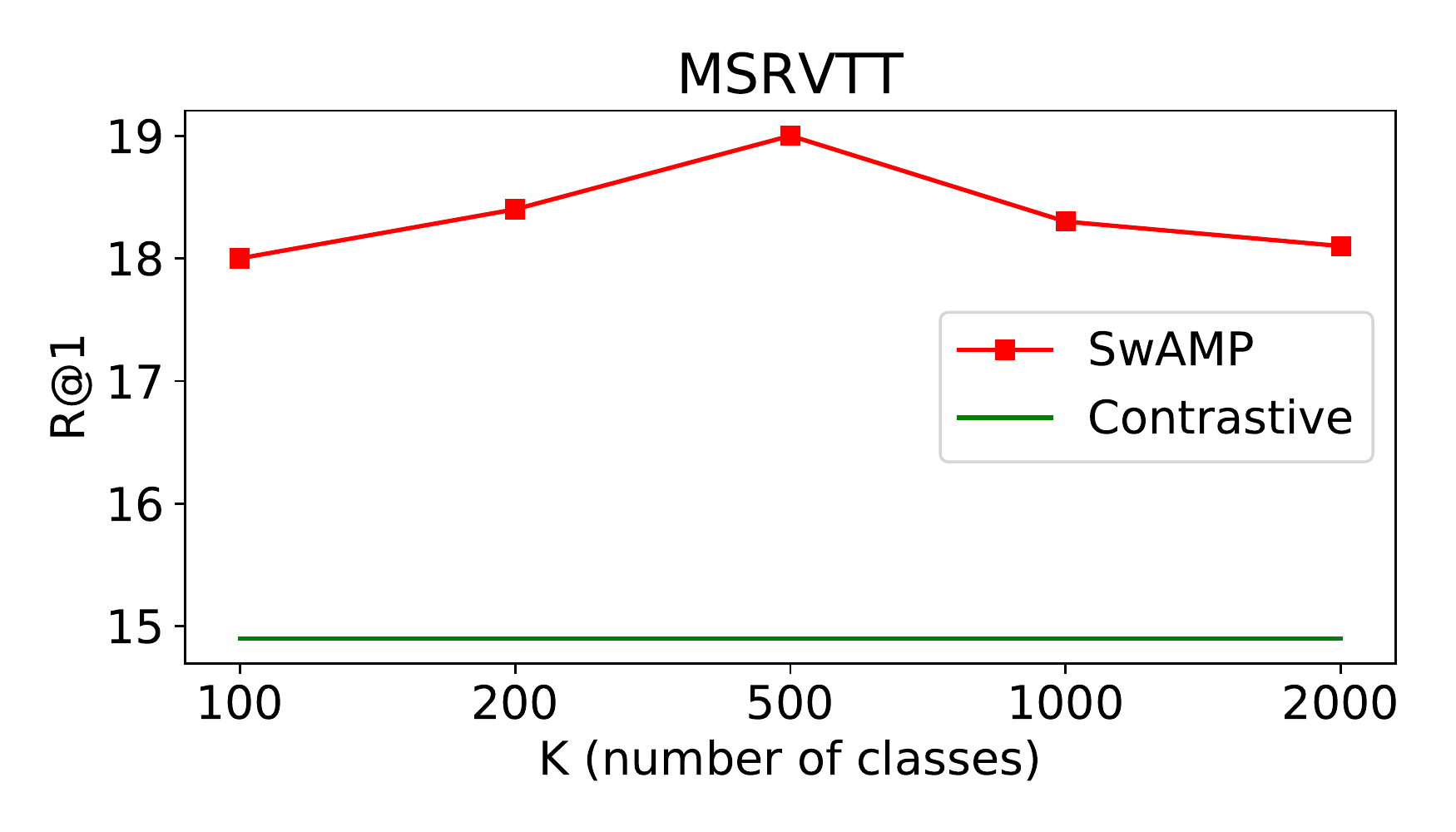}
\end{center}
\vspace{-1.5em}
\caption{(MSRVTT) Impact of the number of classes ($K$). 
}
%\vspace{-1.0em}
\label{appfig:msrvtt_impact_of_K}
\end{figure}
%%%%

%%%%
\begin{figure}[t!]
%\vspace{-1.5em}
\begin{center}
%
%\begin{subfigure}[b]{0.9\textwidth}
\centering
\includegraphics[trim = 5mm 2mm 5mm 4mm, clip, scale=0.34]{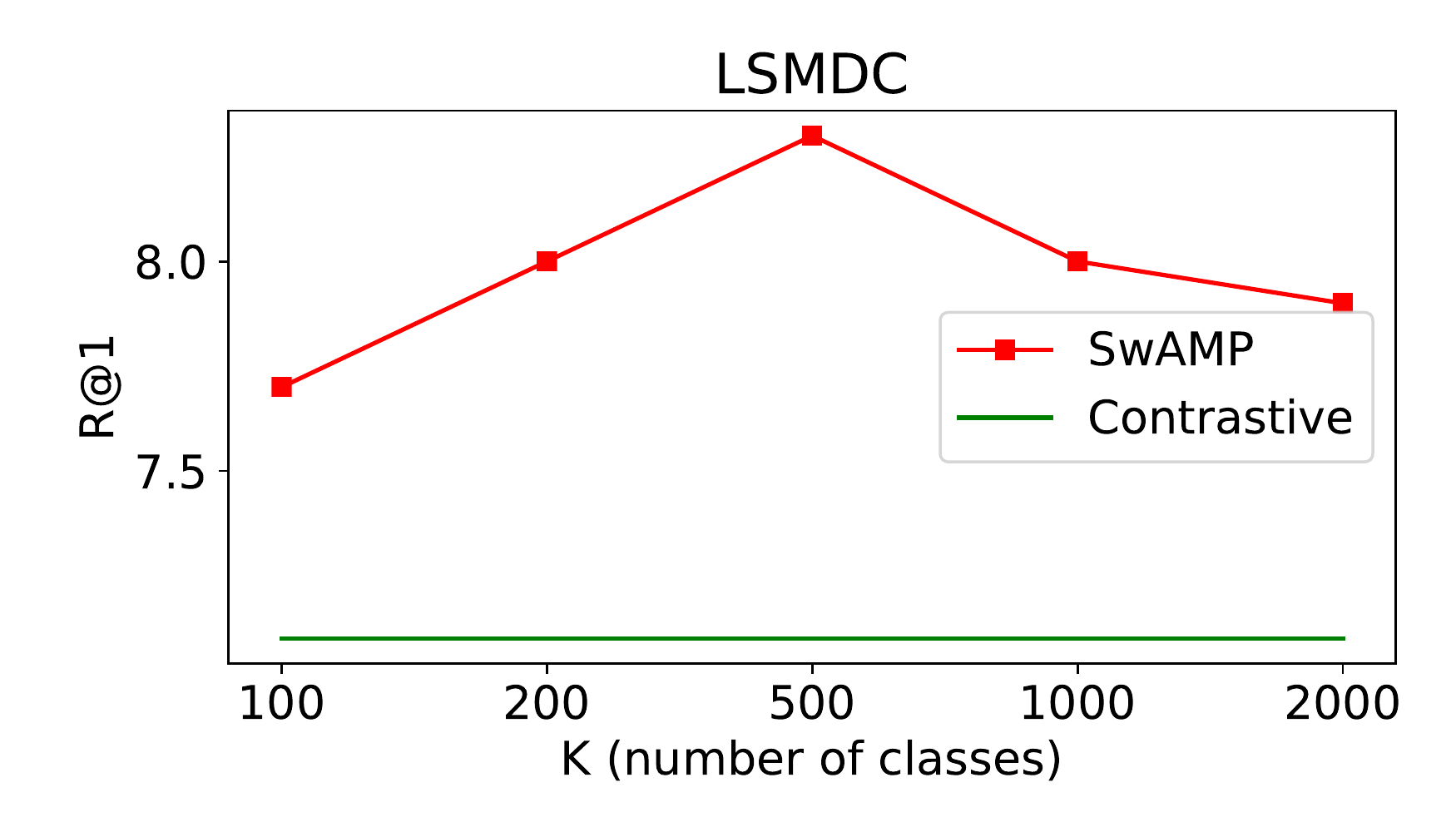}
\end{center}
\vspace{-1.5em}
\caption{(LSMDC) Impact of the number of classes ($K$). 
}
%\vspace{-1.0em}
\label{appfig:lsmdc_impact_of_K}
\end{figure}
%%%%

%%%%
\begin{figure}[t!]
%\vspace{-1.5em}
\begin{center}
%
%\begin{subfigure}[b]{0.9\textwidth}
\centering
\includegraphics[trim = 5mm 2mm 5mm 4mm, clip, scale=0.34]{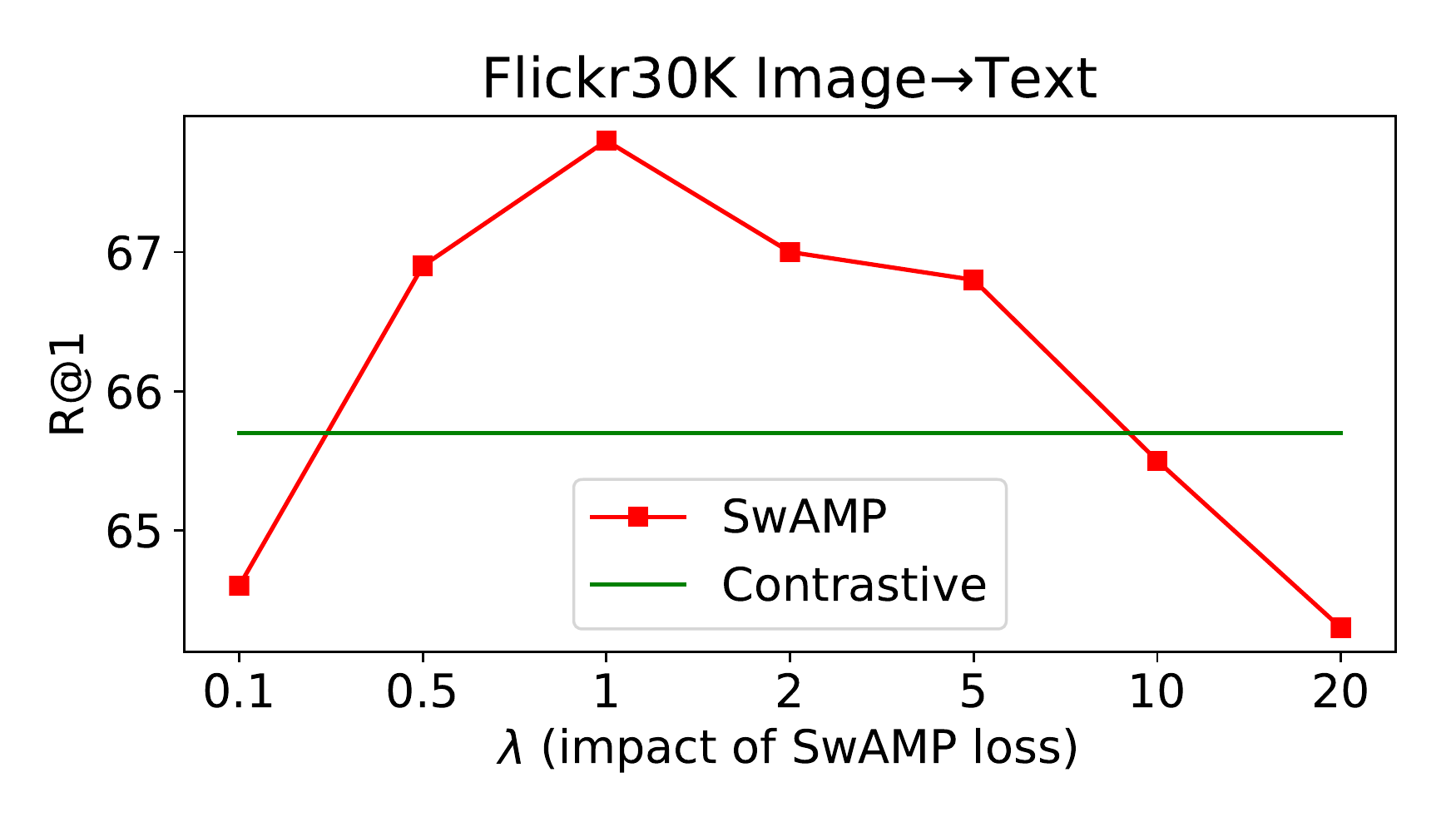} \ \  
\includegraphics[trim = 5mm 2mm 5mm 4mm, clip, scale=0.34]{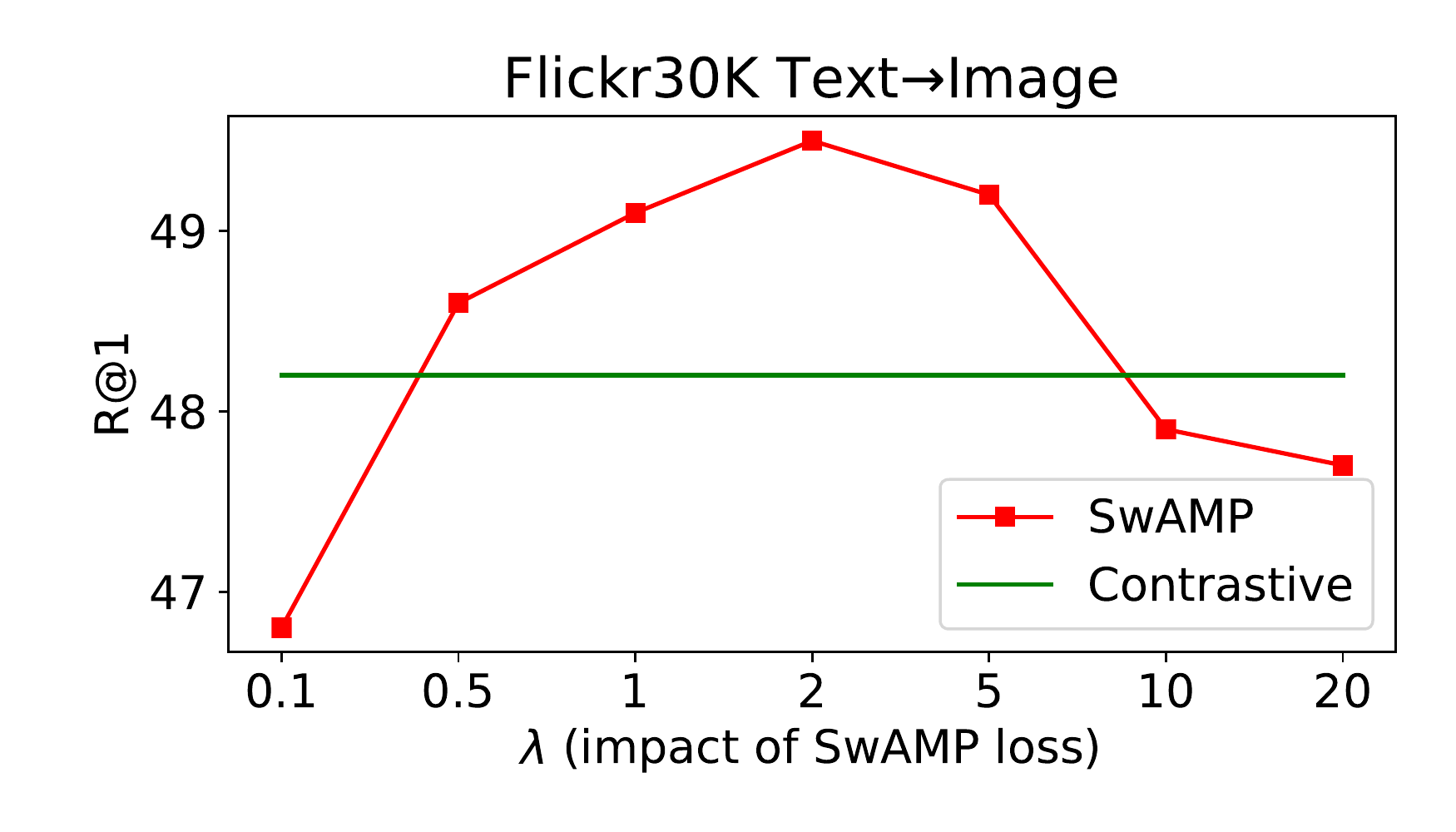}
\end{center}
\vspace{-1.5em}
\caption{(Flickr30K) Impact of the SwAMP loss ($\lambda$). 
}
%\vspace{-1.0em}
\label{appfig:flickr_impact_of_lambda}
\end{figure}
%%%%

%%%%
\begin{figure}[t!]
%\vspace{-1.5em}
\begin{center}
%
%\begin{subfigure}[b]{0.9\textwidth}
\centering
\includegraphics[trim = 5mm 2mm 5mm 4mm, clip, scale=0.34]{figs/coco_impact_of_lambda_i2t.pdf} \ \  
\includegraphics[trim = 5mm 2mm 5mm 4mm, clip, scale=0.34]{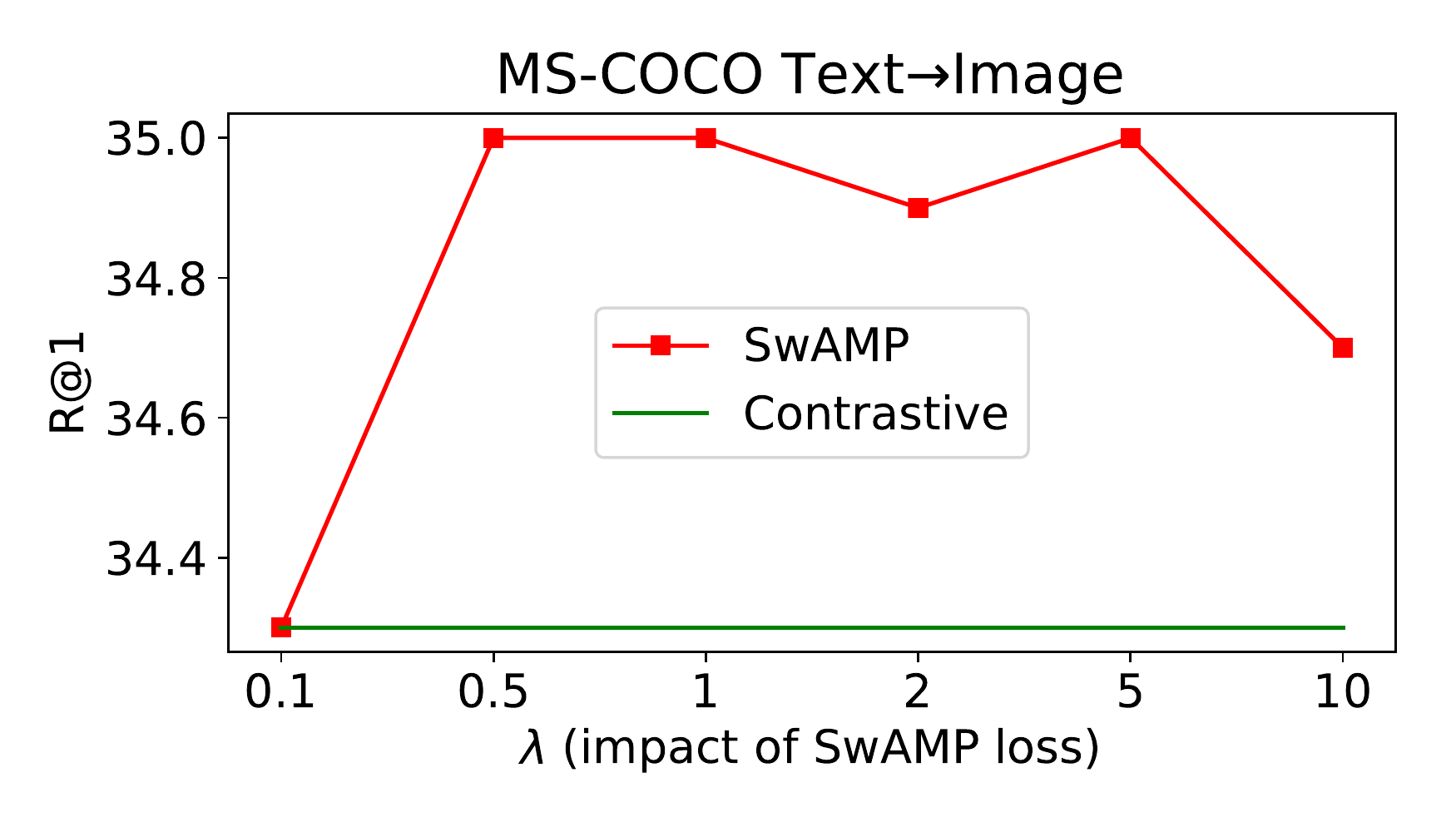}
\end{center}
\vspace{-1.5em}
\caption{(MS-COCO) Impact of the SwAMP loss ($\lambda$). 
}
%\vspace{-1.0em}
\label{appfig:coco_impact_of_lambda}
\end{figure}
%%%%

%%%%
\begin{figure}[t!]
%\vspace{-1.5em}
\begin{center}
%
%\begin{subfigure}[b]{0.9\textwidth}
\centering
\includegraphics[trim = 10mm 2mm 5mm 4mm, clip, scale=0.35]{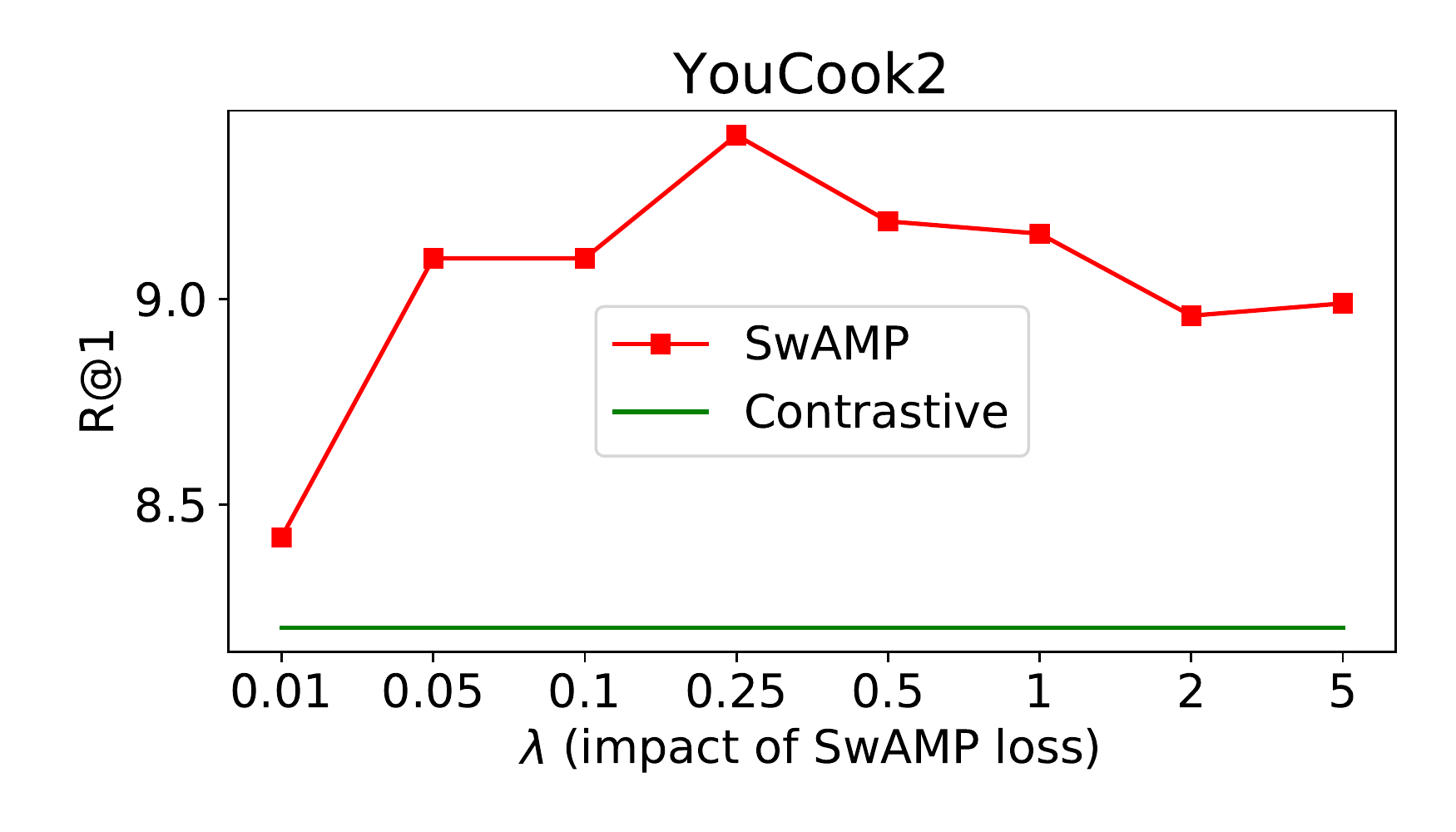}
\end{center}
\vspace{-1.5em}
\caption{(YouCook2) Impact of the SwAMP loss ($\lambda$). 
}
%\vspace{-1.0em}
\label{appfig:yc2_impact_of_lambda}
\end{figure}
%%%%

%%%%
\begin{figure}[t!]
%\vspace{-1.5em}
\begin{center}
%
%\begin{subfigure}[b]{0.9\textwidth}
\centering
\includegraphics[trim = 5mm 2mm 5mm 4mm, clip, scale=0.34]{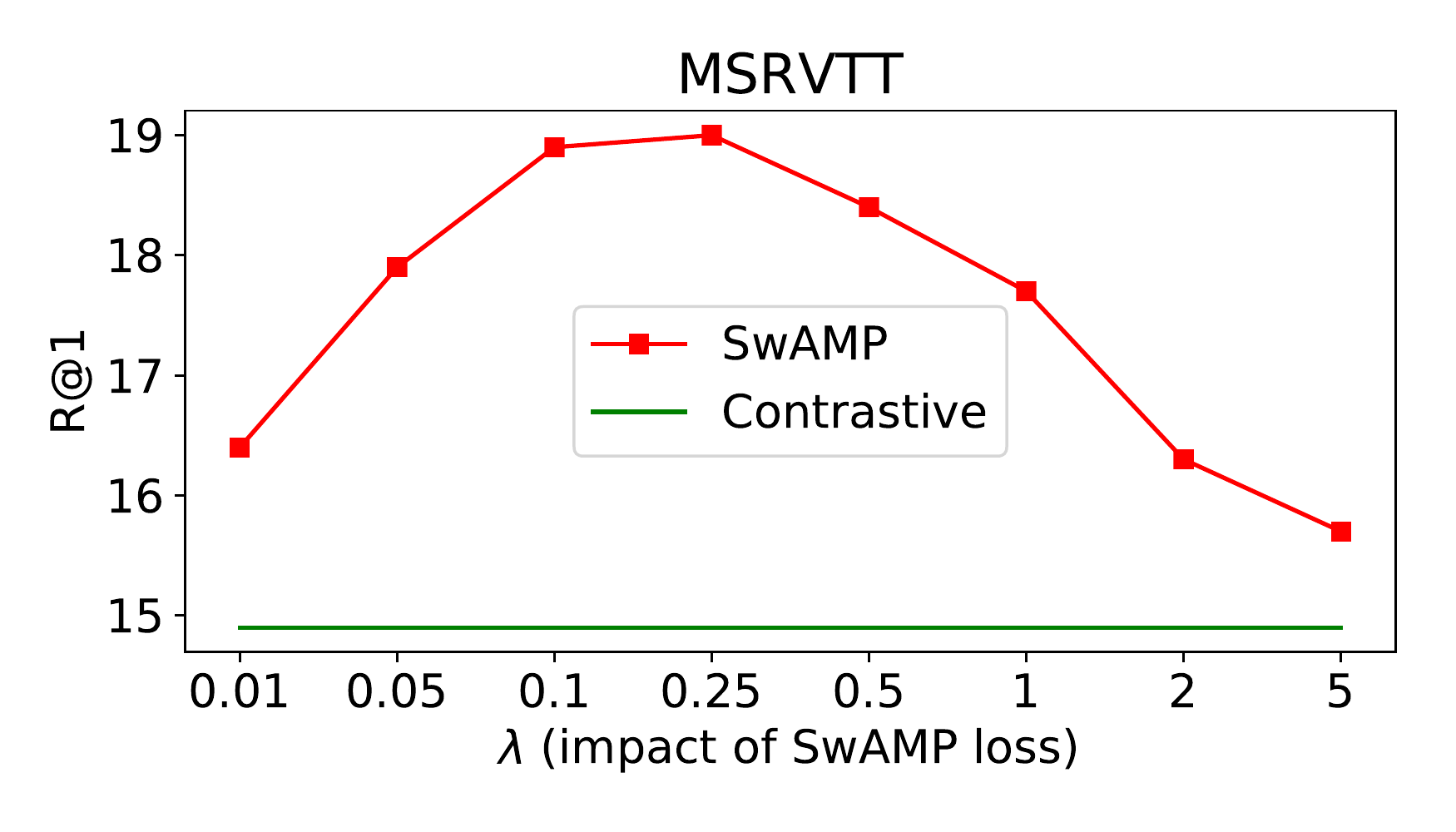}
\end{center}
\vspace{-1.5em}
\caption{(MSRVTT) Impact of the SwAMP loss ($\lambda$). 
}
%\vspace{-1.0em}
\label{appfig:msrvtt_impact_of_lambda}
\end{figure}
%%%%

%%%%
\begin{figure}%[t!]
%\vspace{-1.5em}
\begin{center}
%
%\begin{subfigure}[b]{0.9\textwidth}
\centering
\includegraphics[trim = 5mm 2mm 5mm 4mm, clip, scale=0.34]{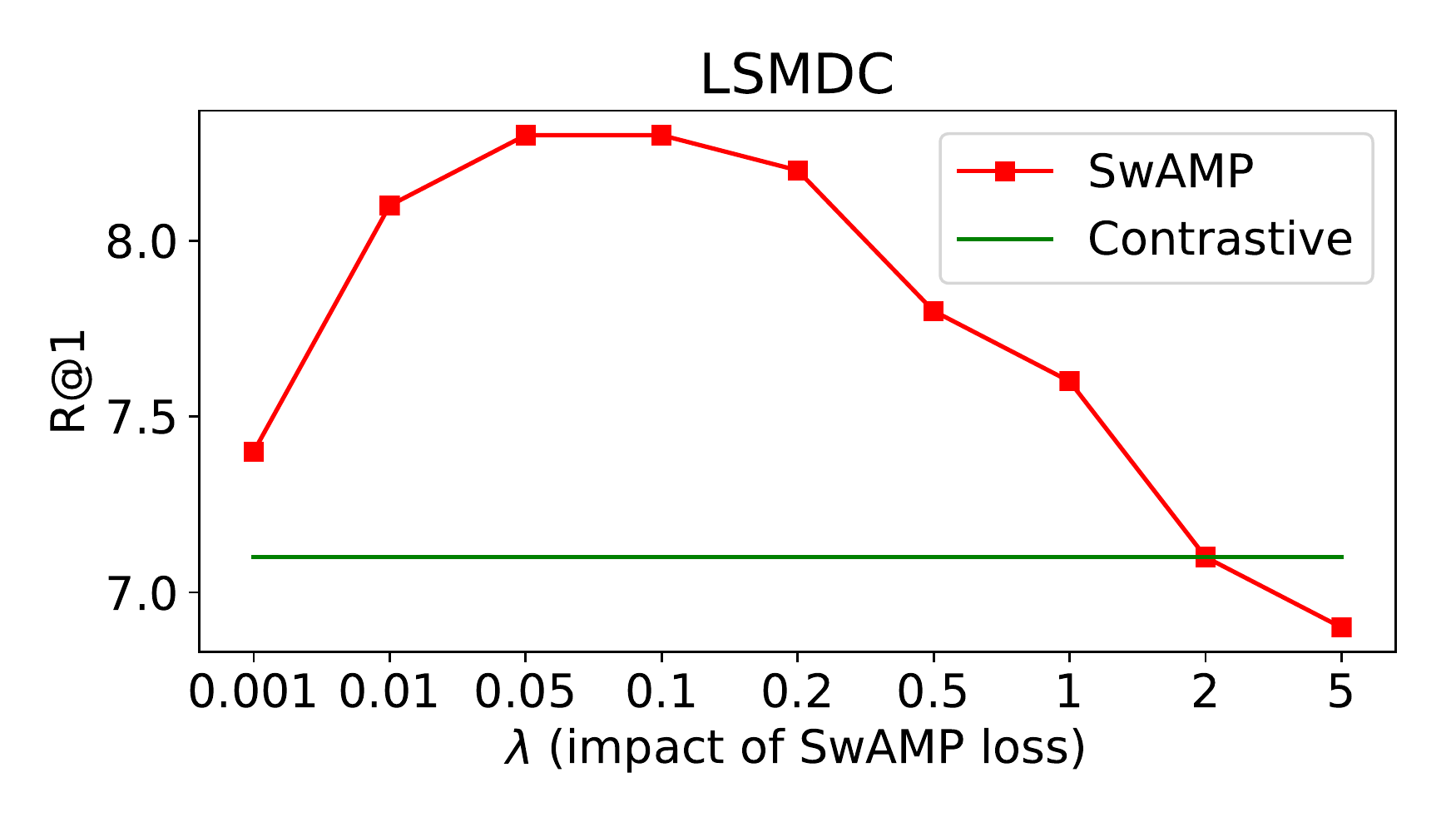}
\end{center}
\vspace{-1.5em}
\caption{(LSMDC) Impact of the SwAMP loss ($\lambda$). 
}
%\vspace{-1.0em}
\label{appfig:lsmdc_impact_of_lambda}
\end{figure}
\section{Text-based Video Retrieval}\label{appsec:video_text}

%For the real-world problems, 
We consider the text-to-video retrieval task where the goal is to find the most relevant video clip for a given natural language text query. We consider three datasets for this task: i) \textbf{YouCook2}~\citep{yc2} of cooking videos and instructions, ii) \textbf{MSR-VTT}~\citep{msrvtt} of generic videos and captions from YouTube, and iii) \textbf{LSMDC}~\citep{lsmdc} of movie clips and subtitles. All these datasets provide pairs of video clip and its text description, forming a multi-modal paired data format $(text,video)$ which conforms to our SwAMP framework. 

For the raw text/video features and the feature extractor networks, as well as the training/test protocols, we follow the methods in~\citep{howto100m}. Whereas the details of the datasets and experimental setups are described in the subsequent sections, the  features are specifically built by the following procedures. 
First, the raw features are obtained by the pretrained networks: (a) raw video features (4096D) are  concatenation of frame-level and video-level features extracted from the pretrained 2D/3D CNNs (the ImageNet pre-trained Resnet-152~\citep{resnet152} for 2D features and the Kinetics~\citep{kinetics} pre-trained ResNeXt-101 16-frame model~\citep{resnext101} for 3D features), (b) raw text features (4096D) are the GoogleNews pre-trained
word2vec embeddings~\citep{word2vec} for the pre-processed transcribed video narrations with the common stop words removed. Then the feature extractor networks $\phi^{video}(\cdot)$ and $\phi^{text}(\cdot)$ transform these raw features into 4096D features by the sigmoid-gated linear transform where the gating functions are two-layer linear networks~\citep{gated_linear}. We fix the raw features and train only the latter sigmoid-gated networks, which comprise about 67M parameters. 

%Given a textual description, the goal is to retrieve representative video clips from a large pool of videos. We evaluate our learned embedding using the standard recall metrics R@1, R@5, R@10 and the median rank (Median R). We provide experimental results for the following domain-specific video description datasets.

Following~\citep{howto100m}, there are two training strategies: i) No-pretraining (No-PT) where the feature extraction networks are randomly initialized, and the training is done on the training split of the dataset, and ii) Pretraining (PT) where the feature extractors are first pretrained on the large-scale HowTo100M dataset~\citep{howto100m}, and finetuned on the target dataset. In~\citep{howto100m}, they adopt the contrastive (triplet) loss for training the feature extractors. Although we also compare our approach with the state-of-the-arts, 
% \footnote{Although there was more recent approach from the same authors~\citep{howto100m_mil}, we have not included it since they formed a zero-shot transfer problem, simply applying the HowTo100M-pretrained model without model fine-tuning on the target dataset. % (aka zero-shot transfer), thus 
% That is, the performance is dominantly
% influenced by the enormous dataset (HowTo100M) rather than the loss function employed.}, 
the main focus in this experiment is to demonstrate the performance improvement achieved by the proposed SwAMP loss against vanilla contrastive learning. The SwAMP hyperparameter $\lambda$, the weight/impact of the SwAMP loss against the contrastive loss  is chosen as $\lambda=0.25$ for all three datasets, except the LSMDC-PT case for which  $\lambda=0.1$. 
 %empirically chosen. % (weight for the contrastive loss is always 1.0). We use $\lambda_{swamp}=0.25$ for all three datasets, except the LSMDC PT case for which  $\lambda_{swamp}=0.1$. 
We also choose temperature in softmax $\tau=0.25$, entropic regularization trade-off in SK $\eta=5.0$, the number of classes $K=500$, and the queue size $2,048$ for the SwAMP. The other learning hyperparameters common in SwAMP and contrastive losses are not changed from~\citep{howto100m}. %, and summarized in Supplement.

%\st{Unlike the synthetic data experiment, training the model with the SwAMP loss alone did not learn the model properly. This can be originated from the high dimensionality of the embedded space and the overfitted feature extractor: if the (initial) class prediction is incorrect, the model  tends to adapt to it faithfully anyway, learning embeddings that conform to the wrong clustering of data points. %if the initial embeddings are poor, then class predictions are incorrect, leading to wrong clustering of data points, which exacerbate the situation preventing the models from learning better embeddings, the vicious cycle. 
%To alleviate this issue, we combine the SwAMP loss and the contrastive loss by the weighted sum, where the latter enforces direct inner product alignment in the latent space, and helps preventing the model from being adapted to wrong clustering.}

%%%%%%%%%%%%%%
%\subsubsection{YouCook2} %~\citep{yc2}}
\textbf{YouCook2.} 
%cooking videos in YouCook2~\citep{yc2}
This cooking video dataset collected from YouTube, contains 89 recipes and 14K video clips %that are 
annotated with textual descriptions from paid human workers. The test data are formed by taking 3.5K clips from the validation set, and the test set comprises of $3,350$ pairs. The retrieval performance metrics are recall-at-$k$ (R@k) with $k=1,5,10$ and the median rank (Med-R). Hence, the random guess attains R@1$=0.03\%$ Med-R=$1,675$. 
%YC2: temperature in softmax $\tau=0.25$, entropic regularization trade-off in SK $\lambda=5.0$, impact of the SwAMP loss is $0.25$, and $K=500$, queue size is 2048. 
%Number of test samples??
The results are summarized in Table~\ref{apptab:yc2}. In the bottom four rows, we see the performance improvement achieved by the proposed SwAMP against the contrastive loss~\citep{howto100m}. For both training strategies, No PT (random model initialization) and PT (initialized with the HowTo100M-pretrained model), our SwAMP improves the retrieval performance significantly (e.g., about $12\%$ reduction in Median Rank for the No PT case). SwAMP also outperform the %As also reported in~\citep{howto100m}, the 
CCA baseline FV-CCA~\citep{fv_cca}. %, %with the features described as above are compared. 
% as well as the model that is trained with the sophisticated multiple-instance learning criteria~\citep{howto100m_mil}\footnote{We note that this comparison may not be fair since they utilize extra assumption about the data: the video clips from the same video footage that are temporally adjacent are considered and exploited as positive/relevant to the text query. We do not make use of such property.}.

%%%% 
\begin{table}[t!]
%\vspace{-2.5em}
\caption{%Performance improvement achieved by the proposed SwAMP against the contrastive learning 
Text-video retrieval results on YouCook2. %constrastiv learning~\citep{howto100m} (denoted by Contrastive). We also include the other approaches.
%The improved scores of SwAMP over contrastive are boldfaced. % for the two training strategies, No PT and PT. 
}
\vspace{+0.3em}
\centering
%\vspace{-0.8em}
%\vskip 0.05in
%\begin{scriptsize}
\begin{footnotesize}
%\begin{small}
%\begin{sc}
\centering
\scalebox{0.95}{
\begin{tabular}{lcccc}
\toprule
Methods & R@1 $\uparrow$ & R@5 $\uparrow$ & R@10 $\uparrow$ & Med-R $\downarrow$
\\ \hline
Random\Tstrut & 0.03 & 0.15 & 0.3 & 1675 \\
FV-CCA%~\citep{fv_cca} 
& 4.6 & 14.3 & 21.6 & 75 \\
\hline
%
% MIL~\citep{howto100m_mil}\Tstrut & 6.1 & 17.3 & 24.8 & 46 \\
% \hline
%
Contrastive (No PT)\Tstrut & 4.2 & 13.7 & 21.5 & 65 \\
%
%Contrastive (No PT) $+$ 
SwAMP (No PT) & ${\bf 4.8}$ & ${\bf 14.5}$ & ${\bf 22.5}$ & ${\bf 57}$ \\
\hline
Contrastive (PT)\Tstrut & 8.2 & 24.5 & 35.3 & 24 \\
%
%Contrastive (PT) $+$ 
SwAMP (PT) & ${\bf 9.4}$ & ${\bf 24.9}$ & 35.3 & ${\bf 22}$ \\
\bottomrule
\end{tabular}
}
%\end{sc}
\end{footnotesize}
%\end{scriptsize}
%\end{small}
\label{apptab:yc2}
%\vspace{-1.0em}
\end{table}
%%%%

%%%%%%%%%%%%%%
%\subsubsection{MSRVTT}
\textbf{MSRVTT.}  
This generic video-text dataset~\citep{msrvtt} collected from YouTube contains videos of specific categories including music, sports, and movie. There are 200K video-caption pairs obtained by human annotation. We follow the retrieval training/test protocol of~\citep{jsfusion,howto100m}. The test set consists of 1K pairs. 
%MSRVTT: (the same as YC2) temperature in softmax $\tau=0.25$, entropic regularization trade-off in SK $\lambda=5.0$, impact of the SwAMP loss is $0.25$, and $K=500$, queue size is 2048. 
As reported in Table~\ref{apptab:msrvtt}, our SwAMP loss improves the performance over the contrastive learning significantly for both no-pretraining and pretraining cases: about $24\%$ in R@1 in the No PT case, and $27\%$ in the PT case. Furthermore, the SwAMP outperforms with large margin the state-of-the-arts: C+LSTM+SA+FC7~\citep{c_lstm}, VSE-LSTM~\citep{vse_lstm}, Temporal Tessellation~\citep{temporal_tessellation}, CT-SAN~\citep{ct-san}, and JSFusion~\citep{jsfusion}.  %fusion approach~\citep{jsfusion} with large margin. 

%%%% 
\begin{table}[t!]
%\vspace{-2.5em}
\caption{%Performance improvement achieved by SwAMP 
Text-video retrieval results on MSRVTT. 
}
\vspace{+0.3em}
\centering
%\vspace{-0.8em}
%\vskip 0.05in
%\begin{scriptsize}
\begin{footnotesize}
%\begin{small}
%\begin{sc}
\centering
\scalebox{0.95}{
\begin{tabular}{lcccc}
\toprule
Methods & R@1 $\uparrow$ & R@5 $\uparrow$ & R@10 $\uparrow$ & Med-R $\downarrow$
\\ \hline
Random\Tstrut & 0.1 & 0.5 & 1.0 & 500 \\
C+LSTM+SA+FC7%~\citep{c_lstm} 
& 4.2 & 12.9 & 19.9 & 55 \\
VSE-LSTM%~\citep{vse_lstm} 
& 3.8 & 12.7 & 17.1 & 66 \\
SNUVL%~\citep{snuvl} 
& 3.5 & 15.9 & 23.8 & 44 \\
Temporal Tessellation%~\citep{temporal_tessellation} 
& 4.7 & 16.6 & 24.1 & 41 \\
CT-SAN%~\citep{ct-san} 
& 4.4 & 16.6 & 22.3 & 35 \\
JSFusion%~\citep{jsfusion} 
& 10.2 & 31.2 & 43.2 & 13 \\
\hline
%
%Contrastive (zero-shot transfer)
% MIL~\citep{howto100m_mil}\Tstrut & 7.5 & 21.2 & 29.6 & 38 \\
% \hline
%
Contrastive (No PT)\Tstrut & 12.1 & 35.0 & 48.0 & 12 \\
%
%Contrastive (w/o PT) $+$ 
SwAMP (No PT) & ${\bf 15.0}$ & ${\bf 38.5}$ & ${\bf 50.3}$ & ${\bf 10}$ \\
\hline
Contrastive (PT)\Tstrut & 14.9 & 40.2 & 52.8 & 9 \\
%
%Contrastive (w/ PT) $+$ 
SwAMP (PT) & ${\bf 19.0}$ & ${\bf 42.4}$ & ${\bf 55.2}$ & ${\bf 8}$ \\
\bottomrule
\end{tabular}
}
%\end{sc}
\end{footnotesize}
%\end{scriptsize}
%\end{small}
\label{apptab:msrvtt}
%\vspace{-1.5em}
\end{table}
%%%%

%%%%%%%%%%%%%%
%\subsubsection{LSMDC}
\textbf{LSMDC.} 
The LSMDC~\citep{lsmdc}\footnote{\url{https://sites.google.com/site/describingmovies/lsmdc-2016/movieretrieval}} is a dataset of movie video clips, comprised of 101K video-caption pairs. The captions are collected either from the movie scripts or the audio descriptions. The test set contains 1K pairs. For this dataset, we use the SwAMP hyperparameter (impact of the SwAMP loss against the contrastive loss) $\lambda=0.1$ for the PT case. The results are shown in Table~\ref{apptab:lsmdc}. Similar to the other two datasets, our SwAMP is consistently better than the contrastive learning (about $7\sim 9\%$ in Median Rank).

%%%% 
\begin{table}[t!]
%\vspace{-2.5em}
\caption{%Performace improvement achieved by SwAMP 
Text-Video retrieval results on LSMDC.}
\vspace{+0.3em}
\centering
%\vspace{-0.8em}
%\vskip 0.05in
%\begin{scriptsize}
\begin{footnotesize}
%\begin{small}
%\begin{sc}
\centering
\scalebox{0.95}{
\begin{tabular}{lcccc}
\toprule
Methods & R@1 $\uparrow$ & R@5 $\uparrow$ & R@10 $\uparrow$ & Med-R $\downarrow$
\\ \hline
Random\Tstrut & 0.1 & 0.5 & 1.0 & 500 \\
C+LSTM+SA+FC7%~\citep{c_lstm} 
& 4.3 & 12.6 & 18.9 & 98 \\
VSE-LSTM%~\citep{vse_lstm} 
& 3.1 & 10.4 & 16.5 & 79 \\
SNUVL%~\citep{snuvl} 
& 3.6 & 14.7 & 23.9 & 50 \\
Temporal Tessellation%~\citep{temporal_tessellation} 
& 4.7 & 15.9 & 23.4 & 64 \\
CT-SAN%~\citep{ct-san} 
& 4.5 & 14.1 & 20.9 & 67 \\
JSFusion%~\citep{jsfusion} 
& 9.1 & 21.2 & 34.1 & 36 \\
\hline
%
% MIL~\citep{howto100m_mil}\Tstrut & 4.0 & 9.8 & 14.0 & 137 \\
% \hline
%
Contrastive (No PT)\Tstrut & 7.2 & 18.3 & 25.0 & 44 \\
%
%Contrastive (No PT) $+$ 
SwAMP (No PT) & ${\bf 7.7}$ & ${\bf 19.3}$ & ${\bf 27.7}$ & ${\bf 40}$ \\
\hline
Contrastive (PT)\Tstrut & 7.1 & 19.6 & 27.9 & 40 \\
%
%Contrastive (w/ PT) $+$ 
SwAMP (PT) & ${\bf 8.3}$ & ${\bf 20.0}$ & ${\bf 28.9}$ & ${\bf 37}$ \\
\bottomrule
\end{tabular}
}
%\end{sc}
\end{footnotesize}
%\end{scriptsize}
%\end{small}
\label{apptab:lsmdc}
\vspace{-1.0em}
\end{table}
\section{Image-Text Retrieval}\label{appsec:img_txt}

For the image-text cross-modal retrieval task, we follow the features and protocols from the well-known {\em stacked cross attention network} (SCAN)~\citep{scan}. In their framework, each image is represented by a set of local features $V=\{v_1,\dots,v_k\}$, where $v_i\ (\in\mathbb{R}^D) = W_v f_i + b_v$ and $f_i$'s are the CNN features extracted from salient image regions detected by the Faster-R-CNN model~\citep{faster_rcnn}. The raw features $f_i$'s are fixed and $\{W_v, b_v\}$ are learnable parameters. 
The text (sentence) is also treated as a set of word features $E=\{e_1,\dots,e_n\}$, where $e_i\ (\in\mathbb{R}^D) = (h^{lr}_i + h^{rl}_i)/2$ and $h^{lr/rl}_i$ are the outputs of the bi-directional GRU~\citep{gru,bidir_rnn} with the sequence of word embeddings as input. Both the word embeddings and GRU parameters are learnable. These image/text features contain rich local information, however, one challenge is that both representations are {\em sets}, hence the number of elements ($k$ and $n$) can vary from instance to instance. 

In the original SCAN paper~\citep{scan}, they proposed a cross-modal attention model, where each local feature from one modality is transformed by the attention~\citep{transformer} with the set of local features in the other modality; e.g., $v_i$ is transformed to $attn(v_i; \{e_j\}_{j=1}^n) = $ the weighted sum of {\em values} $\{e_j\}_{j=1}^n$ with $v_i$ as a {\em query} and $\{e_j\}_{j=1}^n$ as {\em keys} (this denoted by i-t, while the other attention direction t-i can be used alternatively). Then the similarity score between image $V$ and text $E$ is defined as $pool(\{cos(v_i,attn(v_i; \{e_j\}_{j=1}^n))\}_{i=1}^K)$, where $cos(a,b)$ is the cosine similarity and $pool$ is the pooling operation, either of $AVG$ or $LSE$ (log-sum-exp). Then the triplet contrastive loss is employed. For the details, please refer to~\citep{scan}.

Note that in the SCAN, there is no succinct modality-wise embedding vector representation, but the similarity score between instances of two modalities is rather computed by highly complex attention operations. Although this is helpful for capturing the interactions between local features, computing the similarity score takes quadratic time in the number of elements (local features) in the instances. This is time consuming compared to simple dot-product of the modality-wise embedding vectors (See Table~\ref{apptab:image_text_time} for the actual running times compared with the approaches based on modality-wise feature representation). Moreover, it is not applicable to our SwAMP approach since we need to predict the class labels for each modality from modality-wise representation $\phi^{image}(V)$, $\phi^{text}(E)$. 

To have modality-wise representation, we adopt the idea of {\em induced-set attention} (ISA) from the Set Transformer~\citep{set_transformer}. Specifically, we introduce $p$ learnable prototype (query) vectors $\{q_j\}_{j=1}^p$ where $q_j\in \mathbb{R}^D$. Then we compute the attention for each query with $V$ (or $E$), i.e., $z_j = attn(q_j; \{v_i\}_{i=1}^k)$. Then we define $\phi^{image}(V) = concat(z_1,\dots,z_p)$, similarly for $\phi^{text}(E)$, where $concat$ refers to concatenation. Thus the parameters for $\phi^{image}()$ are $\{W_v, b_v\}$ and $\{q_j\}_{j=1}^p$, and the parameters for $\phi^{text}()$ are the word embeddings, GRU parameters, and $\{q_j\}_{j=1}^p$. We share the same $\{q_j\}_{j=1}^p$ for both modalities. We also have multi-head extension by computing these features multiple times and concatenating them. 
%the simple prototype-based attention module, where we essentially maintain a fixed number (denoted by $p$) of prototype vectors (of the same dimensionality as that of the embedding), then we use the prototype vectors as queries and set data $x^A$ or $x^B$ as keys/values in the attention. The transformed queries are concatenated as a ($p \cdot d$-dim vector, which is the succinct vector representation for a set instance. 
We call these modality-wise features as {\em prototype attention representation} (PAR). Note that computing PAR features has linear complexity in the number of local features (assuming $p$ is constant), and the cross-modal similarity is simply dot-product of PAR features, and can be computed in linear time (See also Table~\ref{apptab:image_text_time}).

% %%%%
% \begin{itemize}
% %
% \item SCAN
% %
% \item CosSim-PAR: Contrastive loss with PAR vector representation of a image/text set instance. 
% %
% \item CosSim-ST: Contrastive loss with Set-Transformer representation of a image/text set instance. 
% %
% \item SwAMP-PAR: SwAMP loss with PAR
% %
% \item SwAMP-ST: SwAMP loss with ST
% %
% \end{itemize}
% %%%%

%%%%%%%%%%%%%%
\subsection{Datasets and Results}

We test our approach on the popular image-text retrieval datasets, MS-COCO and Flickr30K. There are 31K images and five captions for each image in Flickr30K. MS-COCO contains $123,287$ images, where each image is annotated with five text descriptions. Following the widely-used split~\citep{karpathy,vsepp}, for the Flickr30K, we have 1K images for validation, 1K images for testing, and the rest for training. For MS-COCO, there are 5K test images (and 25K captions, five captions for each image). We also follow two standard protocols for measuring the test retrieval performance for MS-COCO: 1) using the entire 5K test images or 2) splitting the test set into 5 folds and report the average retrieval performance over the 5 folds. 

The results are summarized in Table~\ref{apptab:flickr} (Flickr) and Table~\ref{apptab:coco} (MS-COCO). We specifically highlight the comparison between the contrastive loss and our SwAMP loss with the modality-wise feature representation (Contrastive-PAR vs.~SwAMP-PAR). For the PAR features, we choose the number of prototypes $p=20$, attention weight temperature $T=0.5$, and the number of heads $H=1$ for Flickr, and $p=10, T=0.5, H=2$ for MS-COCO. For the SwAMP hyperparameters, we use the impact of SwAMP loss $\lambda=1.0$, softmax temperature $\tau=0.025$, the number of classes $K=1,000$, queue size $1,280$ for both datasets. 
As shown, the SwAMP loss performs consistently better than the contrastive loss. SwAMP also outperforms several state-of-the-arts including the recent sophisticated probabilistic embedding strategy~\citep{pcme}.

%%%% 
\begin{table*}[t!]
%\vspace{-2.5em}
\caption{%Performance improvement achieved by SwAMP 
Image-text retrieval results on Flickr30K. %(Report R@1 only??)
}
\vspace{+0.3em}
\centering
%\vspace{-0.6em}
%\vskip 0.05in
%\begin{scriptsize}
%\begin{footnotesize}
\begin{small}
%\begin{sc}
\centering
\scalebox{0.95}{
\begin{tabular}{lcccccc}
\toprule
\multirow{2}{*}{Methods} & 
\multicolumn{3}{c}{Image $\to$ Text} &  \multicolumn{3}{c}{Text $\to$ Image}
\\ \cmidrule(lr){2-4} \cmidrule(lr){5-7} 
& R@1 & R@5 & R@10 & R@1 & R@5 & R@10 \\
\hline
DAN~\citep{dan}\Tstrut & 55.0 & 81.8 & 89.0 & 39.4 & 69.2 & 79.1 \\
DPC~\citep{dpc} & 55.6 & 81.9 & 89.5 & 39.1 & 69.2 & 80.9 \\
VSE++~\citep{vsepp} & 52.9 & - & 87.2 & 39.6 & - & 79.5 \\
SCO~\citep{sco} & 55.5 & 82.0 & 89.3 & 41.1 & 70.5 & 80.1 \\
\hline
%
%SCAN i-t AVG\Tstrut & ${\bf 67.9}$ & 89.0 & 94.4 & 43.9 & 74.2 & 82.8 \\
SCAN i-t AVG\Tstrut & $67.9$ & 89.0 & 94.4 & 43.9 & 74.2 & 82.8 \\
SCAN t-i AVG & 61.8 & 87.5 & 93.7 & 45.8 & 74.4 & 83.0 \\
%
%SCAN t-i AVG + i-t LSE & 67.4 & ${\bf 90.3}$ & ${\bf 95.8}$ & 48.6 & ${\bf 77.7}$ & ${\bf 85.2}$ \\
SCAN t-i AVG + i-t LSE & 67.4 & $90.3$ & $95.8$ & 48.6 & $77.7$ & $85.2$ \\
\hline
Contrastive-PAR\Tstrut & 65.7 & 86.8 & 92.4 & 48.2 & 75.8 & 84.2 \\
%SwAMP-PAR & 67.8 & 88.5 & 94.0 & ${\bf 49.1}$ & 76.1 & 83.7 \\
SwAMP-PAR & ${\bf 67.8}$ & ${\bf 88.5}$ & ${\bf 94.0}$ & ${\bf 49.1}$ & ${\bf 76.1}$ & ${\bf 83.7}$ \\
\bottomrule
\end{tabular}
}
%\end{sc}
%\end{footnotesize}
%\end{scriptsize}
\end{small}
\label{apptab:flickr}
\end{table*}
%%%%

%%%% 
\begin{table*}[t!]
%\vspace{-2.5em}
\caption{%Performance improvement achieved by SwAMP 
Image-text retrieval results on MS-COCO. %(Report R@1 only??)
}
\vspace{+0.3em}
\centering
%\vspace{-0.6em}
%\vskip 0.05in
%\begin{scriptsize}
%\begin{footnotesize}
\begin{small}
%\begin{sc}
\centering
\scalebox{0.95}{
\begin{tabular}{lcccccccccccc}
\toprule
\multirow{3}{*}{Methods} & \multicolumn{6}{c}{5-fold (1K test images)} & \multicolumn{6}{c}{Entire (5K test images)}
\\ \cmidrule(lr){2-7} \cmidrule(lr){8-13} 
& \multicolumn{3}{c}{Image $\to$ Text} &  \multicolumn{3}{c}{Text $\to$ Image} & \multicolumn{3}{c}{Image $\to$ Text} &  \multicolumn{3}{c}{Text $\to$ Image}
\\ \cmidrule(lr){2-4} \cmidrule(lr){5-7} \cmidrule(lr){8-10} \cmidrule(lr){11-13}
& R@1 & R@5 & R@10 & R@1 & R@5 & R@10 & R@1 & R@5 & R@10 & R@1 & R@5 & R@10 \\
\hline
DPC~\citep{dpc}\Tstrut & 65.6 & 89.8 & 95.5 & 47.1 & 79.9 & 90.0 & 41.2 & 70.5 & 81.1 & 25.3 & 53.4 & 66.4 \\
VSE++~\citep{vsepp} & 64.6 & - & 95.7 & 52.0 & - & 92.0 & 41.3 & - & 81.2 & 30.3 & - & 72.4\\
GXN~\citep{gxn} & 68.5 & - & 97.9 & 56.6 & - & 94.5 & 42.0 & - & 84.7 & 31.7 & - & 74.6 \\
SCO~\citep{sco} & 69.9 & 92.9 & 97.5 & 56.7 & 87.5 & 94.8 & 42.8 & 72.3 & 83.0 & 33.1 & 62.9 & 75.5 \\
PCME~\citep{pcme} & 68.8 & - & - & 54.6 & - & - & 44.2 & - & - & 31.9 & - & - \\
\hline
SCAN i-t\Tstrut & 69.2 & 93.2 & 97.5 & 54.4 & 86.0 & 93.6 & 46.4 & 77.4 & 87.2 & 34.4 & 63.7 & 75.7 \\
SCAN t-i + i-t & 72.7 & 94.8 & 98.4 & 58.8 & 88.4 & 94.8 & 50.4 & 82.2 & 90.0 & 38.6 & 69.3 & 80.4 \\
\hline
Contrastive-PAR\Tstrut & 71.8 & 94.3 & 97.9 & 56.8 & 86.9 & 93.8 & 48.4 & 78.1 & 88.1 & 34.3 & 64.4 & 76.2 \\
%SwAMP-PAR & 72.6 & 94.6$ & 98.0$ & 57.4 & 87.6 & 94.1 & 49.7 & 79.1 & 88.3 & 35.0 & 65.1 & 76.6 \\
SwAMP-PAR & ${\bf 72.6}$ & ${\bf 94.6}$ & ${\bf 98.0}$ & ${\bf 57.4}$ & ${\bf 87.6}$ & ${\bf 94.1}$ & ${\bf 49.7}$ & ${\bf 79.1}$ & ${\bf 88.3}$ & ${\bf 35.0}$ & ${\bf 65.1}$ & ${\bf 76.6}$ \\
\bottomrule
\end{tabular}
}
%\end{sc}
%\end{footnotesize}
%\end{scriptsize}
\end{small}
\label{apptab:coco}
%\vspace{-1.0em}
\end{table*}
%%%%

When compared with the computationally expensive SCAN, SwAMP mostly outperforms SCAN except for the SCAN's best attention direction/combination choices. 
Note that SwAMP uses the simple feature aggregation strategy (PAR) to have fast and succinct modality-wise feature representation, whereas SCAN relies on the cross-modal attention similarity scoring model, which is computationally expensive. To see the computational advantage of SwAMP-PAR, we compare the actual training/test times for the two approaches in Table~\ref{apptab:image_text_time}, measured on the same machine with a single GPU (RTX 2080 Ti), Core i7 3.50GHz CPU, and 128 GB RAM. As shown, our SwAMP-PAR is about 4 times faster than SCAN for training on both datasets, while the difference becomes even more pronounced during test; SwAMP-PAR is about two orders of magnitude faster than the cross-modal attention model.

%\hl{Also add a few retrieval examples/figures...}

%%%% 
\begin{table}[t!]
%\vspace{-2.5em}
\caption{Running time comparison for SCAN (cross-modal attention) and our SwAMP-PAR. Running times (seconds) are measured on the same machine (Core i7 3.50GHz CPU, 128GB RAM, and a single GeForce RTX-2080Ti GPU). We report per-batch times for training, and entire retrieval times for test. For MS-COCO test, the running times for 5K test images are reported, where times for 1K test images averaged over 5 folds are shown in the parentheses. For SCAN, when we use features in both directions (e.g., t-i AVG + i-t LSE), the running times are roughly doubled.}
\vspace{+0.3em}
\centering
%\vspace{-0.6em}
%\vskip 0.05in
%\begin{scriptsize}
%\begin{footnotesize}
\begin{small}
%\begin{sc}
\centering
\scalebox{0.95}{
\begin{tabular}{lcccc}
\toprule
\multirow{2}{*}{Methods} & 
\multicolumn{2}{c}{\ \ \ \ Flickr30K \ \ \ \ } &  \multicolumn{2}{c}{MS-COCO}
\\ \cmidrule(lr){2-3} \cmidrule(lr){4-5} 
& Train & Test & Train & Test \\
\hline
SCAN i-t AVG\Tstrut & 
0.35 %0.3470 
& 336.9 %336.9181 
& 0.33 %0.3280 
& 9352.0 (350.3)
%9351.9934 (350.2627) 
\\
\hline
SwAMP-PAR\Tstrut & 
0.09 %0.0968 
& 3.8 %3.7806 
& 0.08 %0.0819 
& 25.9 (16.3)
%25.8951 (16.2914) 
\\
\bottomrule
\end{tabular}
}
%\end{sc}
%\end{footnotesize}
%\end{scriptsize}
\end{small}
\label{apptab:image_text_time}
\end{table}
%%%%

%Hyperparameters: We used the SwAMP loss alone, but it turned out it didn't work well (because of the chicken-egg problem, ie, initial embedding is inaccurate, and the class prediction using the inaccurate embedding is inaccurate, so vicious cycle). Instead, we combine the SwAMP and CosSim losses together with the balancing parameters $\lambda = 1.0$ for the impact of the CosSim-loss. 

% \hl{May put the following benefit over cross-modal attention?}
% %%%%
% \begin{enumerate}
% %
% \item Benefit-1: SwAMP loss requires a linear sampling complexity (unlike quadratic for the contrastive cossim loss), which results in faster convergence during training (??), and makes \hl{the SwAMP loss lead to better retrieval performance than the contrastive cossim loss.}
% %
% \item Benefit-2: On image-text retrieval benchmarks, the SwAMP loss with a simple feature aggregation leads to retrieval performance (sometimes better) comparable to that of the complex cross-modal attention similarity scoring model (SCAN). That is, \hl{the SwAMP loss with simple modality-wise feature representation has retrieval performance comparable to the complex SCAN's joint similarity scoring model, with significantly faster test retrieval time (about two orders of magnitude).}
% %
% \end{enumerate}
% %%%%

%%%%%%%%%%%%%%%%%%%%%%%%%%%%%%%%%%%%%%%%%%%%%%%%%%%%%%%%%%%%%%%%%%%%%%%%%%%%%%%
%%%%%%%%%%%%%%%%%%%%%%%%%%%%%%%%%%%%%%%%%%%%%%%%%%%%%%%%%%%%%%%%%%%%%%%%%%%%%%%
\section{Synthetic Data}\label{appsec:synth}

In this section we devise a synthetic dataset not only for performing the proof-of-concept test of our SwAMP algorithm, but also analyze the impacts of the various hyperparameters and training options in the proposed algorithm. For the former, we especially focus on the retrieval performance improvement achieved by our SwAMP compared to the contrastive loss or its popular variants (e.g., online hard-example mining loss). 

The dataset is constructed by the following procedure: We randomly generate $20$ Gaussians in $\mathbb{R}^5$, each of which is considered to represent a {\em semantic class}. For each Gaussian (class), we sample a latent vector $z\in\mathbb{R}^5$, and a pair of instances $(x^A\in\mathbb{R}^{100}, x^B\in\mathbb{R}^{100})$ is then generated by $x^A=f_A(z)$ and $x^B = f_B(z)$ where $f_A$ and $f_B$ are randomly initialized fully-connected DNNs with two hidden layers of 50 units. We generate 500 pairs for each class that leads to $10,000$ data pairs, and split them into 7000/1000/2000 train/validation/test sets. The validation recall-at-1 (R@1) performance is evaluated at every training epoch, and the model at the epoch with the best validation performance is selected as the final model. Note that during training we only use the paired data $(x^A, x^B)$ with the semantic class labels hidden to the training algorithms. 

For training, we adopt the embedding networks $\phi^A(x^A)$ and $\phi^B(x^B)$ as fully-connected neural nets with two hidden layers of 50 units. The embedding dimension is chosen as $5$. We train the model with this same network architecture, using the contrastive loss and our SwAMP loss. For both loss functions, the batch size is 128, and the Adam optimizer~\citep{adam} is used with learning rate $10^{-3}$, and the maximum epoch is 100. 

For the contrastive loss, we adopt the (online) hard-example mining with the margin parameter $\alpha=0.1$. % in (\ref{appeq:contrastive_loss}). 
For the SwAMP loss, the defaults parameters are as follows: temperature $\tau=0.01$ for the softmax classifier, the reciprocal impact of the max-entropy regularizer for the Sinkhorn-Knopp $\eta=1/0.05$ (i.e., we add the entropic regularizer with the weight $\eta^{-1} = 0.05$ to the objective of the OT problem. Also, by default, we choose the number of classes $K=1000$ and the queue size $1,280$, 10 times the batch size (and greater than $K$). %These are the default values for the hyperparameters, and refer to below the results for other hyperparameter values. 
For both loss functions, the embedding networks are initialized randomly. 

For test, we perform the cross-modal retrieval task $x^A \to x^B$, treating each $x^A$ in the test set as a query, retrieving $x^B$ from the test set. There are two ways to define the retrieval error: i) {\em pair-based} which treats the retrieved $x^B$ as a correct retrieval only if the query $x^A$ and the retrieved $x^B$ are found as a pair in the data, and ii) {\em class-based} which compares only the classes of the query $x^A$ and the retrieved $x^B$. Hence the pair-based error is more strict than the class-based since it counts only the data item that appears in the data as correct retrieval, without comparing the semantic classes of the retrieved item and the query. 

%%%% 
\begin{table}[t!]
%\vspace{-2.5em}
\caption{Retrieval results on the synthetic data.}
\vspace{+0.3em}
\centering
%\vspace{-0.6em}
%\vskip 0.05in
%\begin{scriptsize}
\begin{footnotesize}
%\begin{small}
%\begin{sc}
\centering
\scalebox{0.95}{
\begin{tabular}{llcccc}
\toprule
Error type & Method & R@1 $\uparrow$ & R@5 $\uparrow$ & R@10 $\uparrow$ & Med-R $\downarrow$
\\ \hline
\multirow{2}{*}{Pair-based} & Contrastive\Tstrut & 84.10 & 98.60 & 99.55 & 1 \\
& SwAMP & ${\bf 90.80}$ & ${\bf 99.95}$ & ${\bf 100.0}$ & ${\bf 1}$ \\
\hline
\multirow{2}{*}{Class-based} & Contrastive\Tstrut & 91.60 & 99.70 & 99.90 & 1 \\
& SwAMP & ${\bf 95.70}$ & ${\bf 99.95}$ & ${\bf 100.0}$ & ${\bf 1}$ \\
\bottomrule
\end{tabular}
}
%\end{sc}
\end{footnotesize}
%\end{scriptsize}
%\end{small}
\label{apptab:synth}
\vspace{-1.0em}
\end{table}
%%%%

%%%%%%%%%%%%%%
\subsection{Ablation study on hyperparameters} 

%\hl{Include the impact-of-lambda study in the ablation... Note We only had results on  $1/\lambda=0$. Show results when we gradually increases $1/lambda$ to infty (corresponding to contrastive loss).}

There are several hyperparameters in our SwAMP model, and we have conducted several ablation-type study on the impacts of the hyperparameters. The hyperaparameters that are deemed to be the most critical are: i) the number of classes $K$, ii) the size of the queues, iii) initialization of the feature extraction networks (either random initialization or pretrained one with the contrastive loss), iv) entropic regularization trade-off $\eta$ in Sinkhorn-Knopp, and v) the soft/hard cluster assignment after OT clustering. 

%and iv) whether the logit term is used or not in the cost function of the OT. The experimental results on these variations are summarized below. %\footnote{In the supplement, we also report results on other impacts including the entropic regularization trade-off $\lambda$ and the soft/hard cluster assignment after OT clustering).}. 

%\hl{Impact of lambda, the impact of SwAMP loss; Currently lambda is 1e6}

\begin{figure}[t!]
%\vspace{-1.5em}
\begin{center}
%
%\begin{subfigure}[b]{0.9\textwidth}
\centering
\includegraphics[trim = 5mm 2mm 5mm 4mm, clip, scale=0.33]{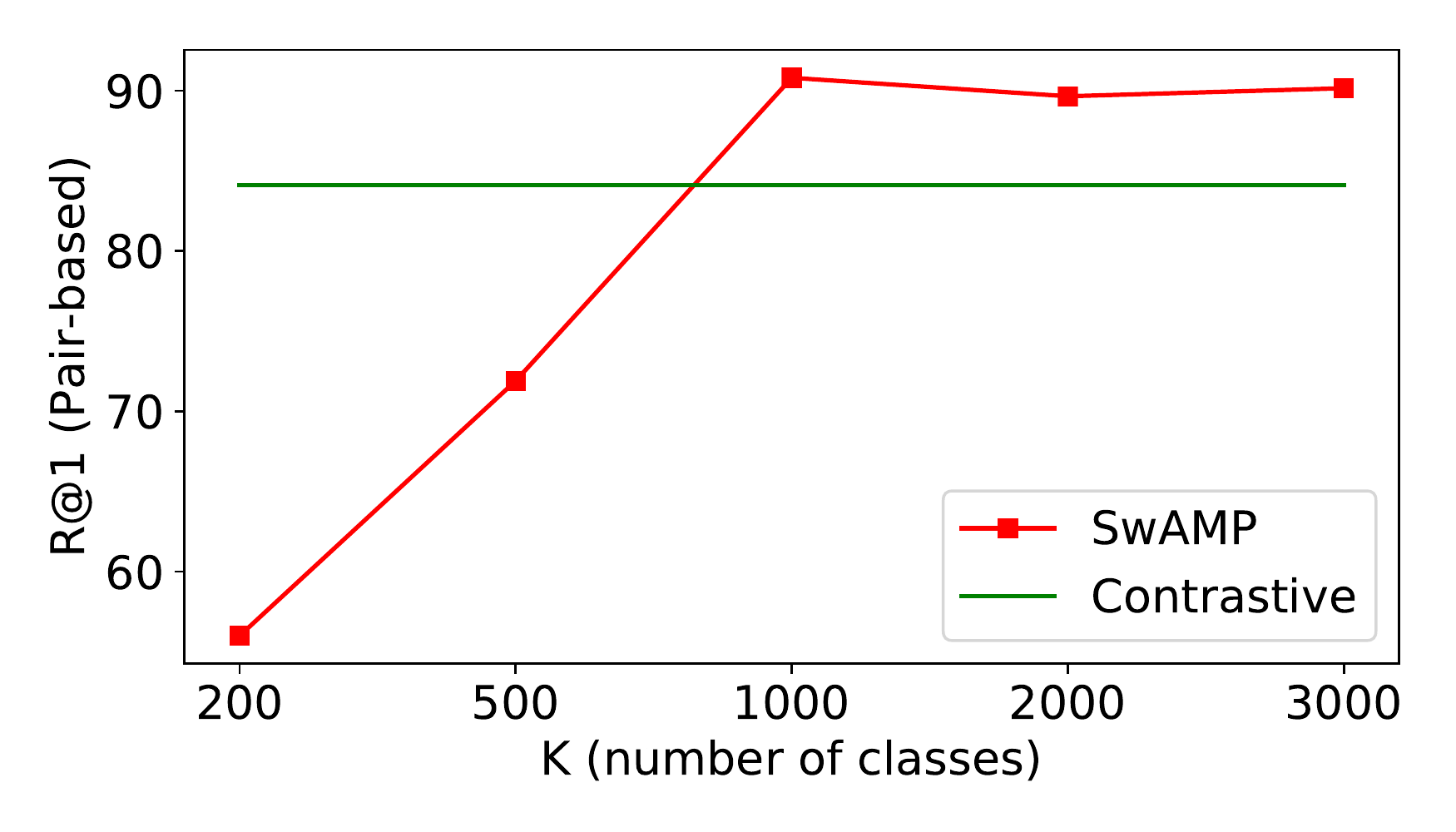} \ \ 
\includegraphics[trim = 5mm 2mm 5mm 4mm, clip, scale=0.33]{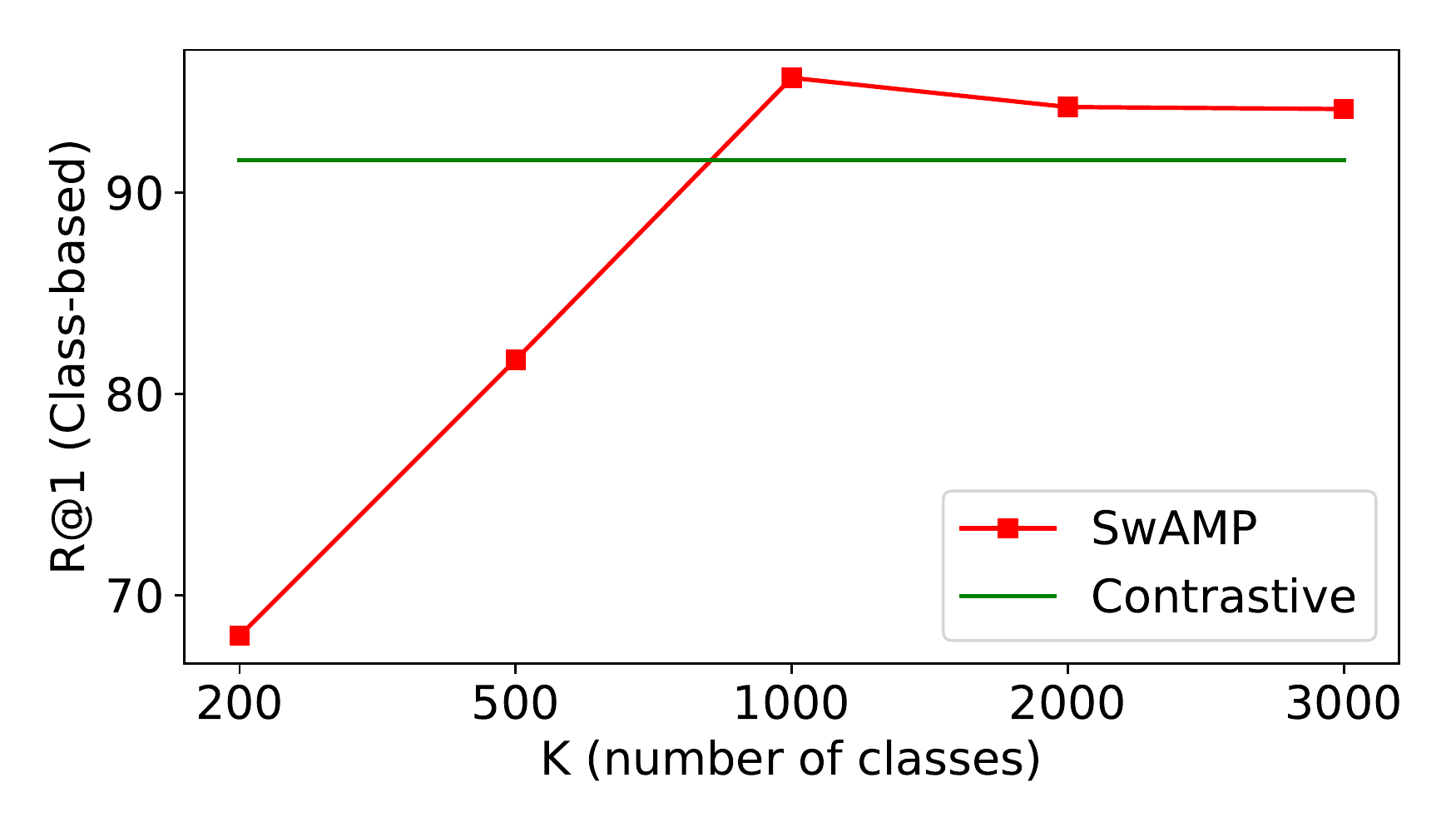}
\end{center}
\vspace{-1.5em}
\caption{(Synthetic data) Impact of the number of classes ($K$). 
}
%\vspace{-1.0em}
\label{appfig:impact_of_K}
\end{figure}

%%
\textbf{Number of classes ($K$).} We vary the number of classes $K$ for  200, 500, 1000, 2000, 3000, and record the R@1 scores for both pair and class based error types for our SwAMP model. The results are shown in Fig.~\ref{appfig:impact_of_K}.
We see that allowing more clusters improves the performance. %, which implies that the SwAMP promotes the instance discrimination, the widely successful pretext task in self-supervised learning~[CITE].
However, once $K$ is around $1000$ or greater than $1000$, there is no significant benefit of increasing $K$. This implies that SwAMP does not merely do instance discrimination, but seeks for grouping/clustering of similar instances. 
Although we did not include it in the figure, having $K=20$, i.e., the true number of semantic classes, yielded poor performance (worse than $K=200$). This means that it is very difficult to expect that the model would discover the underlying semantic classes correctly. 

%%%%
%%%%

\begin{figure}[t!]
%\vspace{-1.5em}
\begin{center}
%
%\begin{subfigure}[b]{0.9\textwidth}
\centering
\includegraphics[trim = 5mm 2mm 5mm 4mm, clip, scale=0.33]{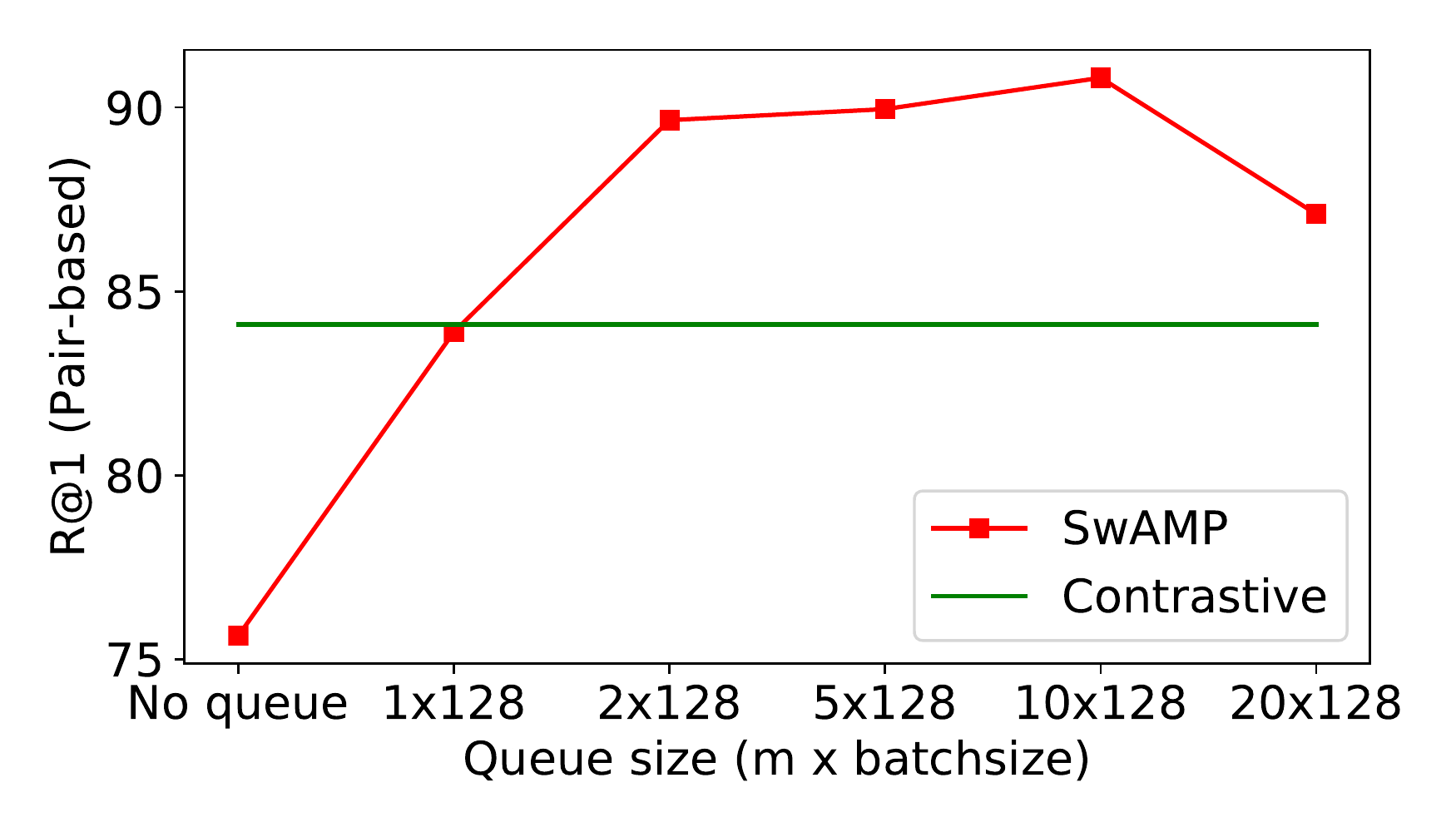} \ \ 
\includegraphics[trim = 5mm 2mm 5mm 4mm, clip, scale=0.33]{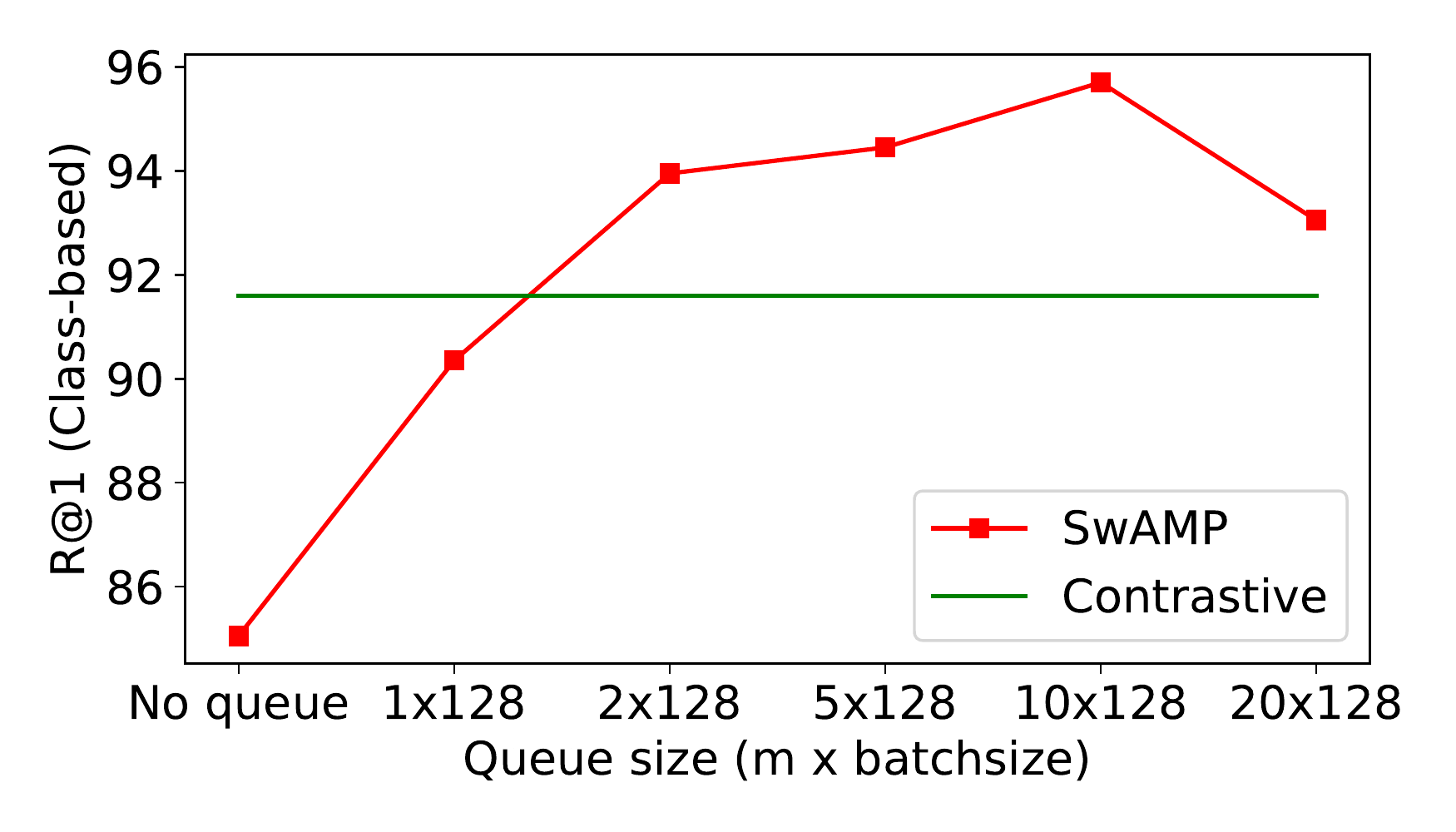}
\end{center}
\vspace{-1.5em}
\caption{(Synthetic data) Impact of the size of the queues. 
}
%\vspace{-1.0em}
\label{appfig:impact_of_queue_size}
\end{figure}

%%
\textbf{Size of queues.} Another important hyperparameter is the size of the queues, where the OT clustering is performed on the latest features that are stored in the queues. In addition to the default queue size $1280 = 10 \times 128$ (batch size), we try with different queue sizes $\{0,1,2,5,20\} \times 128$. Note that the OT clustering is performed on the union of the features in the queue and the current batch, hence zero queue size implies that we only use the current batch for OT clustering. The results are reported in Fig.~\ref{appfig:impact_of_queue_size}. 
As shown, increasing the queue sizes accordingly improves the performance, where with the queue size of two times the batch size outperforms the contrastive loss. Also, not using the queues (``No queue'') resulted in poor performance, signifying the importance of using the queues. Interestingly, too large queue size ($20 \times$) deteriorates the performance, which might be explained by the negative effects of the stale features obtained several iterations ago from the old feature extractor networks. This suggests the trade off of the queue size: too small queue size does not generalize well to the clustering of entire data, while too large queue size can be harmful due to the inconsistent stale features. 

%%%%
%%%%

\begin{figure}[t!]
%\vspace{-1.5em}
\begin{center}
%
%\begin{subfigure}[b]{0.9\textwidth}
\centering
\includegraphics[trim = 5mm 2mm 5mm 4mm, clip, scale=0.33]{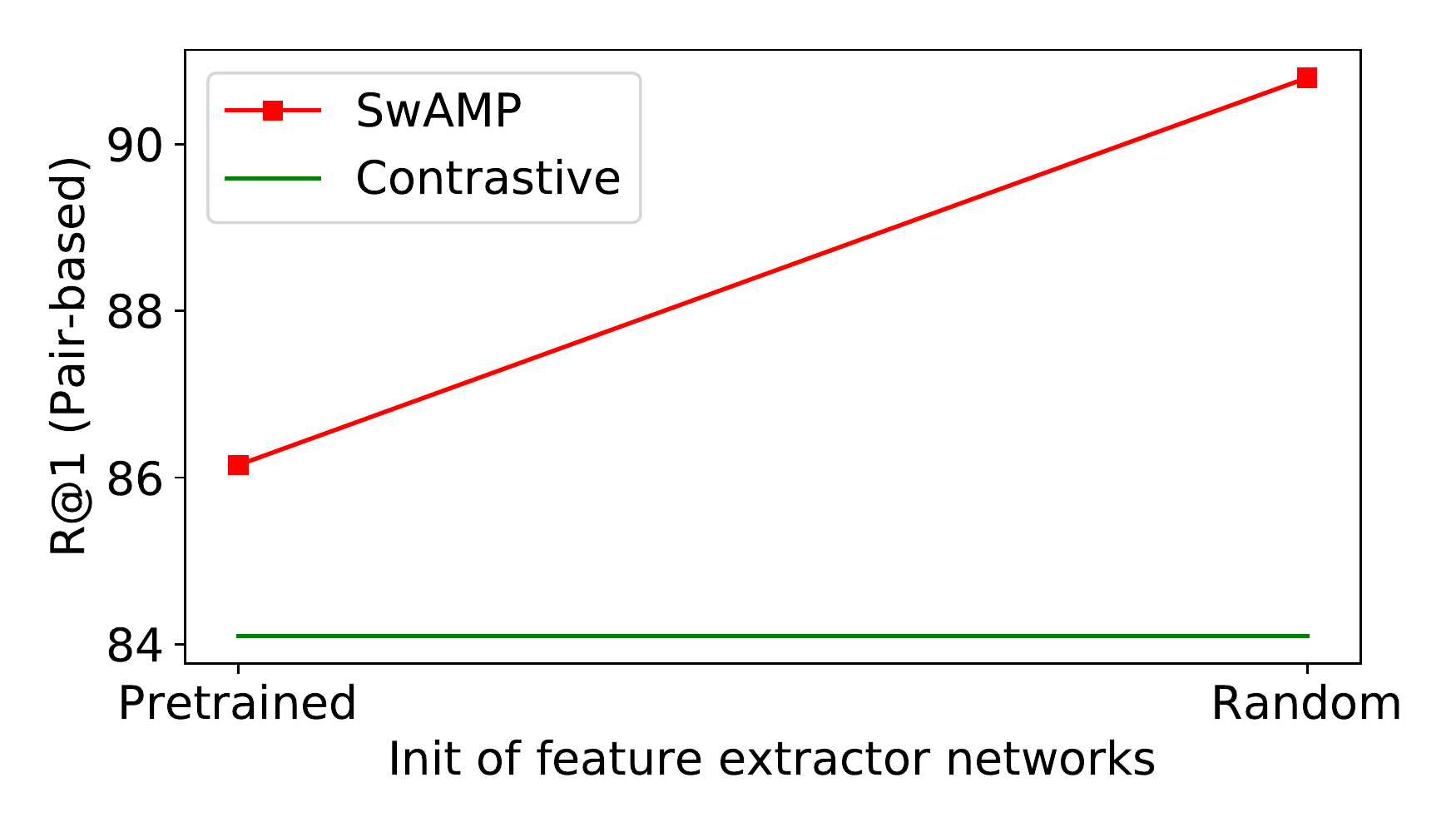} \ \ 
\includegraphics[trim = 5mm 2mm 5mm 4mm, clip, scale=0.33]{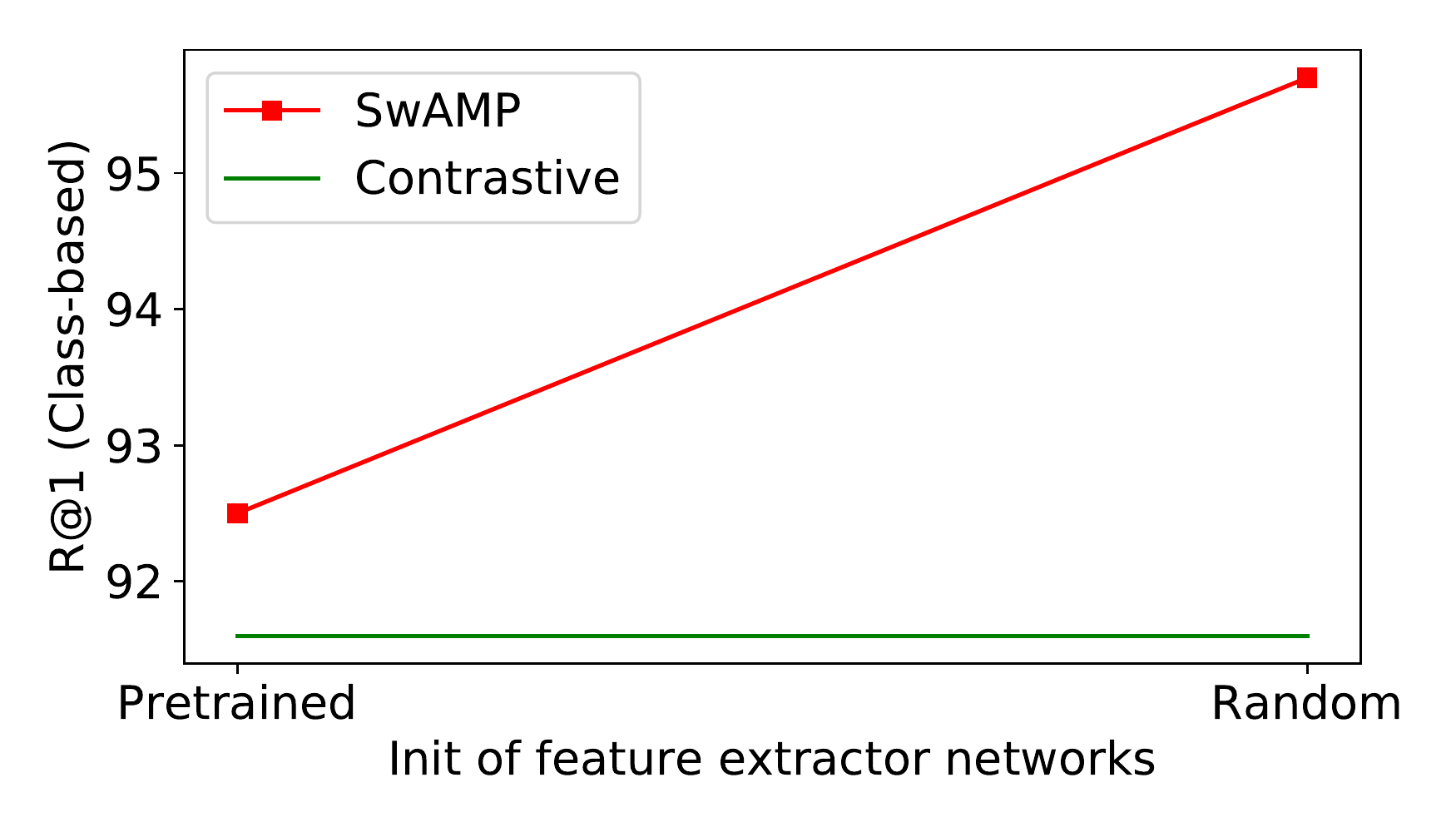}
\end{center}
\vspace{-1.5em}
\caption{(Synthetic data) Impact of the initialization of feature extractor networks. 
}
%\vspace{-1.0em}
\label{appfig:impact_of_init}
\end{figure}

%%
\textbf{Initialization of feature extractor networks.} In our default setup, the feature extractor networks $\phi^A(\cdot)$ and $\phi^B(\cdot)$ are initialized randomly. Now we test the performance of the SwAMP when the feature extractor networks are initialized from the pretrained ones by the contrastive loss training. We initially expected that this warm-start training may expedite the training with the SwAMP loss, however, as the results in Fig.~\ref{appfig:impact_of_init} indicates, it does not outperform the random initialization although the warm-start is still better than contrastive loss training. This may imply that the SwAMP loss defines a very different loss landscape from the contrastive loss, and the contrastive-loss optimized model may lie at the region far from the optima of the SwAMP loss, thus the warm-start even hinders convergence to the SwAMP optima. 

%%%%
%%%%

\begin{figure}[t!]
%\vspace{-1.5em}
\begin{center}
%
%\begin{subfigure}[b]{0.9\textwidth}
\centering
\includegraphics[trim = 5mm 2mm 5mm 4mm, clip, scale=0.33]{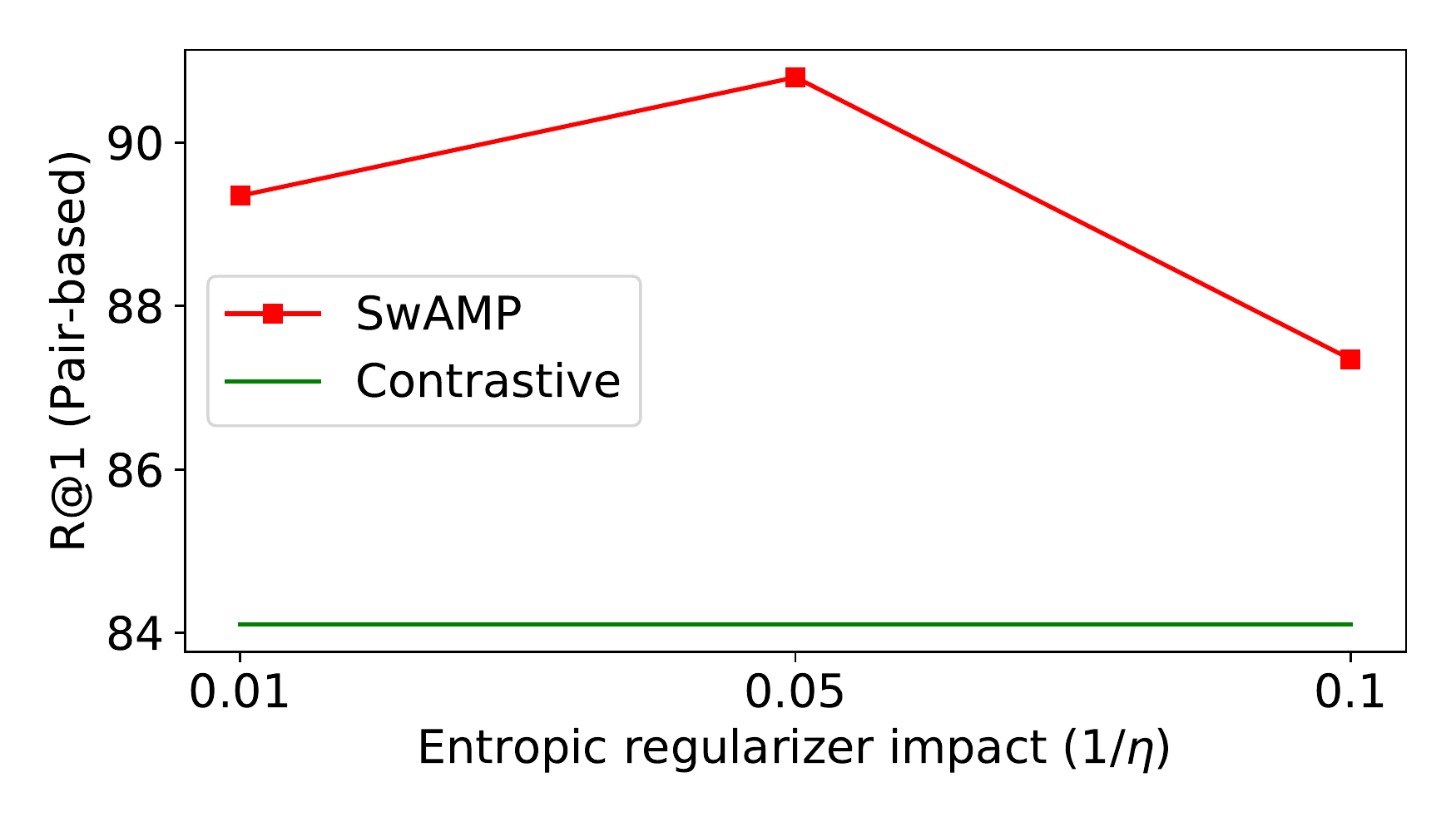} \ \ 
\includegraphics[trim = 5mm 2mm 5mm 4mm, clip, scale=0.33]{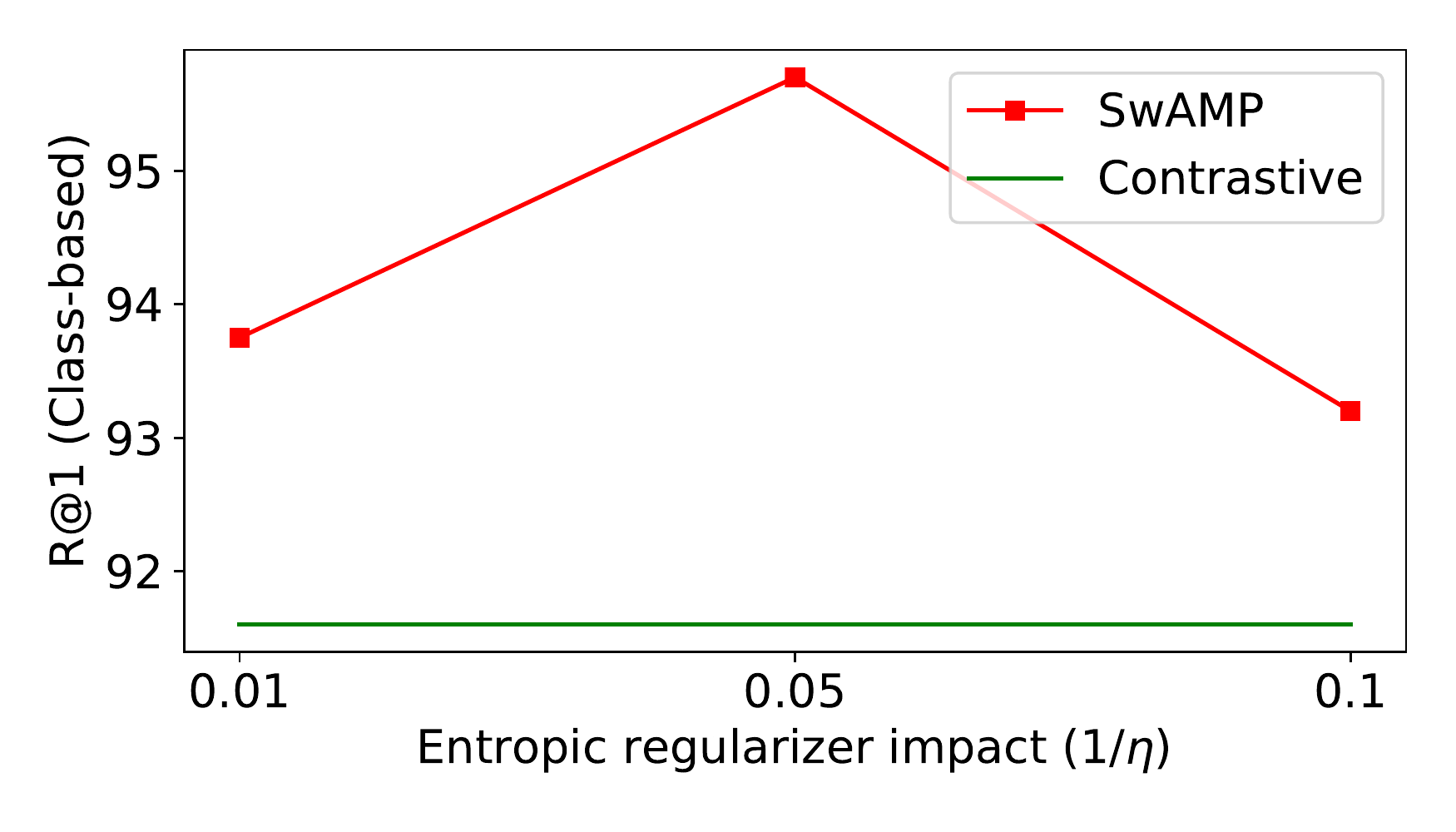}
\end{center}
\vspace{-1.5em}
\caption{(Synthetic data) Impact of entropic regularization ($1/\eta$) in Sinkhorn-Knopp. 
}
\vspace{-1.0em}
\label{appfig:impact_of_eta}
\end{figure}

%%
\textbf{Impact of the entropic regularization ($1/\eta$).} In the Sinkhorn-Knopp (SK) algorithm, we have the reciprocal trade-off $1/\eta$ for the entropy term of the optimization variables $q(y|x)$. Too much emphasizing the entropy term (by increasing $1/\eta$ or decreasing $\eta$) would lead to near uniform $q(y|x)$, which means that it carries little information about the meaningful classes, and cluster assignment can be more or less random. On the other hand, having too small impact of the entropy term would make the SK algorithm converge too slowly, and the output of the SK with only a few iterations would produce non-optimal solutions. To see the impact, we vary $1/\eta$ from 0.01, 0.05 (default), and 0.1, and the results are shown in Fig.~\ref{appfig:impact_of_eta}. We see that there is slight performance degradation for small and large $1/\eta$ values from the optimal choice.

%%%%
%%%%

\begin{figure}[t!]
%\vspace{-1.5em}
\begin{center}
%
%\begin{subfigure}[b]{0.9\textwidth}
\centering
\includegraphics[trim = -1mm 2mm 5mm 4mm, clip, scale=0.345]{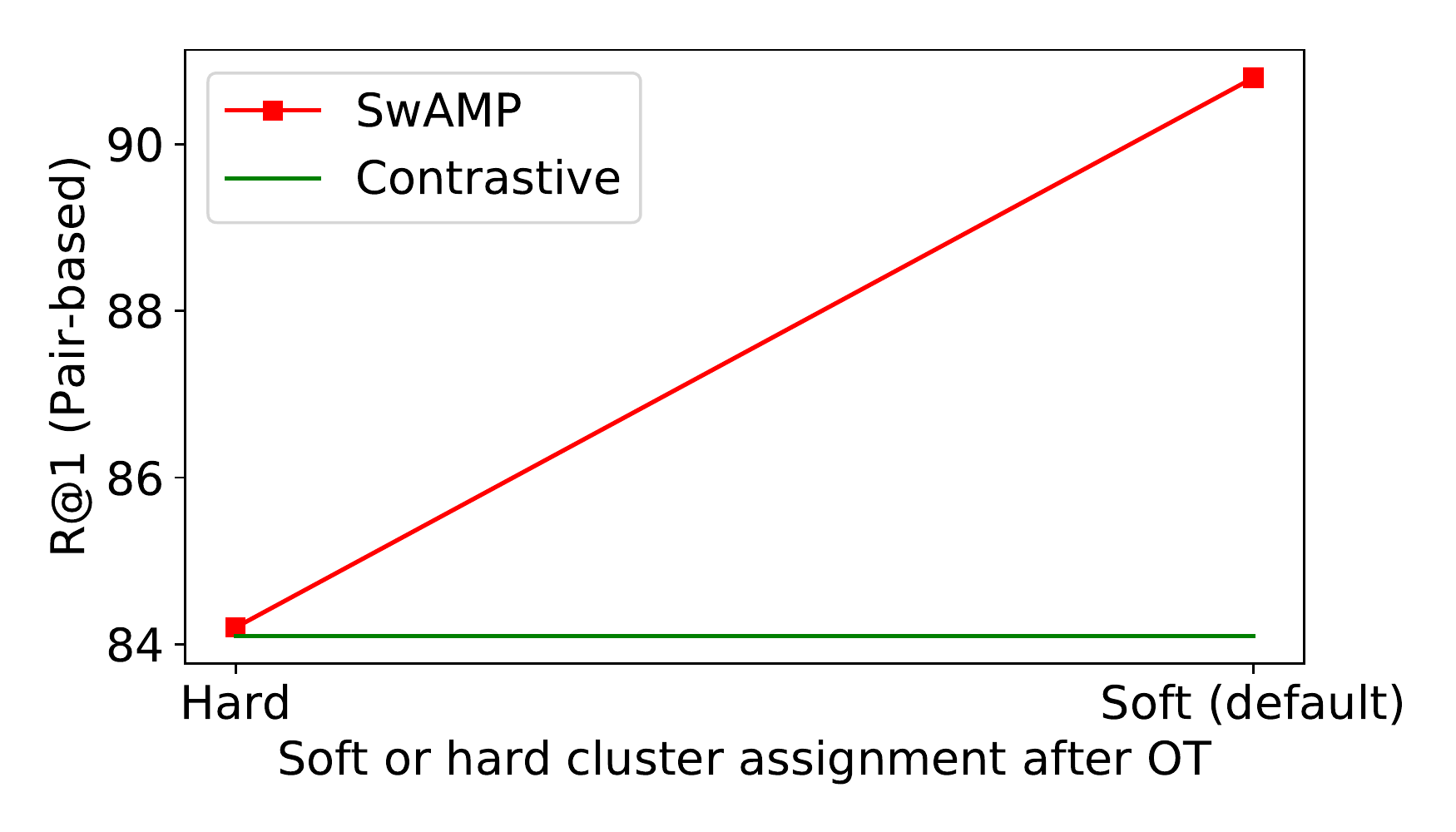} \ \ 
\includegraphics[trim = 5mm 2mm 5mm 4mm, clip, scale=0.345]{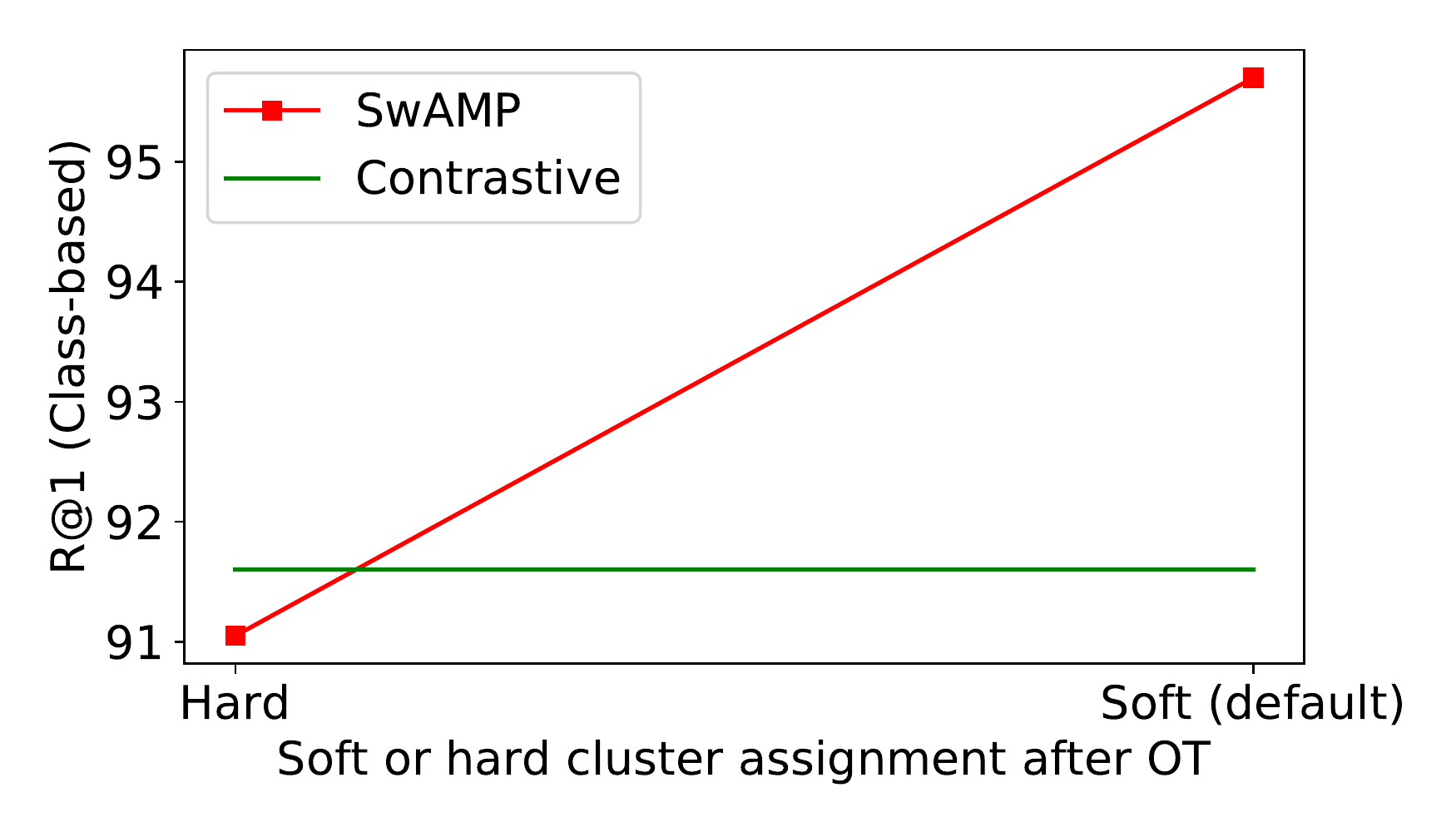}
\end{center}
\vspace{-1.5em}
\caption{(Synthetic data) Soft (default) or hard cluster assignment after OT. 
}
%\vspace{-1.0em}
\label{appfig:soft_hard_ot}
\end{figure}

%%
\textbf{Soft or hard cluster assignment after OT.} We also check if the hard cluster assignment thresholding after OT optimization would be beneficial or not. Recall that the default is to use the output $q(y|x)$ of the SK algorithm as it is (i.e., soft cluster assignment). In the hard assignment we further threshold $q(y|x)$ to have one-hot encoding, which is then used in the cross-entropy loss optimization. As shown in Fig.~\ref{appfig:soft_hard_ot}, the hard assignment is harmful, which implies that retaining uncertainty in cluster estimation is important to have accurate clustering and feature learning. 

%%%%
%%%%

\end{document}